\documentclass[10pt,twocolumn,letterpaper]{article}

\usepackage[pagenumbers]{iccv} 

\usepackage[dvipsnames]{xcolor}
\usepackage{multirow,booktabs}
\usepackage{makecell}
\usepackage{algorithm}
\usepackage{algpseudocode}
\usepackage{subcaption}
\usepackage{balance}
\usepackage{colortbl}
\usepackage{adjustbox} 
\makeatletter

\newcommand{\Rmnum}[1]{\expandafter\@slowromancap\romannumeral #1@}
\makeatother

\definecolor{forestgreen}{RGB}{34,139,34}
\definecolor{colorTab}{rgb}{0.92,0.95,0.92}
\definecolor{light-blue}{rgb}{0.2, 0.5, 1}

\newcommand\blfootnote[1]{%
  \begingroup
  \renewcommand\thefootnote{}\footnote{#1}%
  \addtocounter{footnote}{-1}%
  \endgroup
}

\definecolor{iccvblue}{rgb}{0.21,0.49,0.74}
\usepackage[pagebackref,breaklinks,colorlinks,allcolors=iccvblue]{hyperref}

\def\paperID{9694} 
\def\confName{ICCV}
\def\confYear{2025}

\title{DPoser-X: Diffusion Model as Robust 3D Whole-body Human Pose Prior}

\author{
    Junzhe Lu\textsuperscript{1,*} \quad
    Jing Lin\textsuperscript{2,*} \quad
    Hongkun Dou\textsuperscript{3} \quad
    Ailing Zeng\textsuperscript{4} \quad
    Yue Deng\textsuperscript{3} \\
    Xian Liu\textsuperscript{5} \quad
    Zhongang Cai\textsuperscript{6} \quad
    Lei Yang\textsuperscript{6} \quad
    Yulun Zhang\textsuperscript{7} \quad
    Haoqian Wang\textsuperscript{1,$\dagger$} \quad
    Ziwei Liu\textsuperscript{2,$\dagger$}
    \vspace{1mm}
    \\
    \textsuperscript{1}Tsinghua University \quad
    \textsuperscript{2}Nanyang Technological University \quad
    \textsuperscript{3}Beihang University \quad \\
    \textsuperscript{4}Independent Researcher \thickspace
    \textsuperscript{5}NVIDIA Research \thickspace
    \textsuperscript{6}SenseTime Research \thickspace
    \textsuperscript{7}Shanghai Jiao Tong University \\
    \url{https://dposer.github.io/}
    \vspace{-7mm} 
}

\begin{document}

\twocolumn[{%
    \renewcommand\twocolumn[1][]{#1}%
    \maketitle
    \begin{center}
    \centering
    \captionsetup{type=figure}
    \includegraphics[width=\textwidth]{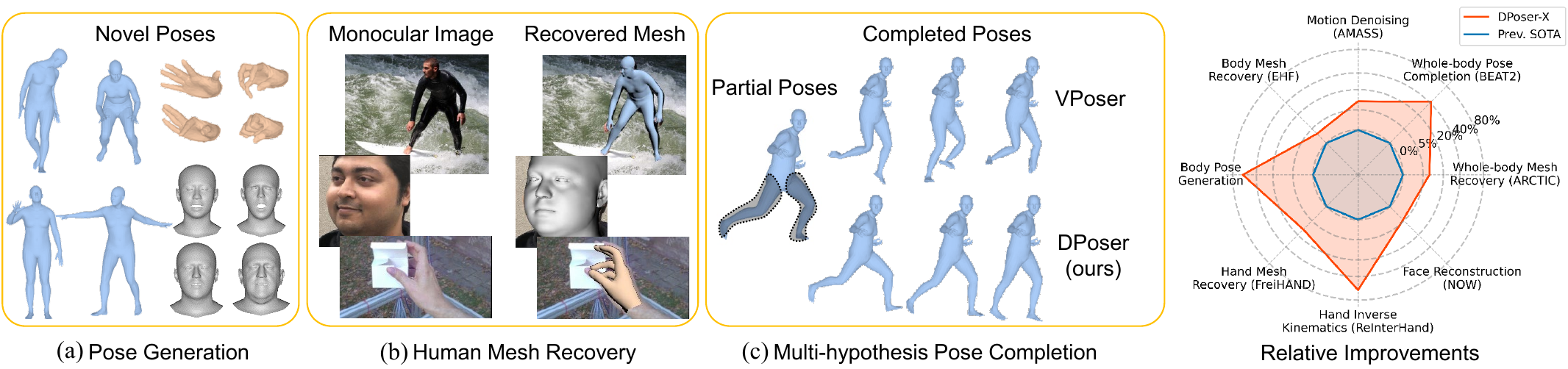}
    \vspace{-7mm}
    \captionof{figure}{An overview of DPoser-X's versatility and performance across multiple pose-related tasks. Built on diffusion models, DPoser-X serves as a robust and adaptable prior for 3D whole-body human pose modeling. Shown are scenarios in (a) pose generation, (b) human mesh recovery, and (c) pose completion. With up to 61\% improvement across 8 benchmarks, DPoser-X consistently outstrips existing priors like VPoser~\cite{pavlakos2019expressive} and NRDF~\cite{he2024nrdf}, proving its superiority in tasks involving the human body, hand, and face. \\} 
    \vspace{-2.5mm}
    \label{fig:teaser}
    \end{center}%
}]

\maketitle

\begin{abstract}
We present \textbf{DPoser-X}, a diffusion-based prior model for 3D whole-body human poses. Building a versatile and robust full-body human pose prior remains challenging due to the inherent complexity of articulated human poses and the scarcity of high-quality whole-body pose datasets. To address these limitations, we introduce a Diffusion model as body Pose prior (DPoser) and extend it to DPoser-X for expressive whole-body human pose modeling.
Our approach unifies various pose-centric tasks as inverse problems, solving them through variational diffusion sampling. To enhance performance on downstream applications, we introduce a novel truncated timestep scheduling method specifically designed for pose data characteristics. We also propose a masked training mechanism that effectively combines whole-body and part-specific datasets, enabling our model to capture interdependencies between body parts while avoiding overfitting to specific actions.
Extensive experiments demonstrate DPoser-X's robustness and versatility across multiple benchmarks for body, hand, face, and full-body pose modeling. Our model consistently outperforms state-of-the-art alternatives, establishing a new benchmark for whole-body human pose prior modeling. 
\vspace{0pt}
\blfootnote{* Equal contribution. $\dagger$ Corresponding authors.}
\end{abstract}

\setlength{\abovedisplayskip}{5pt}
\setlength{\belowdisplayskip}{5pt}

\vspace{-8mm}
\vspace{1.5mm}
\section{Introduction}
\label{sec:intro}

Human pose modeling is a fundamental research topic with broad applications, ranging from human-robot interaction to augmented and virtual reality experiences. Obtaining plausible and realistic human poses requires learning effective pose distribution from large-scale datasets, which can then serve as priors for downstream tasks like body model fitting, motion capture, and gesture recognition.  
Previous approaches to human pose prior modeling have primarily adopted techniques such as Gaussian Mixture Models (GMMs)~\cite{bogo2016keep}, Variational Auto-encoders (VAEs)~\cite{pavlakos2019expressive}, and Neural Distance Fields (NDFs)~\cite{tiwari2022pose, he2024nrdf}. However, each of these approaches faces inherent limitations. GMMs can generate implausible poses due to their unbounded nature, and VAEs enforce a Gaussian prior in the latent that undermines expressiveness. NDFs, while promising for 3D surface modeling, struggle to generalize across the complex, high-dimensional manifolds of human pose. Moreover, due to the scarcity of whole-body pose data, most existing prior models focus on body-only poses, neglecting the whole-body domain which is crucial for detailed human modeling.

To overcome these limitations, recent advances in diffusion models~\cite{ho2020denoising, song2020score} offer a compelling new paradigm. Unlike VAEs with their restrictive latent bottlenecks, diffusion models show greater expressiveness in learning complex distributions. This has led to state-of-the-art results in domains from image synthesis~\cite{dhariwal2021diffusion, karras2022elucidating} to human motion generation~\cite{lu2023humantomato, shafir2023human} and multi-hypothesis pose estimation~\cite{holmquist2023diffpose, ci2023gfpose}. However, these pose-related applications are typically designed for specific generation tasks or tailored to work with conditional inputs. The potential of diffusion models to serve as a universal human pose prior that benefits various pose-related tasks remains largely unexplored.

In this work, we introduce DPoser, a diffusion-based human pose prior that can be seamlessly integrated across diverse pose-related tasks via test-time optimization. Our approach begins with training an unconditional diffusion model. We then formulate various pose-centric tasks as inverse problems, solving them based on variational diffusion sampling~\cite{mardani2023variational}, where DPoser serves as a regularization component. 
Furthermore, our investigations reveal that key pose information during diffusion is concentrated in the later timesteps, leading us to develop a novel truncated timestep scheduling strategy to enhance optimization performance.
Finally, we extend DPoser to DPoser-X for whole-body pose modeling with a mixed training strategy to address the data scarcity issue. Extensive experiments demonstrate that DPoser-X outshines state-of-the-art pose priors in a range of downstream tasks involving the body, hand, and face. 
An overview of DPoser-X's versatility and performance across pose-related tasks is shown in Fig.~\ref{fig:teaser}.

In summary, our main contributions are as follows:
\begin{itemize}
\item We introduce DPoser, a novel framework based on diffusion models that creates a robust and flexible human pose prior applicable across diverse pose-related tasks.

\item We analyze the impact of diffusion timesteps in the pose domain and propose truncated timestep scheduling for more efficient test-time optimization.

\item We present DPoser-X, the first whole-body pose prior, which incorporates a mixed training strategy that effectively leverages both whole-body and part-only datasets.

\item Extensive experiments demonstrate that DPoser-X outstrips previous pose priors on multiple benchmarks for body, hand, face, and full-body modeling.

\end{itemize}

\vspace{-1mm}
\section{Methodology}

\vspace{-1mm}
\subsection{Preliminary: Diffusion Models}
\vspace{-1mm}
\label{sec:preliminary}
Diffusion models~\cite{sohl2015deep, song2019generative, song2020score, ho2020denoising} are a class of generative models that operate by reversing a predefined forward noising process. This forward process systematically corrupts data $\mathbf{x}_0 \sim p_{\mathrm{data}}$ by adding Gaussian noise over a continuous or discrete time variable $t \in [0, 1]$. A noisy sample $\mathbf{x}_t$ at any given time $t$ is produced according to:
\begin{equation}
    \mathbf{x}_t = \alpha_{t}\mathbf{x}_0+\sigma_{t}\epsilon, \quad \epsilon \sim \mathcal{N}(\mathbf{0},\mathbf{I})
    \label{eq:forward diffusion}
\end{equation}
Here, $\alpha_t$ and $\sigma_t$ are predefined schedule functions such that as $t$ increases from 0 to 1, the data distribution is gradually transformed into a tractable prior, typically an isotropic Gaussian distribution $\mathcal{N}(\mathbf{0},\mathbf{I})$.

The generative aspect lies in learning to reverse this process. This is achieved by training a neural network, typically parameterized as a noise predictor $\epsilon_\phi(\mathbf{x}_t; t)$, to estimate the noise component $\epsilon$ from a given noisy sample $\mathbf{x}_t$. The network is optimized using the following L2-loss objective~\cite{ho2020denoising}, where $w(t)$ is a positive weighting function:
\begin{equation}
    \mathbb{E}_{\mathbf{x}_0\sim p_{\mathrm{data}},\epsilon\sim\mathcal{N}(\mathbf{0}, \mathbf{I}),t\sim\mathcal{U}[0,1]} \left[w(t)||\epsilon-\epsilon_\phi(\mathbf{x}_t;t)||_2^2\right],
    \label{eq:training objective}
\end{equation}

Once trained, the model can generate novel data by simulating the reverse diffusion trajectory. This procedure starts with a random sample from the prior, $\mathbf{x}_1 \sim \mathcal{N}(\mathbf{0},\mathbf{I})$, and iteratively applies the learned denoiser $\epsilon_\phi$ to produce a clean sample $\mathbf{x}_0$. In practice, this reverse simulation is implemented by numerical solvers~\cite{song2020score}, such as the discrete samplers introduced in DDPM~\cite{ho2020denoising} and DDIM~\cite{song2020denoising}.

\vspace{-1mm}
\subsection{Learning Pose Prior with Diffusion Models}
\vspace{-1mm}
\label{sec:train_body}
\noindent \textbf{SMPL-based pose representation.}\
To build a flexible 3D human pose prior for the body, we propose to utilize the SMPL body model~\cite{loper2015smpl}, which can be viewed as a differentiable function \([J, V] = M(\theta,\beta)\) that maps body joint angles \(\theta \in \mathbb{R}^{3\times21}\) and shape parameters \(\beta \in \mathbb{R}^{10}\) to mesh vertices \(V \in \mathbb{R}^{3\times6890} \) and joint positions \(J \in \mathbb{R}^{3\times22} \). Our target is to model the distribution of joint angles \(p(\theta)\). 

\noindent \textbf{Training of unconditional diffusion models.}\
To this end, we adopt an unconditional diffusion model to learn the pose representation \(\theta\). 
This approach aligns with a task-agnostic strategy, focusing solely on the distribution of 3D poses. 
For the diffusion process, we employ the sub-VP SDE parameterization proposed in~\cite{song2020score}. Specifically, the coefficients in Eq.~\eqref{eq:forward diffusion} can be obtained as: \(\alpha_{t}=\exp\left(-\frac{1}{2}\int_{0}^{t}\xi(s)\mathrm{d}s\right)\) and \(\sigma_{t}=1-\exp\left(-\int_{0}^{t}\xi(s)\mathrm{d}s\right)\), where \(\xi(t)\) denotes linear scheduled noise scales.

During training, we sample a clean pose \(\theta\) (also \(\mathbf{x}_0\)) from datasets and introduce noise to generate noisy samples \(\mathbf{x}_t\) according to the forward process (Eq.~\eqref{eq:forward diffusion}).
Then we apply the objective in Eq.~\eqref{eq:training objective} to train the noise predictor \(\epsilon_\phi(\mathbf{x}_t;t)\) with weights \(w(t)=\sigma_{t}^2\) as suggested in \cite{song2020score}.

\subsection{Optimization Leveraging Diffusion Priors}
\vspace{-1mm}
\label{sec:DPoser}

The acquired noise predictor, denoted as \(\epsilon_\phi\), permits the generation of novel poses through various samplers. Yet, integrating diffusion priors into general downstream tasks remains largely unexplored. We address this by reframing pose-related tasks as inverse problems and applying variational diffusion sampling~\cite{mardani2023variational} for efficient resolution.

\noindent \textbf{Inverse problem formulation.}\
Given an original signal \(\mathbf{x}_0\) and a degraded measurement $\mathbf{y}$, a typical inverse problem can be formulated as:
\begin{equation}
    \mathbf{y}=\mathcal{A}(\mathbf{x}_0)+\mathbf{n},\quad \mathbf{y},\mathbf{n}\in\mathbb{R}^d,~\mathbf{x}_0\in\mathbb{R}^n,
    \label{eq:inverse problem}
\end{equation}
where \(\mathcal{A}\) symbolizes the degradation pattern and \(\mathbf{n}\) constitutes noise, assumed to be white Gaussian. \(\mathbf{x}_0\) refers to pose representations in human models like SMPL~\cite{loper2015smpl}. This formulation allows us to approach various pose-centric tasks by adapting \(\mathcal{A}\) and interpreting \(\mathbf{y}\) accordingly:
\begin{itemize}
    \item \textbf{Pose completion}: Here, \(\mathcal{A}\) serves as a mask matrix, with \(\mathbf{y}\) being the observed incomplete pose data.
    \item \textbf{Inverse kinematics \& Motion denoising}: In this scenario, \(\mathcal{A}\) applies forward kinematics of human models, treating \(\mathbf{y}\) as the observed noisy 3D joints.
    \item \textbf{Human mesh recovery}: \(\mathcal{A}\) integrates forward kinematics and perspective camera projection to relate \(\mathbf{y}\) to 2D joint observations in images (\ie 2D keypoints).
\end{itemize}
The aim is to recover the original signal \(\mathbf{x}_0\) based on the degraded measurement $\mathbf{y}$. Specifically, our objective shifts to sampling from the posterior distribution \(p\left(\mathbf{x}_0 \mid \mathbf{y}\right)\).

\noindent \textbf{Solving inverse problems with diffusion models.}\
To simulate this posterior sampling process, we adopt the variational diffusion sampling technique~\cite{mardani2023variational}. It employs a variational distribution \( q\left(\mathbf{x}_0 \mid \mathbf{y}\right):= \mathcal{N}(\mu, \sigma^2\mathbf{I}) \) and aims to minimize the KL divergence between this variational distribution and the true posterior \(p\left(\mathbf{x}_0 \mid \mathbf{y}\right)\). 
Further, under the assumption of zero variance (\( \sigma \approx 0 \)), the optimization problem of seeking \(\mathbf{x}_0\) (\ie, \(\mu\)) is demonstrated to be equivalent to minimizing the following loss function~\cite{song2021maximum, mardani2023variational}:
\begin{equation}
    \mathcal{L} = \|\mathbf{y}-\mathcal{A}(\mathbf{x}_0)\|^2+w_t(\mathtt{sg}[\epsilon_\phi(\mathbf{x}_t;t)-\epsilon])^\top\mathbf{x}_0,
    \label{eq:RED}
\end{equation}
where the first term represents the task-specific measurement loss, and the second term corresponds to the regularization loss. Here, \(w_t\) denotes the loss weights, \(\epsilon\) is the standard Gaussian noise, and \( \mathtt{sg} \) signifies stopped-gradient. The regularization procedure initiates by selecting a timestep \( t \) and perturbs the optimization variable \( \mathbf{x}_0 \) as per Eq.~\eqref{eq:forward diffusion}, resulting in \( \mathbf{x}_t \). Then, the gradients \( [\epsilon_\phi(\mathbf{x}_t;t) - \epsilon] \) are applied. 

\begin{figure}[t]
\centering
\includegraphics[width=\linewidth]{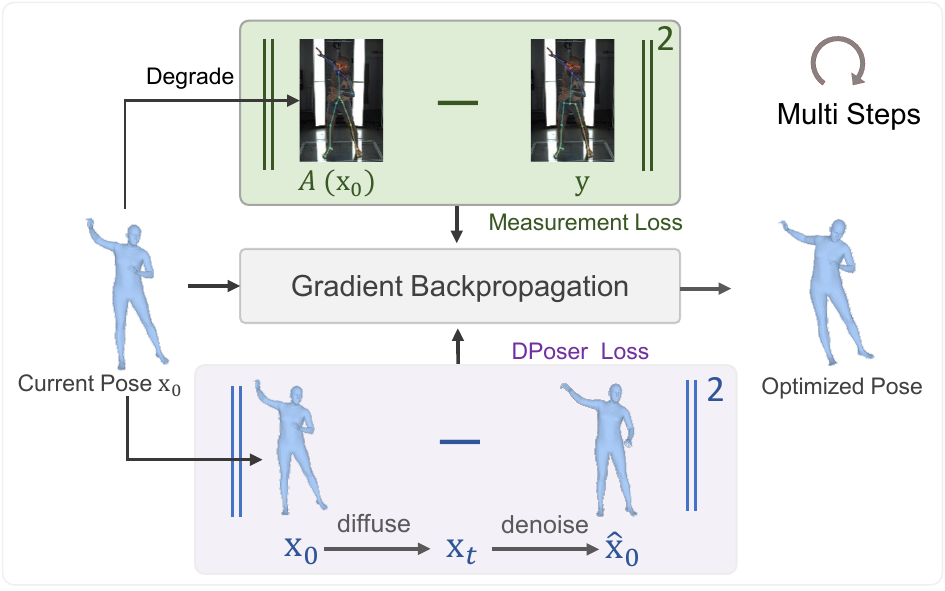}
\vspace{-7mm}
\caption{Overview of the DPoser-regularized optimization framework. Task inputs (\eg, 2D keypoints in human mesh recovery) and current poses are used to compute the measurement loss based on the degradation pattern \( \mathcal{A}(\cdot) \) (\eg, camera projection). 
Meanwhile, DPoser regularization introduces noise to the current pose and applies a one-step denoiser to compute DPoser loss \(L_\text{DPoser}\).}
\label{fig:overview}
\vspace{-6mm}
\end{figure}

\noindent \textbf{Introducing DPoser regularization.}\
To shed more light on the working mechanism, we propose an equivalent but more intuitive regularization term defined as:
\begin{align}
    L_\mathrm{DPoser} &= w_t||\mathbf{x}_0-\mathtt{sg}[\mathbf{\hat{x}}_0(t)]||_2^2, \text{where} \label{eq:DPoser}\\
    \mathbf{\hat{x}}_0(t) &= \frac{\mathbf{x}_t - \sigma_{t}\epsilon_\phi(\mathbf{x}_t;t)}{\alpha_{t}}. 
\end{align}
Here, \(\mathbf{\hat{x}}_0(t)\) functions as one-step denoising estimation using the diffusion model \(\epsilon_\phi(\mathbf{x}_t;t)\), which recovers clean pose from diffused sample \(\mathbf{x}_t\) at any timestep \(t\). The straightforward L2-loss encourages the current pose \(\mathbf{x}_0\) towards a denoised, plausible pose distribution. Further, it is theoretically consistent with the gradient direction of the regularization loss during variational diffusion sampling in Eq.~\eqref{eq:RED}.

\emph{Proof: Differentiating Eq.~\eqref{eq:DPoser} with respect to \(\mathbf{x}_0\) yields:}
\begin{align}
\nabla_{\mathbf{x}_0} L_{\mathrm{DPoser}}&=2w_t(\mathbf{x}_0-\mathbf{\hat{x}}_0(t)) \notag\\
&=2w_t(\frac{\mathbf{x}_t - \sigma_{t}\epsilon}{\alpha_{t}} -\frac{\mathbf{x}_t - \sigma_{t}\epsilon_\phi(\mathbf{x}_t;t)}{\alpha_{t}} ) \notag\\
&=2w_t\frac{\sigma_{t}}{\alpha_{t}} (\epsilon_\phi(\mathbf{x}_t;t)-\epsilon) \notag\\
&\propto (\epsilon_\phi(\mathbf{x}_t;t)-\epsilon). 
\label{eq:SDS gradient}
\vspace{-1.5mm}
\end{align}

\noindent \textbf{DPoser across pose-related tasks.}\ 
As a regularization term, DPoser can be combined with task-specific measurement losses in various pose-related tasks. We demonstrate the applications like human mesh recovery (illustrated in Fig.~\ref{fig:overview}) and inverse kinematics.
Section~\ref{sec:exp} provides test-time details about more domains (\eg, hand and face) and tasks such as pose completion and motion denoising.

Human mesh recovery (HMR) aims to deduce the human pose and shape from monocular images.
In this context, we refine the optimization function in SMPLify~\cite{bogo2016keep}, integrating DPoser as a regularization term, \(L_\mathrm{DPoser}\), and omitting the original intricate interpenetration error component. The modified objective, engaging both body pose \(\theta\) and shape \(\beta\) parameters from the SMPL model~\cite{loper2015smpl}, is defined as: 
\vspace{-1mm}
\begin{equation}
    L(\theta ,\beta ) =  L_\text{HMR} + w_\beta L_\beta + w_\theta L_\mathrm{DPoser}.
    \label{eq:HMR}
\vspace{-1mm}
\end{equation}
The re-projection loss \(L_\text{HMR}\), acting as the measurement loss, is defined by:
\vspace{-1mm}
\begin{equation}
    L_\text{HMR} = \sum_{i \in \text{Joints}} \lambda_i \rho\left(\Pi_C\left(M_J(\theta, \beta)_i\right) - J^{\text{est}}_i\right),
    \label{eq:reprojection}
\vspace{-1mm}
\end{equation}
where \(M_J(\theta, \beta)\) denotes SMPL's forward kinematics. The camera function \(\Pi_C\) maps 3D joint coordinates into 2D space. \(J^{\text{est}}\) refers to the 2D keypoints estimated using off-the-shelf 2D pose estimators, with \(\lambda_i\) reflecting the confidence score. The Geman-McClure error function (\(\rho\)) is employed to assess the discrepancy in 2D joint locations. 
To avoid unrealistic poses when minimizing re-projection loss, our DPoser regularization \(L_\mathrm{DPoser}\) is introduced on the body pose \(\theta\). Moreover, a shape regularization term \(L_\beta = \|\beta\|_2^2\) is utilized to constrain the body shapes. Their weights are expressed as \(w_\theta\), and \(w_\beta\) respectively.

Another application is inverse kinematics (IK), which estimates poses from noisy or incomplete 3D joint positions, $J^{\text{obs}}$, where the set of known joints is assumed to be provided. The task-specific measurement loss, $L_\text{IK}$, is an L2-loss between the model's 3D joints and the observed ones:
\begin{equation}
    L_\text{IK} = \sum_{i \in \text{Known Joints}} ||M_J(\theta, \beta)_i - J^{\text{obs}}_i||_2^2.
    \label{eq:IK}
\end{equation}
As with HMR, the full objective combines \(L_\text{IK}\) with our DPoser regularization \(L_\mathrm{DPoser}\). The inclusion of \(L_\mathrm{DPoser}\) is crucial for ensuring plausible poses, especially when 3D joint observations are sparse or noisy.

Given the structure of \(L_\mathrm{DPoser}\), selecting the suitable diffusion timestep \(t\) is essential in the iterative optimization process. In the subsequent section, we address this by introducing our novel truncated timestep scheduling.

\vspace{-1mm}
\subsection{Test-time Truncated Timestep Scheduling}
\begin{figure}[t]
  \centering
  \includegraphics[width=0.45\textwidth]{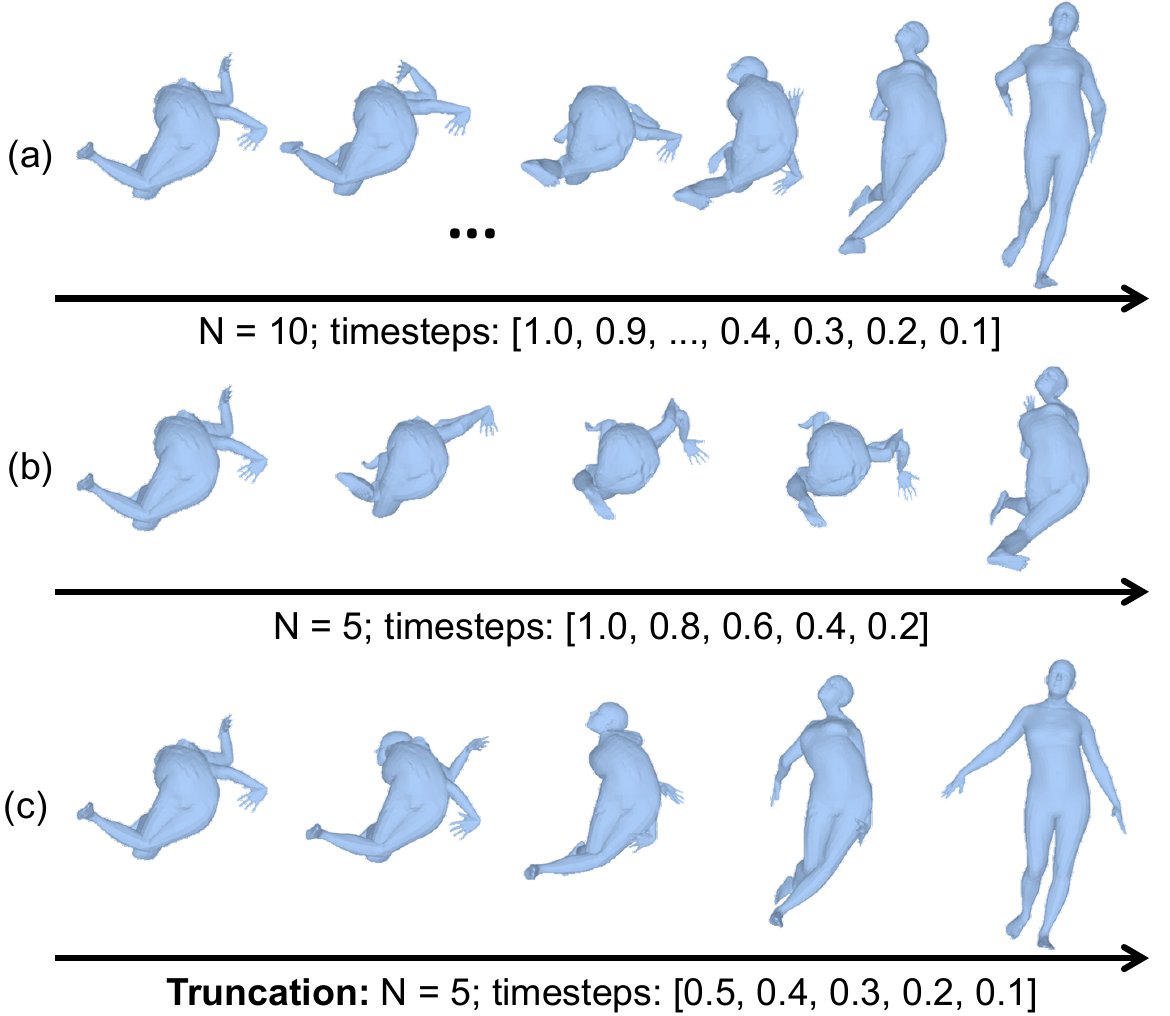}
  \vspace{-2mm}
  \caption{Illustration of the rationale behind the proposed truncated timestep scheduling. We employ the DDIM sampler~\cite{song2020denoising} with limited steps and visualize the generated poses. Our observations reveal that pose refinement occurs at later timesteps.}
  \label{fig:truncation}
\vspace{-3mm}
\end{figure}

\vspace{-1mm}
\label{sec:truncated_timestep}
In the diffusion process, previous research~\cite{choi2022perception} on images shows that initial timesteps (larger \(t\)) correspond to the perceptual content, while later timesteps refine details. Pose data, however, lacks the similar structured layering and spatial redundancy, implying a need for a tailored timestep scheduling approach.
As depicted in Fig.~\ref{fig:truncation}, we find that pose generation does not benefit from the early timesteps as image generation does. The significant stages of pose refinement occur at smaller \(t\), specifically when \(t\le0.3\). With limited steps, the uniform scheduling, as tested in (b), proves less effective. In contrast, allocating these steps toward the latter end of the diffusion process, as in (c), yields better samples. This indicates that critical pose information is concentrated more heavily in the later timesteps.

Based on the above insights, we propose a shift from standard uniform timestep scheduling~\cite{chung2022diffusion, mardani2023variational} to a truncated strategy for pose data. 
Specifically, the timestep \(t\) for each optimization step can be expressed as:
\vspace{-1mm}
\begin{equation}
     t = t_{\text{max}} - \frac{(t_{\text{max}} - t_{\text{min}}) \times \text{iter}}{N-1}.
\vspace{-1mm}
\end{equation}
where \(N\) denotes the number of optimization iterations, and \(\text{iter}\) signifies the current iteration. Note that when the interval \([t_{max}, t_{min}]\) is set to ([1.0, 0.0]), this strategy degenerates to the uniform linear timestep schedule used in prior works~\cite{chung2022diffusion, mardani2023variational}. This formulation is integral to our optimization framework, which is summarized in Alg. \ref{alg:test-time-optimization}. 
The timestep interval is chosen based on the task's noise scale (typically \([0.15, 0.05]\)). See Section~\ref{sec:timestep_analysis} for details.

\begin{algorithm}[t]
\caption{Test-time Optimization with DPoser}
\label{alg:test-time-optimization}
\begin{algorithmic}[1]
\Require 
A trained diffusion model \( \epsilon_\phi(\mathbf{x}_t; t) \), task-specific loss \( L_{\text{task}} \), range of diffusion timesteps \([t_{\text{max}}, t_{\text{min}}]\), number of optimization iterations \( N \).
\Ensure
Initialization of pose parameters \( \mathbf{x}_0 \)
\For{\( \text{iter} = 0, 1, \ldots, N-1 \)}
    \State \( t \leftarrow t_{\text{max}} - \frac{(t_{\text{max}} - t_{\text{min}}) \times \text{iter}}{N-1} \)   \Comment{Timestep scheduling}
    \State Sample \( \epsilon \sim \mathcal{N}(0, I) \)
    \State \( \mathbf{x}_t \leftarrow \alpha_t \mathbf{x}_0 + \sigma_t \epsilon \)  \Comment{Forward diffusion}
    \State \( \mathbf{\hat{x}}_0(t) \leftarrow \frac{\mathbf{x}_t - \sigma_t \epsilon_\phi(\mathbf{x}_t; t)}{\alpha_t} \) \Comment{One-step denoiser}
    \State \( L_{\text{DPoser}} \leftarrow w_t \lVert \mathbf{x}_0 - \text{sg}[\mathbf{\hat{x}}_0(t)] \rVert_2^2 \) \Comment{Regularization}
    \State \( L_{\text{total}} \leftarrow L_{\text{task}} + L_{\text{DPoser}} \)
    \State Update \( \mathbf{x}_0 \) via backpropagation on \( L_{\text{total}} \)
\EndFor
\State \textbf{return} \( \mathbf{x}_0 \)
\end{algorithmic}
\end{algorithm}

\vspace{-1mm}
\subsection{Building Whole-body Pose Prior}
\label{sec:DPoser_wholebody}
\begin{figure*}[t]
  \centering
  \includegraphics[width=0.98\textwidth]{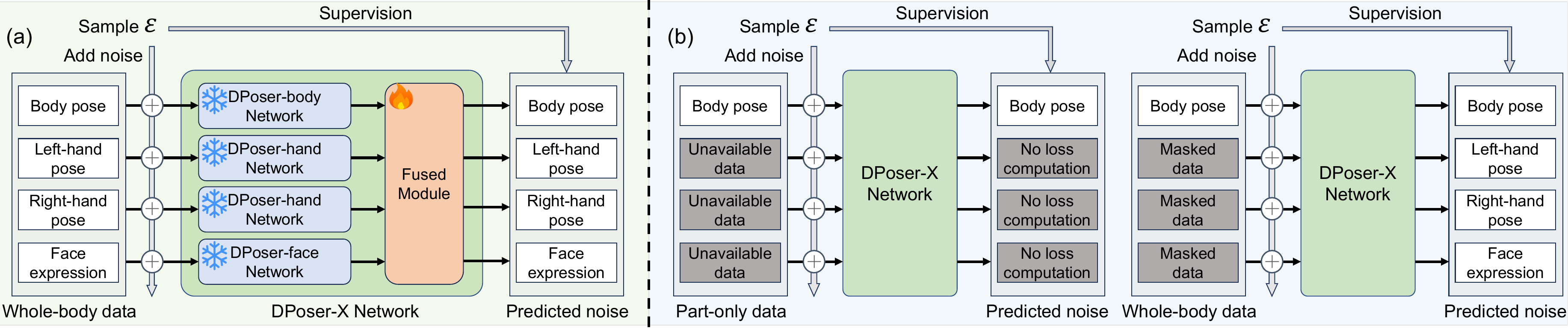}
  \vspace{-2mm}
  \caption{Overview of the DPoser-X methodology. (a) The whole-body network consists of frozen part-only networks, and a fused module trained on whole-body datasets. (b) The mixed training strategy utilizes part-only datasets by applying loss only to available parts. To prevent arbitrary predictions on unavailable parts, the whole-body data is sometimes randomly masked, and loss is applied to all parts.}
  \label{fig:wholebody_pipeline}
\vspace{-4mm}
\end{figure*}

To model the whole-body pose, we separately train three part-specific DPoser models to capture the distribution of body pose, hand pose, and facial expressions, following the training mechanism described in Section \ref{sec:train_body}. 
For DPoser-hand, we adopt the MANO~\cite{romero2022embodied} model with the learned target being hand joint angles \(\theta_{hand} \in \mathbb{R}^{3 \times 15}\). For DPoser-face, we utilize the FLAME~\cite{li2017learning} model, where the modeling target \(\theta_{face} \in \mathbb{R}^{103}\) includes 100 facial expression coefficients and a 3-dimensional jaw pose. 
As a baseline, the three models are combined directly, referred to as \textit{DPoser-X-base}. Specifically, it splits the whole-body pose input into body, hand, and face parts, processing each through its respective part-specific model.

DPoser-X-base, however, fails to capture the interactions between different parts, which is crucial for whole-body pose modeling. For instance, in relaxed standing poses, the left and right hand poses are usually mirrored. To address this issue, we introduce a fused module after the DPoser-X-base model and train it on whole-body datasets, resulting in \textit{DPoser-X-fused}. 
While this model performs well in tasks like completion, it exhibits limited pose diversity, as demonstrated in Section \ref{sec:ablation}. This limitation stems from the fact that existing whole-body pose datasets mainly capture specific actions (e.g., speech gestures or object-grabbing scenarios), resulting in degraded generalization.

To improve this, we propose a mixed training strategy, which utilizes whole-body, body-only, hand-only, two-hands, and face-only datasets, leading to the \textit{DPoser-X-mixed} model. Specifically, we treat part-only data as incomplete whole-body data, applying loss only to the available parts. 
In addition, to reduce the data type gap, we randomly mask the whole-body data and apply loss across all parts, forcing the network to predict masked parts.
In practice, we implement the unavailable and masked data parts as the mean poses and enable masking at a probability of 20\%.
This mixed training strategy enables DPoser-X-mixed to maintain whole-body modeling ability while leveraging part-only datasets to enhance generalization.

\vspace{-1mm}
\section{Experiments}
\vspace{-1mm}
In this section, we showcase the robustness and versatility of DPoser-X across a wide range of pose-centric tasks involving the human body, hand, face, and whole-body. Section~\ref{sec:exp} and \ref{sec:addtional_experiments} provide evaluation metrics and complete assessments including but not limited to body pose completion, hand mesh recovery, and face generation.

\vspace{-1mm}
\subsection{Experimental Setup}
\vspace{-1mm}
\label{sec:exp_setup}
\noindent \textbf{Implementation details.}\
DPoser-body is trained on the AMASS dataset~\cite{mahmood2019amass} with the same splits as prior works~\cite{pavlakos2019expressive, tiwari2022pose}.
The model employs axis-angle representation for joint rotations, which we normalize to have zero mean and unit variance. The architecture consists of a fully connected neural network with about 8.28M parameters. It draws inspiration from GFPose~\cite{ci2023gfpose} but omits conditional input pathways for our unconditional setting.
We train this model for 800,000 iterations using the Adam optimizer with a learning rate of \(2 \times 10^{-4}\) and a batch size of 1280.

DPoser-hand and DPoser-face networks use a similar architecture.
DPoser-hand uses the FreiHAND, DexYCB, HO3D, H2O, and ReInterHand datasets~\cite{zimmermann2019freihand,chao2021dexycb,hampali2020honnotate,kwon2021h2o,moon2024dataset}. DPoser-face employs WCPA and MICA datasets~\cite{kao2022single, zielonka2022towards}.
The DPoser-X network integrates last-layer features from pre-trained part models using a fully connected neural network with identical residuals. 
Whole-body datasets include BEAT2, GRAB, ARCTIC, and EgoBody~\cite{liu2024emage,taheri2020grab,fan2023arctic,zhang2022egobody}. 
After data source weight balancing, based on our mixed training strategy, DPoser-X-mixed is trained on around 65\% whole-body, 14\% body-only, 12\% single-hand, 4\% two-hand, and 5\% face-only data.

\noindent \textbf{Comparison settings.}\
We compare against SOTA pose priors including GMM~\cite{bogo2016keep}, VPoser~\cite{pavlakos2019expressive}, GAN-S~\cite{davydov2022adversarial}, Pose-NDF~\cite{tiwari2022pose}, and NRDF~\cite{he2024nrdf}.
The above works focus on the body, so we have trained the hand versions of VPoser and NRDF. Since NRDF relies on quaternion representations, VPoser is trained for face and whole-body comparisons. We also include an L2-regularization baseline. See Section~\ref{sec:comparative_implementation} for details of comparative methods implementation.

\begin{figure}[t]
    \centering
    \subfloat[GAN-S~\cite{davydov2022adversarial}]{
        \includegraphics[width=0.31\linewidth]{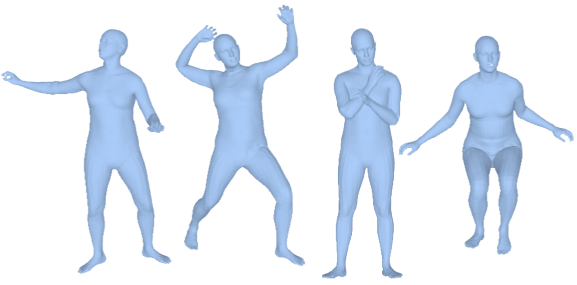}
    }
    \subfloat[Pose-NDF~\cite{tiwari2022pose}]{
        \includegraphics[width=0.31\linewidth]{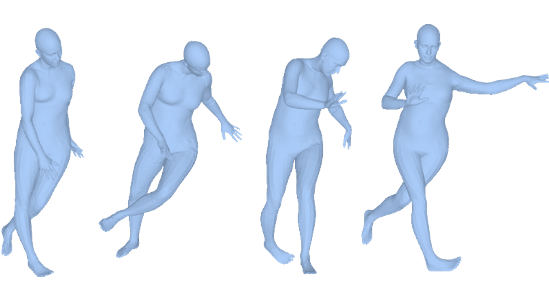}
    }
    \subfloat[NRDF~\cite{he2024nrdf}]{
        \includegraphics[width=0.31\linewidth]{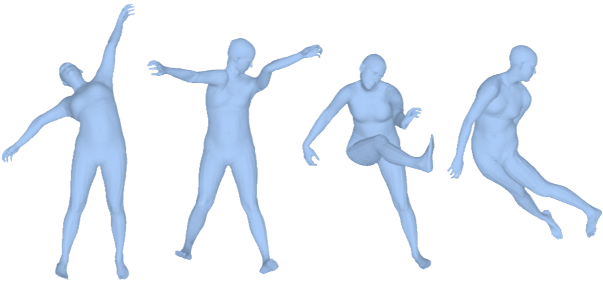}
    }
    \hfill
    \subfloat[VPoser~\cite{pavlakos2019expressive}]{
        \includegraphics[width=0.31\linewidth]{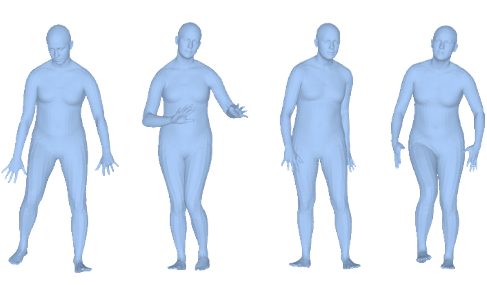}
    }
    \subfloat[DPoser (ours)]{
        \includegraphics[width=0.31\linewidth]{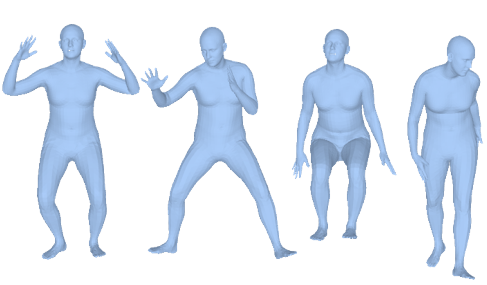}
    }
    \subfloat[DPoser (ours)*]{
        \includegraphics[width=0.31\linewidth]{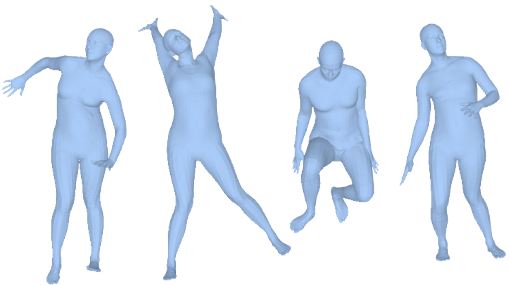}
    }
    \vspace{-3mm}
    \caption{Qualitative comparison of generated human poses: (e) illustrates naturalistic poses aligned with real-world data, whereas (f) shows poses that, despite superior APD, lack natural appearance. *We use a DDIM sampler~\cite{song2020denoising} with only 10 steps.}
    \label{fig:generation}
    \vspace{-3mm}
\end{figure}

\begin{table}[t]
  \centering
  \footnotesize
  \begin{tabular}{lccccc}
    \toprule[\heavyrulewidth]
    Sample source & APD $\uparrow$ & FID $\downarrow$ & Prec. $\uparrow$ & Rec. $\uparrow$ & \(d_{NN}\)$\downarrow$ \\
    \midrule[\lightrulewidth]
    Real-world~\cite{mahmood2019amass} & 15.44 & 0.005 & 0.86 & 0.90 & 0.001 \\
    \midrule[\lightrulewidth]
    GMM~\cite{bogo2016keep} & 16.28 & 1.02 & 0.13 & 0.34 & 4.37 \\
    VPoser~\cite{pavlakos2019expressive} & 10.75 & 0.66 & 0.29 & 0.42 & 3.74 \\
    GAN-S~\cite{davydov2022adversarial} & 15.68 & \underline{0.18} & \underline{0.61} & 0.41 & 2.98 \\
    Pose-NDF~\cite{tiwari2022pose} & 18.75 & 5.92 & 0.02 & 0.00 & 9.08 \\
    NRDF~\cite{he2024nrdf} & \textbf{22.82} & 0.64 & 0.03 & \textbf{0.99} & 6.69 \\
    \rowcolor{colorTab}
    DPoser & 14.28 & \textbf{0.07} & \textbf{0.72} & 0.80 & \textbf{2.63} \\
    \rowcolor{colorTab}
    DPoser* & \underline{19.03} & \underline{0.58} & 0.10 & \underline{0.95} & \underline{2.95} \\
    \bottomrule[\heavyrulewidth]
  \end{tabular}
  \vspace{-2mm}
  \caption{Comparative analysis of pose generation metrics. *Indicates the use of a 10-step DDIM sampler.}
  \label{tab:generation}
  \vspace{-6.5mm}
\end{table}

\begin{figure*}[t]
    \centering
    \includegraphics[width=\linewidth]{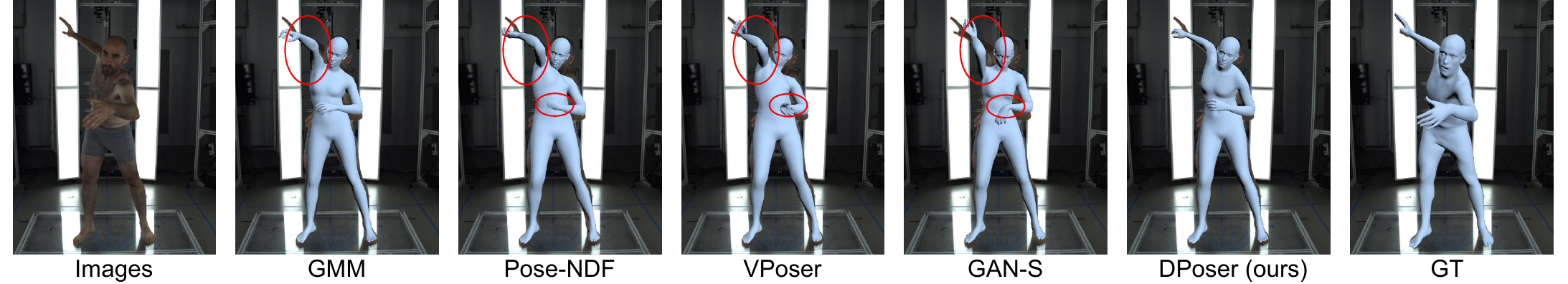}
    \vspace{-7mm}
    \caption{Visualization of human mesh recovery results (body only) on EHF~\cite{pavlakos2019expressive} when fitting from scratch.}
    \label{fig:HMR-compare}
    \vspace{-2.3mm}
\end{figure*}

\begin{table*}[!t]
  \centering
  \footnotesize
    \begin{tabular}{lccccccc}
    \toprule[\heavyrulewidth]
    Initialization & w/o fitting & GMM~\cite{bogo2016keep} & VPoser~\cite{pavlakos2019expressive} & Pose-NDF~\cite{tiwari2022pose} & NRDF~\cite{he2024nrdf} & GAN-S~\cite{davydov2022adversarial} & DPoser \\
    \midrule[\lightrulewidth]
    from scratch  & 108.57 & 58.32 & 58.08 & 57.87 & 57.38 & 57.26 & \textbf{56.05} \\
    CLIFF~\cite{li2022cliff}  & 56.62 & 51.02 & 49.39 & 49.50 & 49.27 & 49.58 & \textbf{49.05} \\
    \bottomrule[\heavyrulewidth]
    \end{tabular}
  \vspace{-2mm}
  \caption{Performance comparison of human mesh recovery on the EHF dataset~\cite{pavlakos2019expressive}. PA-MPJPE is reported as the metric.}
\label{tab:HMR}
\vspace{-5mm}
\end{table*}

\vspace{-2mm}
\subsection{Body-only Tasks}
\vspace{-1mm}
\label{sec:exp_body}
\noindent \textbf{Body pose generation.}\
We conduct generation experiments to assess the learned data distribution of pose priors. Since generation is not an inverse problem, we employ a standard Euler-Maruyama discretization~\cite{song2020score} with 1000 steps for DPoser's pose generation.
As shown in Fig.~\ref{fig:generation}, DPoser generates visually diverse and realistic poses, indicating a well-learned prior distribution. In contrast, VPoser~\cite{pavlakos2019expressive} exhibits limited diversity due to its mean-centric nature, constrained by the explicit Gaussian latents. Additionally, both Pose-NDF~\cite{tiwari2022pose} and NRDF~\cite{he2024nrdf} struggle to project pure noise to plausible poses, resulting in unnatural outputs due to their limited generalization capabilities.

In terms of quantitative evaluation (Table~\ref{tab:generation}), Pose-NDF and NRDF show high Average Pairwise Distance (APD) but poor Precision and \(d_{NN}\). The high APD is likely due to exaggerated poses, which should be avoided in pose priors.
To verify this, we test a 10-step DDIM sampler~\cite{song2020denoising} named DPoser* that is suboptimal by design. Despite superior APD, DPoser* performs poorly in FID and Precision, indicating that too few steps lead to divergence from the expected distribution.
Our findings highlight the need for a balanced evaluation of quantitative and qualitative results.

\noindent \textbf{Human mesh recovery.}\
We probe the efficacy of DPoser in HMR, focusing on estimating human body pose and shape from monocular images. Following Pose-NDF~\cite{tiwari2022pose}, we conduct experiments on the EHF dataset~\cite{pavlakos2019expressive} and benchmark our method against existing SOTA priors. 
Our optimization-based framework incorporates two initialization paradigms: (1) a baseline that utilizes mean poses, and (2) an advanced scheme that employs CLIFF~\cite{li2022cliff}, a pre-trained regression-based model tailored for HMR. We further compare with recent generation-based methods GFPose~\cite{ci2023gfpose} and HuProSO3~\cite{dunkel2024normalizing} in Section~\ref{sec:addtional_hmr}.

As presented in Table~\ref{tab:HMR} and Fig.~\ref{fig:HMR-compare}, when fitting from scratch, DPoser surpasses established SOTA priors like GAN-S~\cite{davydov2022adversarial} and NRDF~\cite{he2024nrdf}. 
Moreover, DPoser can refine initial pose estimations from SOTA regression-based models like CLIFF~\cite{li2022cliff}, aligning better with images.

\begin{table}[t]
    \centering
    \footnotesize
    \begin{tabular}{lcc}
        \toprule[\heavyrulewidth]
        Methods & AMASS~\cite{mahmood2019amass} & HPS~\cite{guzov2021human}\\
        \midrule[\lightrulewidth]
        w/o prior & 24.19 & 23.67 \\
        VPoser~\cite{pavlakos2019expressive} & 23.42 & 22.78 \\
        Pose-NDF~\cite{tiwari2022pose} & 22.13 & 21.60 \\
        MVAE~\cite{ling2020character} & 26.80 & N/A \\
        HuMoR~\cite{rempe2021humor} & 22.69 & N/A \\
        \rowcolor{colorTab}
        DPoser & \textbf{19.87} & \textbf{20.54} \\
        \bottomrule[\heavyrulewidth]
    \end{tabular}
    \vspace{-1.9mm}
    \caption{Performance metrics (MPJPE) for motion denoising.}
    \label{tab:motion}
    \vspace{-2.6mm}
\end{table}

\noindent \textbf{Motion denoising.}\
Though not initially designed for temporal tasks, DPoser adapts well in motion denoising. The task aims to recover clean body poses from noisy 3D joint positions in motion sequences.
Following Pose-NDF~\cite{tiwari2022pose} and HuMoR~\cite{rempe2021humor}, we apply Gaussian noise (standard deviation of 40 mm) to the 60-frame sequences from the AMASS dataset~\cite{mahmood2019amass}. We also test on the HPS dataset~\cite{guzov2021human} without additional training to validate generalization.
As shown in Table \ref{tab:motion}, DPoser sets a new standard in motion denoising, outperforming even specialized motion priors like HuMoR.

\vspace{-1.5mm}
\subsection{Hand-only and Face-only Tasks}
\vspace{-1.5mm}
\label{sec:exp_hand_face}

\begin{table}[t]
  \centering
  \footnotesize
  \begin{tabular}{lccccc}
    \toprule[\heavyrulewidth]
    \makecell[c]{\multirow{2}{*}[-1.0ex]{Methods}} & \makecell[c]{\multirow{2}{*}[-1.0ex]{Setting}} & \multicolumn{3}{c}{MPJPE $\downarrow$} & MPVPE $\downarrow$ \\
    \cmidrule(lr){3-5} \cmidrule(l){6-6}
    & & Vis. & Occ. & All. & All. \\
    \midrule[\lightrulewidth]
    w/o prior & sparse & 0.07 & 14.46 & 8.98 & 10.07 \\
    L2 prior & sparse & 0.84 & 13.84 & 8.89 & 8.94 \\
    VPoser~\cite{pavlakos2019expressive} & sparse & 0.37 & 13.10 & 8.25 & 8.84 \\
    NRDF~\cite{he2024nrdf} & sparse & 0.11 & 13.92 & 8.66 & 9.53 \\
    \rowcolor{colorTab}
    DPoser-hand & sparse & \textbf{0.06} & \textbf{5.15} & \textbf{3.21} & \textbf{3.43} \\
    \midrule[\lightrulewidth]
    w/o prior & fingertip & 0.13 & 4.00 & 2.89 & 4.59 \\
    L2 prior & fingertip & 1.02 & 3.58 & 2.85 & 3.15 \\
    VPoser~\cite{pavlakos2019expressive} & fingertip & 0.48 & 4.35 & 3.25 & 3.93 \\
    NRDF~\cite{he2024nrdf} & fingertip & 0.15 & 3.95 & 2.93 & 3.95 \\
    \rowcolor{colorTab}
    DPoser-hand & fingertip & \textbf{0.07} & \textbf{2.40} & \textbf{1.74} & \textbf{1.99} \\
    \bottomrule[\heavyrulewidth]
  \end{tabular}
  \vspace{-2mm}
  \caption{Quantitative evaluation of hand inverse kinematics on the ReInterhand dataset~\cite{moon2024dataset} under various masking settings.}
  \label{tab:inverse_kinematics_hand}
  \vspace{-5.8mm}
\end{table}

\noindent \textbf{Hand inverse kinematics.}\
We assess DPoser’s performance in hand inverse kinematics (IK) tasks using the ReInterhand dataset~\cite{moon2024dataset}, considering various challenging conditions. Table~\ref{tab:inverse_kinematics_hand} summarizes results across the two settings: sparse (60\% keypoints masked) and fingertip (only 5 fingertip keypoints visible). DPoser consistently outperforms baselines, achieving the lowest error. Notably, in the sparse setting, DPoser reduces MPJPE by over 50\% compared to other methods, showcasing its robustness in recovering accurate hand poses from limited observations.

\begin{table}[t]
  \centering
  \footnotesize
  \begin{tabular}{lcc}
    \toprule[\heavyrulewidth]
    Methods & all & side-view \\
    \midrule[\lightrulewidth]
    w/o prior & 12.97/16.07/13.15 & 13.15/16.22/13.26 \\
    L2 prior & 12.21/15.15/12.74 & 12.29/15.22/12.62 \\
    VPoser~\cite{pavlakos2019expressive} & 12.23/15.13/12.71 & 12.20/15.35/12.98  \\
    \rowcolor{colorTab}
    DPoser-face & \textbf{11.68}/\textbf{14.58}/12.17 & \textbf{11.77}/\textbf{14.67}/12.34 \\
    \midrule[\lightrulewidth]
    MICA~\cite{zielonka2022towards} & 9.03/11.12/9.24 & 9.29/11.71/10.04 \\
    + w/o prior & 9.01/11.09/9.15 & 9.47/11.80/9.84 \\
    + L2 prior & 9.96/12.37/10.44 & 10.01/12.49/10.62 \\
    + VPoser~\cite{pavlakos2019expressive} & 9.93/12.34/10.43 & 10.00/12.50/10.65 \\
    \rowcolor{colorTab}
    + DPoser-face & \textbf{8.76}/\textbf{10.78}/9.00 & \textbf{9.18}/\textbf{11.47}/9.73 \\
    \bottomrule[\heavyrulewidth]
  \end{tabular}
  \vspace{-1.8mm}
  \caption{Face reconstruction performance on the NOW benchmark~\cite{sanyal2019learning}. Results are reported as median/mean/std of MPVPE.}
  \label{tab:face_recon_Now}
  \vspace{-4mm}
\end{table}

\noindent \textbf{Face reconstruction.}\
We evaluate DPoser on face reconstruction tasks using the NOW~\cite{sanyal2019learning} benchmark. The target is to estimate face shape accurately from a single image.
To assess model performance in more challenging scenarios, we collect and test on a side-view subset of NOW. For initialization, in addition to fitting from scratch, we use the SOTA face reconstruction model MICA~\cite{zielonka2022towards}.

\begin{figure}[t]
    \centering
    \includegraphics[width=0.98\linewidth]{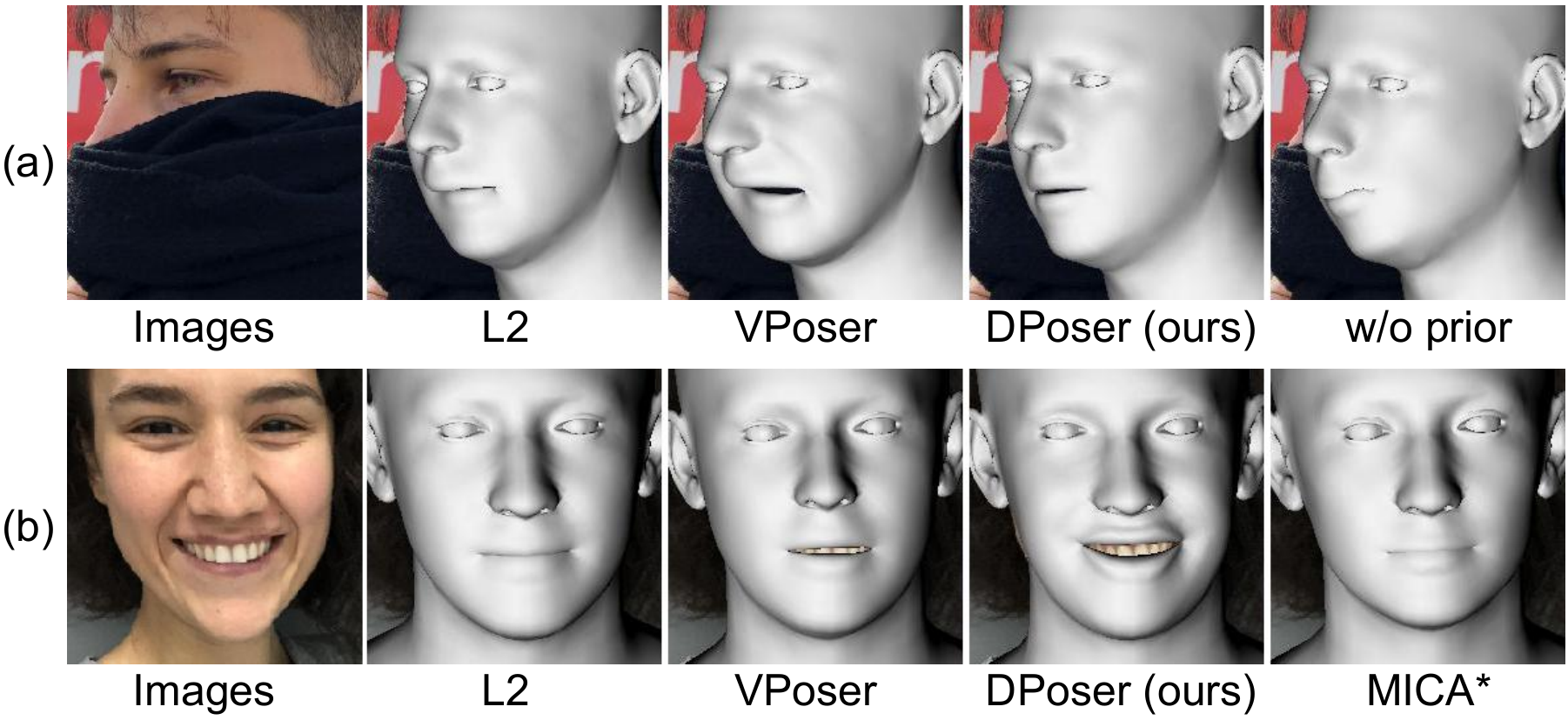}
    \vspace{-2.4mm}
    \caption{Visualization of face reconstruction on the NOW benchmark~\cite{sanyal2019learning}. (a) Fitting from scratch. (b) Initialization using MICA~\cite{zielonka2022towards}. *MICA predicts only face shape without expressions; translational and global orientation are fitted for visualization.}
    \label{fig:now_comparison}
    \vspace{-2mm}
\end{figure}

As shown in Table~\ref{tab:face_recon_Now}, when fitting from scratch, DPoser achieves the lowest reconstruction errors across both overall and side-view cases. With MICA initialization, DPoser achieves the best performance, reducing mean error to 8.76 mm. In contrast, due to their mean-centric characteristics, L2 prior and VPoser~\cite{pavlakos2019expressive} do not improve on MICA’s results, getting even worse results than the baseline without pose prior. Visualizations in Fig.~\ref{fig:now_comparison} show DPoser’s ability to reconstruct realistic faces, handling occlusion effectively.

\vspace{-1.2mm}
\subsection{Whole-body Tasks}
\vspace{-1.2mm}
\label{sec:exp_wholebody}

\begin{table}[!t]
  \vspace{-1mm}
  \centering
  \footnotesize
  \setlength{\tabcolsep}{3pt}
  \begin{tabular}{lcccccc}
    \toprule[\heavyrulewidth]
    Methods & $\text{APD}_{\text{body}}
 $ $\uparrow$ & $\text{APD}_{\text{hands}}
 $ $\uparrow$ & FID $\downarrow$ & Prec. $\uparrow$ & Rec. $\uparrow$ \\
    \midrule[\lightrulewidth]
    VPoser-X~\cite{pavlakos2019expressive} & 7.79 & 1.16 & 14.52 & \textbf{0.64} & 0.07 \\
    DPoser-X-base & \textbf{14.45} & \textbf{2.34} & 90.78 & 0.00 & 0.00 \\
    DPoser-X-fused & 11.78 & {1.93} & \textbf{3.71} & \underline{0.32} & \underline{0.77} \\
    \rowcolor{colorTab}
    DPoser-X-mixed & \underline{14.08} & \underline{2.04} & \underline{13.97} & 0.02 & \textbf{0.81} \\
    \bottomrule[\heavyrulewidth]
  \end{tabular}
  \vspace{-1.8mm}
  \caption{Quantitative evaluation of whole-body pose generation.}
  \label{tab:wholebody_generation}
  \vspace{-3mm}
\end{table}

\noindent \textbf{Whole-body pose generation.}\
We evaluate whole-body pose generation in Table \ref{tab:wholebody_generation} for VPoser-X and three DPoser-X variants. See Fig.~\ref{fig:wholebody_generation_paper} for visualization results.
DPoser-X-fused produces more diverse body and hand poses compared to VPoser-X, which prioritizes dataset-consistent realism but has limited diversity. 
The DPoser-X-base model exhibits the highest diversity but deviates from the whole-body data distribution, as indicated by the high FID. 
Benefiting our mixed training strategy, DPoser-X-mixed strike a good balance between learning whole-body actions (\eg, expressive grabbing and talking) and preserving generalization on more diverse data sources. For conciseness, in subsequent experiments, we denote our DPoser-X-mixed model simply as DPoser-X when no confusion arises.

\noindent \textbf{Whole-body pose completion.}\
We conduct a challenging pose completion experiment where one hand is masked randomly. This task evaluates the ability to model interdependencies between whole-body parts, particularly between two hands. Given the inherent uncertainties, we obtain 10 hypotheses and evaluate them based on their min/mean/std of errors against the GT. APD across multiple solutions is computed to assess the solution diversity.
The testing datasets include ARCTIC~\cite{fan2023arctic} and BEAT2~\cite{liu2024emage}, focusing separately on cooperative manipulations and speech gestures involving two hands. As shown in Table~\ref{tab:wholebody_pose_completion}, DPoser-X achieves the lowest min-MPVPE and reflects task uncertainty well with high APD. 
This indicates that DPoser-X learns the correlation between whole-body parts effectively.

\begin{table}[t]
\centering
\footnotesize
\setlength{\tabcolsep}{3pt}
\begin{tabular}{lcccc}
    \toprule[\heavyrulewidth]
    \makecell[c]{\multirow{2}{*}[-1.0ex]{Methods}} & \multicolumn{2}{c}{ARCTIC~\cite{fan2023arctic}} & \multicolumn{2}{c}{BEAT2~\cite{liu2024emage}} \\
    \cmidrule(r){2-3} \cmidrule(l){4-5} 
    & \makecell[c]{MPVPE $\downarrow$} & \makecell[c]{APD $\uparrow$} & \makecell[c]{MPVPE $\downarrow$} & \makecell[c]{APD $\uparrow$} \\
    \midrule[\lightrulewidth]
    VPoser-X~\cite{pavlakos2019expressive} & 37.34/43.24/4.60 & 0.59 & 27.49/35.46/5.06 & 0.66 \\
    \rowcolor{colorTab}
    DPoser-X & \textbf{21.81}/30.99/6.10 & \textbf{1.24} & \textbf{15.92}/25.89/6.04 & \textbf{1.18} \\
    \bottomrule[\heavyrulewidth]
\end{tabular}
\vspace{-2mm}
\caption{Performance metrics (min/mean/std of MPVPE and APD) for whole-body pose completion on multiple datasets.}
\label{tab:wholebody_pose_completion}
\vspace{-3mm}
\end{table}

\begin{table}[t]
  \centering
  \footnotesize
  \resizebox{\linewidth}{!}{
  \begin{tabular}{lcccc}
    \toprule[\heavyrulewidth]
    \makecell[c]{\multirow{2}{*}[-1.0ex]{Methods}} & \multicolumn{3}{c}{PA-MPVPE$\downarrow$} & {PA-MPJPE$\downarrow$} \\
    \cmidrule(r){2-4} \cmidrule(l){5-5}
    & \makecell[c]{All} & \makecell[c]{Hands} & \makecell[c]{Face} & \makecell[c]{Body} \\
    \midrule[\lightrulewidth]
    w/o prior & 72.94 & 18.80 & 11.60 & 86.40 \\
    GMM~\cite{bogo2016keep} \& L2 prior & 67.25 & 17.93 & 10.92 & 79.08 \\
    VPoser-X~\cite{pavlakos2019expressive} & 66.74 & 17.44 & 10.99 & 79.88 \\
    \rowcolor{colorTab}
    DPoser-X & \textbf{60.98} & \textbf{15.60} & \textbf{9.75} & \textbf{73.00} \\
    \midrule[\lightrulewidth]
    SMPLer-X~\cite{cai2024smpler} & 26.15 & 11.21 & 2.95 & 29.31 \\
    + w/o prior & 26.33 & 10.34 & 2.86 & 28.69 \\
    + GMM~\cite{bogo2016keep} \& L2 prior & 25.60 & 9.99 & 2.78 & 28.12 \\
    + VPoser-X~\cite{pavlakos2019expressive} & 25.41 & 9.50 & 2.83 & 28.37 \\
    \rowcolor{colorTab}
    + DPoser-X & \textbf{24.65} & \textbf{7.33} & \textbf{2.73} & \textbf{27.87} \\
    \bottomrule[\heavyrulewidth]
  \end{tabular}
  }
  \vspace{-2mm}
  \caption{Whole-body mesh recovery results on ARCTIC~\cite{fan2023arctic}.}
  \label{tab:HMR_wholebody_arctic}
  \vspace{-5.5mm}
\end{table}

\begin{figure*}[t]
    \centering
    \includegraphics[width=0.98\linewidth]{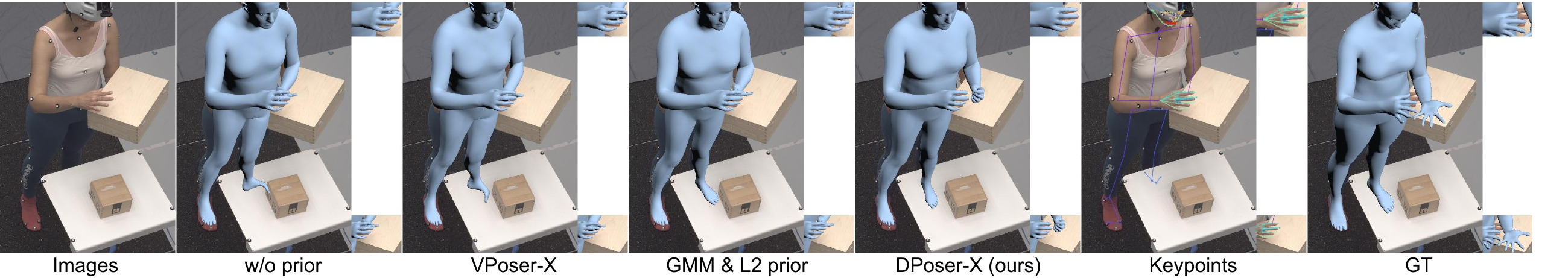}
    \vspace{-2.8mm}
    \caption{Visualization of whole-body mesh recovery on the ARCTIC dataset~\cite{fan2023arctic}. DPoser-X can recover plausible whole-body poses from imperfect 2D keypoints detected by RTMPose~\cite{jiang2023rtmpose}, while other methods struggle to handle noisy inputs effectively.}
    \label{fig:arctic_comparison}
    \vspace{-5.3mm}
\end{figure*}

\noindent \textbf{Whole-body mesh recovery.}\
We evaluate DPoser-X on the ARCTIC dataset~\cite{fan2023arctic} for whole-body mesh recovery. For fitting from scratch, 2D whole-body keypoints are detected by RTMPose~\cite{jiang2023rtmpose}, while for SMPLer-X~\cite{cai2024smpler} initialization, GT 2D keypoints are used. For comparison, we choose VPoser-X and a pose prior baseline that employs GMM~\cite{bogo2016keep} for body poses and L2 prior for hands and face.

Results in Table~\ref{tab:HMR_wholebody_arctic} show that DPoser-X outperforms VPoser-X~\cite{pavlakos2019expressive} and GMM baselines across all metrics for the entire body. 
When used alongside SMPLer-X, DPoser-X further refines the initialization, especially on hands, an area often overlooked by the SOTA whole-body mesh recovery models.
Qualitative results in Fig.~\ref{fig:arctic_comparison} demonstrate that DPoser-X can recover plausible poses from noisy keypoints observations, unlike VPoser-X, which struggles with occlusion and produces unrealistic poses.

\begin{table}[t]
\centering
\resizebox{\linewidth}{!}{
\begin{tabular}{lcccc}
\toprule[\heavyrulewidth]
\makecell[c]{\multirow{2}{*}[-1.0ex]{Scheduling}}& \multicolumn{2}{c}{Whole-body Mesh Recovery} & \multicolumn{2}{c}{Motion Denoising} \\
    \cmidrule(r){2-3} \cmidrule(l){4-5}
    & \makecell[c]{$\text{PA-MPVPE}_{\text{all}}$ $\downarrow$} & \makecell[c]{$\text{PA-MPVPE}_{\text{hands}}$ $\downarrow$} & \makecell[c]{MPVPE $\downarrow$} & \makecell[c]{MPJPE $\downarrow$} \\
    \midrule[\lightrulewidth]
    Random~\cite{poole2022dreamfusion} & 62.28 & 16.63 & 43.33 & 23.87 \\
    Fixed~\cite{cho2023generative} & 61.69 & 15.71 & 45.69 & 22.54 \\
    Uniform~\cite{chung2022diffusion, mardani2023variational} & 62.13 & 17.32 & 39.72 & 20.80 \\
    \rowcolor{colorTab}
    Truncated (ours) & \textbf{60.98} & \textbf{15.60} & \textbf{38.21} & \textbf{19.87} \\
    \bottomrule[\heavyrulewidth]
\end{tabular}
}
\vspace{-2mm}
\caption{Ablation of timestep scheduling on key pose-related tasks}
\vspace{-2mm}
\label{tab:scheduling}
\end{table}

\begin{table}[t]
\centering
\footnotesize
\vspace{-1mm}
\setlength{\tabcolsep}{3pt} 
\begin{tabular}{lcccc}
    \toprule[\heavyrulewidth]
    \makecell[c]{\multirow{2}{*}[-1.0ex]{Methods}} & \multicolumn{2}{c}{ARCTIC~\cite{fan2023arctic}} & \multicolumn{2}{c}{BEAT2~\cite{liu2024emage}} \\
    \cmidrule(r){2-3} \cmidrule(l){4-5} 
    & \makecell[c]{MPVPE $\downarrow$} & \makecell[c]{APD $\uparrow$} & \makecell[c]{MPVPE $\downarrow$} & \makecell[c]{APD $\uparrow$}\\
    \midrule[\lightrulewidth]
    DPoser-X-base & 25.49/36.94/8.13 & 1.43 & 21.98/33.41/8.20 & 1.37 \\
    DPoser-X-fused & 21.51/30.37/5.96 & 1.14 & 15.51/24.58/6.25 & 1.12 \\
    \rowcolor{colorTab}
    DPoser-X-mixed & 21.81/30.99/6.10 & 1.24 & 15.92/25.89/6.04 & 1.18 \\
    \bottomrule[\heavyrulewidth]
\end{tabular} 
\vspace{-2mm}
\caption{Ablation of training strategies for whole-body pose completion. min/mean/std of MPVPE and APD are reported.}
\label{tab:wholebody_pose_completion_ablation}
\vspace{-2mm}
\end{table}

\begin{table}[t]
  \centering
  \footnotesize
  \vspace{-1mm}
  \begin{tabular}{lcccc}
    \toprule[\heavyrulewidth]
    \makecell[c]{\multirow{2}{*}[-1.0ex]{Methods}} & \multicolumn{3}{c}{PA-MPVPE$\downarrow$} & {PA-MPJPE$\downarrow$} \\
    \cmidrule(r){2-4} \cmidrule(l){5-5}
    & \makecell[c]{All} & \makecell[c]{Hands} & \makecell[c]{Face} & \makecell[c]{Body} \\
    \midrule[\lightrulewidth]
    DPoser-X-base & 72.79 & 17.21 & 5.41 & 74.68 \\
    DPoser-X-fused & 72.06 & 18.12 & 5.35 & 75.27 \\
    \rowcolor{colorTab}
    DPoser-X-mixed & \textbf{70.91} & \textbf{15.83} & \textbf{5.27} & \textbf{74.33} \\
    \bottomrule[\heavyrulewidth]
  \end{tabular}
  \vspace{-1.5mm}
  \caption{Ablation of training strategies for whole-body mesh recovery on the Fit3D dataset~\cite{fieraru2021aifit}.}
  \label{tab:HMR_wholebody_arctic_ablation}
  \vspace{-5mm}
\end{table}

\vspace{-1.5mm}
\subsection{Ablation Study}
\vspace{-1mm}
\label{sec:ablation}
We conduct ablation studies to evaluate the effectiveness of the proposed truncated timestep scheduling and mixed training strategy. Experiments on data representations, network settings, and comparisons with other diffusion-based inverse problem solvers are available in Section~\ref{sec:training} and \ref{sec:testing}.

\noindent \textbf{Effectiveness of truncated timestep scheduling.}\
We contrast our proposed scheduling strategy against three established methods—random~\cite{poole2022dreamfusion}, fixed~\cite{cho2023generative}, and uniform~\cite{chung2022diffusion, mardani2023variational} scheduling.
The results in Table~\ref{tab:scheduling} demonstrate that our scheduling outperforms existing strategies on all the evaluated tasks.
As outlined in Section~\ref{sec:truncated_timestep}, the timestep range should be selected based on the noise characteristic of each task. This actually provides a perspective to explain the performance.
The uniform scheduling (\ie, range as \([1.0, 0.0]\)) performs poorly on mesh recovery due to the low noise scale of this task. 
Meanwhile, the fixed scheduling (\ie, \(t_{max} = t_{min}\)) yields the worst results for motion denoising, since the poses are denoised gradually during optimization and the timestep \(t\) should decrease to adapt it.

\noindent \textbf{Advantage of mixed training strategy.}\
The mixed training strategy for DPoser-X combines part-only and whole-body datasets, allowing whole-body modeling and effectively preserving generalization.
Beyond the generation experiments in Table~\ref{tab:wholebody_generation}, we evaluate DPoser-X variants on more downstream tasks.
In whole-body pose completion (Table \ref{tab:wholebody_pose_completion_ablation}), DPoser-X-mixed delivers comparable accuracy to DPoser-X-fused in MPVPE, showing the ability to learn correlations between whole-body parts.
In contrast, DPoser-X-base struggles to model such interactions, evidenced by much higher MPVPE. 
Furthermore, we conduct whole-body mesh recovery experiments on Fit3D~\cite{fieraru2021aifit}, which contains challenging sports poses, without additional training.
As shown in Table~\ref{tab:HMR_wholebody_arctic_ablation}, DPoser-X-mixed achieves the best overall PA-MPVPE and outperforms both DPoser-X-fused and DPoser-X-base, especially in hand-related metrics.
These results highlight the strength of the mixed training strategy, enabling DPoser-X-mixed to generalize well while maintaining whole-body modeling ability.

\vspace{-3mm}
\section{Conclusion}
\vspace{-1.5mm}
We present DPoser, an unconditional diffusion-based pose prior model designed to support a wide range of pose-related tasks. DPoser is engineered for versatility, functioning as a simple L2-loss regularizer, and is further enhanced by our novel truncated timestep scheduling for test-time optimization. Unlike prior methods focused solely on the human body, DPoser models are developed for body, hand, and face, demonstrating their efficiency and robustness across various downstream tasks.
Additionally, we introduce a mixed training strategy to construct the whole-body model, DPoser-X, which effectively integrates both whole-body and part-only datasets. This approach enables DPoser-X to capture correlations between whole-body parts while maintaining strong generalization.
Comprehensive experiments substantiate DPoser-X's superior performance over existing state-of-the-art pose priors, highlighting its potential for broad application.

\section{Acknowledgments}
This research was supported by the National Key Research and Development Program of China (Project No. 2022YFB36066) and the Shenzhen Science and Technology Project (Grant Nos. KJZD20240903103210014, JCYJ20220818101004). This work was also supported by the Ministry of Education, Singapore, under its MOE AcRF Tier 2 (MOE-T2EP20221-0012, MOE-T2EP20223-0002), and under the RIE2020 Industry Alignment Fund – Industry Collaboration Projects (IAF-ICP) Funding Initiative, as well as cash and in-kind contribution from the industry partner(s).

{
\small
\bibliographystyle{ieeenat_fullname}
\bibliography{references}
}

\vfill
\newpage

\setcounter{section}{0}
\setcounter{table}{0}
\setcounter{figure}{0}

\renewcommand{\thesection}{\Alph{section}}   
\renewcommand {\thetable} {S-\arabic{table}}
\renewcommand {\thefigure} {S-\arabic{figure}}


\twocolumn[{
	\renewcommand\twocolumn[1][]{#1}
	\begin{center}
		\textbf{\Large Supplementary Material:\\DPoser-X: Diffusion Model as Robust 3D Whole-body Human Pose Prior}
        \vspace{0.8cm}
        \end{center}
}]

This appendix provides comprehensive details to supplement our main work. Section~\ref{sec:related_works} presents an in-depth discussion of related studies. The parameterization of diffusion models and their connection to score functions are recapped in Section~\ref{sec:detail_intro}, followed by the perspective of Score Distillation Sampling (SDS) to understand our DPoser regularization in Section~\ref{sec:sds}.

Section~\ref{sec:run_time} examines the runtime and computational overhead introduced by DPoser, while Section~\ref{sec:timestep_analysis} explores test-time timestep scheduling across both pose and image domains. The datasets used for training and evaluation are detailed in Section~\ref{sec:datasets}. Section~\ref{sec:exp} covers detailed experimental setup including task-specific loss, evaluation metrics used in each task and implementation of comparative methods.

Further evaluations of DPoser-X on additional tasks and datasets are provided in Section~\ref{sec:addtional_experiments}. Section~\ref{sec:training} outlines the training process for DPoser, whereas Section~\ref{sec:testing} discusses extended optimization techniques. Section~\ref{sec:limitation} outlines the method's limitations, presents failure cases, and suggests avenues for future research. Lastly, Section~\ref{sec:visual} showcases additional qualitative results.

\blfootnote{* Equal contribution. $\dagger$ Corresponding authors.}

\section{Related Work}
\label{sec:related_works}
\subsection{Human Pose Priors}
Human body models like SMPL~\cite{loper2015smpl} and SMPL-X~\cite{pavlakos2019expressive} serve as powerful tools for parameterizing both pose and shape, thereby offering a comprehensive framework for describing human gestures. Within the SMPL model, body poses are represented using rotation matrices or joint angles linked to a kinematic skeleton. Adjusting these parameters enables the representation of diverse human actions. Nonetheless, feeding unrealistic poses into these models can result in non-viable human figures, primarily because plausible human poses are confined within a complex, high-dimensional manifold due to biomechanical constraints.

Various strategies~\cite{bogo2016keep, pavlakos2019expressive, tiwari2022pose, davydov2022adversarial, he2024nrdf} have been put forward to build human pose priors. Generative frameworks like GMMs, VAEs~\cite{kingma2013auto}, and Generative Adversarial Networks 
(GANs)~\cite{creswell2018generative} have shown promise in encapsulating the multifaceted pose distribution, facilitating advancements in tasks like human mesh recovery~\cite{kanazawa2018end, georgakis2020hierarchical}. 
Further, some studies have delved into conditional pose priors tailored to specific tasks, incorporating extra information such as image features~\cite{qiu2023learning, cho2023generative}, 2D joint coordinates~\cite{ci2023gfpose}, or sequences of preceding poses~\cite{ling2020character, rempe2021humor}.
Our initiative leans towards an unconditional pose prior without relying on additional inputs like images or text, aiming for a versatile application across various pose-related scenarios.

\subsection{Diffusion Models for Pose-centric Tasks}
Diffusion models~\cite{song2019generative, song2020score, ho2020denoising, song2020denoising} have emerged as powerful tools for capturing intricate data distributions, aligning well with the demands of multi-hypothesis estimation in ambiguous human poses.
Notable works include DiffPose~\cite{holmquist2023diffpose}, which employs a Graph Convolutional Network (GCN) architecture conditioned on 2D pose sequences for 3D pose estimation by learned reverse process (\ie, generation).
Similarly, DiffusionPose~\cite{qiu2023learning} and GFPose~\cite{ci2023gfpose} employ the generation-based pipeline but take different approaches in conditioning. 
Further, ZeDO~\cite{jiang2023back} concentrates on 2D-to-3D pose lifting, while Diff-HMR~\cite{cho2023generative} and DiffHand~\cite{li2023diffhand} explore estimating SMPL parameters and hand mesh vertices, respectively.
EgoWholeBody~\cite{wang2024egocentric} and RoHM~\cite{zhang2024rohm} focus on refining noisy motion sequences via diffusion-based generation.
BUDDI~\cite{muller2023generative} stands out for using diffusion models to capture the joint distribution of interacting individuals and leveraging SDS loss~\cite{poole2022dreamfusion, wang2023score} for optimization during testing phases.

While DPoser shares similar optimization implementation with BUDDI, it sets itself apart by introducing a wider perspective of inverse problems and equipping an innovative timestep scheduling strategy tailored to human poses. 
Unlike other approaches~\cite{jiang2023back, qiu2023learning, ci2023gfpose, holmquist2023diffpose} that primarily focus on 3D location-based representation, DPoser takes on the more demanding task of modeling SMPL-based rotation pose representation. 
Furthermore, DPoser-X improves whole-body modeling with detailed hand and facial expressions, making it a versatile choice for pose-centric tasks.

\section{Parameterization of Score-based Diffusion Models}
\label{sec:detail_intro}
In the seminal work by Song \etal~\cite{song2020score}, it is demonstrated that both score-based generative models~\cite{song2019generative} and diffusion probabilistic models~\cite{ho2020denoising} can be interpreted as discretized versions of stochastic differential equations (SDEs) defined by score functions. This unification allows the training objective to be interpreted either as learning a time-dependent denoiser or as learning a sequence of score functions that describe increasingly noisy versions of the data.

We begin by revisiting the training objective for score-based models~\cite{song2019generative} to elucidate the link with diffusion models~\cite{ho2020denoising}. Consider the transition kernel of the forward diffusion process \(p_{0t}(\mathbf{x}_t | \mathbf{x}_0)=\mathcal{N}(\mathbf{x}_t;\alpha_{t}\mathbf{x}_0,\sigma_{t}^2\mathbf{I})\). Our goal is to learn score functions \( \nabla_{\mathbf{x}_t} \log p_t(\mathbf{x}_t) \) through a neural network \( s_\theta(\mathbf{x}_t; t) \), by minimizing the L2 loss as follows (we omit the expectation operator for conciseness) :
\begin{equation}
    \mathbb{E} \left[w(t)||s_\theta(\mathbf{x}_t;t)-\nabla_{\mathbf{x}_t}\log p_t\left(\mathbf{x}_t\right)||_2^2\right].
    \label{eq:ESM}
\end{equation}
Here, \(\mathbf{x}_t = \alpha_{t}\mathbf{x}_0+\sigma_{t}\epsilon\), where \(\epsilon \sim \mathcal{N}(\mathbf{0}, \mathbf{I})\).

Based on denoising score matching~\cite{vincent2011connection}, we know the minimizing objective Eq.~\eqref{eq:ESM} is equivalent to the following tractable term:
\begin{equation}
    \mathbb{E} \left[w(t)||s_\theta(\mathbf{x}_t;t)-\nabla_{\mathbf{x}_t}\log p_{0t}(\mathbf{x}_t | \mathbf{x}_0)||_2^2\right].
    \label{eq:DSM}
\end{equation}
To link this with the noise predictor \( \epsilon_\theta(\mathbf{x}_t; t) \) in diffusion models, we can employ the reparameterization \( s_\theta(\mathbf{x}_t; t) = - \frac{\epsilon_\theta(\mathbf{x}_t; t)}{\sigma_t} \). Then, Eq.~\eqref{eq:DSM} can be simplified as follows:
\begin{align}
&w(t)||- \frac{\epsilon_\theta(\mathbf{x}_t;t)}{\sigma_t} - \nabla_{\mathbf{x}_t}\log p_{0t}(\mathbf{x}_t \mid \mathbf{x}_0)||_2^2 \notag\\
=&w(t)||- \frac{\epsilon_\theta(\mathbf{x}_t;t)}{\sigma_t} + \frac{(\mathbf{x}_t-\alpha _t\mathbf{x}_0)}{\sigma_{t}^2} ||_2^2 \notag\\
=&w(t)||- \frac{\epsilon_\theta(\mathbf{x}_t;t)}{\sigma_t} + \frac{\sigma_{t}\epsilon}{\sigma_{t}^2})||_2^2 \notag\\
=&\frac{w(t)}{\sigma_{t}^2} ||\epsilon_\theta(\mathbf{x}_t;t) - \epsilon)||_2^2 \label{eq:DSM2}
\end{align}

The resulting form of Eq.~\eqref{eq:DSM2} aligns precisely with the noise prediction form of diffusion models~\cite{ho2020denoising} (refer to Eq.~(4) in the main text). This implies that by training \( \epsilon_\theta(\mathbf{x}_t; t) \) in a diffusion model context, we simultaneously get a handle on the score function, approximated as \( \nabla_{\mathbf{x}_t} \log p_t(\mathbf{x}_t) \approx - \frac{\epsilon_\theta(\mathbf{x}_t; t)}{\sigma_t} \).

\begin{table*}[t]
  \centering
    \begin{tabular}{lcccccc}
    \toprule[\heavyrulewidth]
    \makecell[c]{\multirow{2}{*}[-1.0ex]{Strategy}}& \multicolumn{2}{c}{Whole-body Mesh Recovery} & \multicolumn{2}{c}{Body Pose Completion} & \multicolumn{2}{c}{Motion Denoising} \\
        \cmidrule(r){2-3} \cmidrule(lr){4-5} \cmidrule(l){6-7}
        & \makecell[c]{PA-MPVPE (all) $\downarrow$} & \makecell[c]{PA-MPVPE (hands) $\downarrow$} & \makecell[c]{MPVPE $\downarrow$} & \makecell[c]{APD $\uparrow$} & \makecell[c]{MPVPE $\downarrow$} & \makecell[c]{MPJPE $\downarrow$} \\
        \midrule[\lightrulewidth]
        \rowcolor{colorTab}
        1 step & \textbf{60.98} & \textbf{15.60} & \textbf{38.79}/78.31/27.13 & 6.53 & \textbf{38.21} & \textbf{19.87} \\
        5 step & 61.39 & 15.70 & 40.15/85.01/31.96 & 7.72 & 40.22 & 21.21 \\
        10 step & 61.52 & 15.74 & 41.04/87.36/32.51 & \textbf{8.07} & 40.69 & 21.34 \\
        \bottomrule[\heavyrulewidth]
    \end{tabular}
  \caption{Ablation of different denoising steps in DPoser's optimization.}
  \label{tab:denoiser steps}
\end{table*}

\section{View DPoser as Score Distillation Sampling}
\label{sec:sds}

Interestingly, the gradient of DPoser (Eq.~(10) in the main text) coincides with Score Distillation Sampling (SDS)~\cite{poole2022dreamfusion, wang2023score}, which can be interpreted as aiming to minimize the following KL divergence:
\begin{equation}
    KL\big(p_{0t}\left(\mathbf{x}_t\mid \mathbf{x}_0\right) \parallel p_t^\mathtt{SDE}\left(\mathbf{x}_t;\theta \right)\big),
    \label{eq:SDS KL}
\end{equation}
where \(p_t^\mathtt{SDE}\left(\mathbf{x}_t;\theta\right)\) denote the marginal distribution whose score function is estimated by \(\epsilon_\theta(\mathbf{x}_t;t)\).
For the specific case where \(t \to 0\), this term encourages the Dirac distribution \(\delta (\mathbf{x}_0)\) (\ie, the optimized variable) to gravitate toward the learned data distribution \(p_0^\mathtt{SDE}\left(\mathbf{x}_0;\theta\right)\), while the Gaussian perturbation like Eq.~\eqref{eq:SDS KL} softens the constraint.
Building on this understanding, we can borrow advanced techniques from SDS—a rapidly evolving area ripe for methodological innovations~\cite{daras2024survey}. To extend this, we experiment with a multi-step denoising strategy adapted from HiFA~\cite{zhu2023hifa}, substituting our original one-step denoising process. This alternative, however, yields suboptimal results across most evaluation metrics, as demonstrated in Table~\ref{tab:denoiser steps}. A plausible explanation could be that our proposed truncated timestep scheduling effectively manages low noise levels (\ie, small \(t\)), thus negating the need for more denoising steps. 
In our main experiments, we keep the efficient one-step denoiser.

\begin{table}[t]
    \small
    \setlength{\tabcolsep}{2pt}
    \begin{center}
    \resizebox{0.47\textwidth}{!}{\noindent
    \begin{tabular}{cccccc}
        \toprule[\heavyrulewidth]
        w/o prior & GMM~\cite{bogo2016keep} & VPoser~\cite{pavlakos2019expressive} & Pose-NDF~\cite{tiwari2022pose} & GAN-S~\cite{davydov2022adversarial} & DPoser \\
        \midrule[\lightrulewidth]
        15.64 & 16.14 & 16.83 & 21.88 & 74.60 & 17.34 \\
      \bottomrule[\heavyrulewidth]
      \end{tabular}
      }
    \vspace{-2.5mm}
    \caption{Runtime comparison (in seconds) of different prior models for human mesh recovery on 100 images, evaluated using an RTX 3090Ti GPU.}
    \label{tab:run_time}
    \end{center}
\end{table}

\section{Runtime Comparison}
\label{sec:run_time}
Diffusion models generally require iterative steps for gradual denoising, making them less efficient than VAEs and GANs in generation tasks. However, when applied to downstream optimization processes, DPoser introduces minimal additional computational overhead. This is due to two key factors: (1) DPoser regularization involves only a single-step denoising at each optimization step, and (2) the stop-gradient operator ensures that the regularization does not require backpropagation through the trained network. 

To assess DPoser's efficiency, we benchmarked its runtime against various prior models (including a baseline without pose prior) for human mesh recovery across 100 images in a consistent execution environment. As shown in Table~\ref{tab:run_time}, incorporating DPoser results in only a modest (10\%) increase in optimization runtime compared to the baseline. In contrast, GAN-S~\cite{davydov2022adversarial} incurs a significant computational cost due to its required GAN-inversion phase, which converts initial poses into their latent representations.

\section{Analysis of Test-time Timestep Scheduling}
\label{sec:timestep_analysis}

\begin{figure}[t]
    \includegraphics[width=0.49\textwidth]{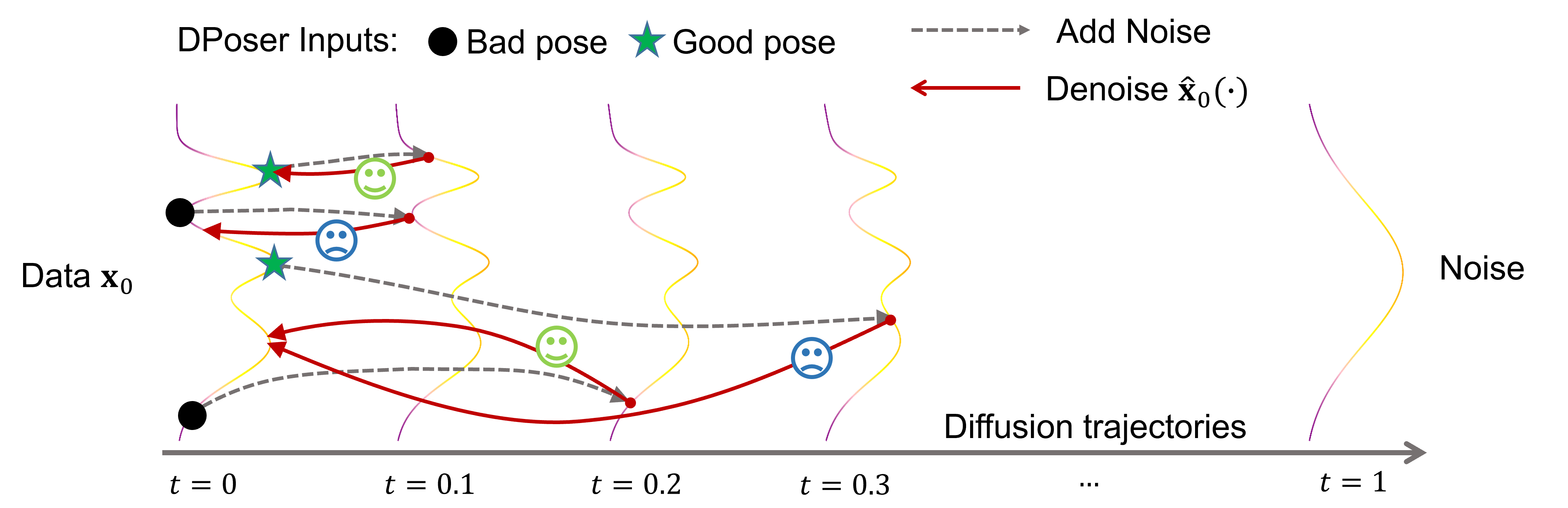}
    \vspace{-3mm}
    \caption{Visualization of the impact of different timestep values in DPoser regularization. A larger \(t\) effectively corrects undesirable poses but may excessively alter well-posed inputs, resulting in plausible yet unrelated poses. Conversely, a smaller \(t\) better preserves the original pose but struggles to correct implausible ones.}
    \label{fig:traj}
\end{figure}

\begin{table}[t]
    \centering
    \vspace{-1mm}
    \resizebox{0.48\textwidth}{!}{\noindent
      \begin{tabular}{lcccc}
        \toprule[\heavyrulewidth]
        Noise std & \([0.15, 0.05]\) & \([0.2, 0.05]\) & \([0.2, 0.1]\)& \([0.25, 0.1]\) \\
        \midrule[\lightrulewidth]
        40 mm & \textbf{19.83} & \underline{19.87} & 21.68 &  22.14\\
        100 mm & 36.13 & 34.15 & \textbf{33.18} & \underline{33.83} \\
      \bottomrule[\heavyrulewidth]
      \end{tabular}
    }
    \caption{Ablation of timestep range for motion denoising on the AMASS dataset~\cite{mahmood2019amass}. MPJPE is reported as the metric.}
    \label{tab:interval_ablation}
    \vspace{-2.5mm}
\end{table}

During optimization, the selection of timestep is crucial for downstream tasks. 
As discussed in Section~2.4, the key information of pose data emerges at small \(t\) values (\(t\le0.3\)), which serves as a coarse range.
Moreover, the L2 loss format of our DPoser regularization gives an intuitive view of the impact of timestep. 
As shown in Fig.~\ref{fig:traj}, while \(t\) is small, since the adding noise and denoising path is short, the denoised pose is close to the origin and the DPoser guidance is weak. 
Specifically, considering the extreme case where \(t \to 0\), in \(\mathbf{\hat{x}}_0(t) = \frac{\mathbf{x}_t - \sigma_{t}\epsilon_\phi(\mathbf{x}_t;t)}{\alpha_{t}}\), the coefficient \(\sigma_{t} \to 0\) while \(\alpha_{t} \to 1\), causing \(\mathbf{\hat{x}}_0(t)\) to approach \(\mathbf{x}_0\), which leads to a near-zero DPoser loss.
On the contrary, suitably large \(t\) means strong DPoser guidance and can correct implausible poses better. 
Thus, we tailor \([t_{max}, t_{min}]\) intervals to specific tasks based on their noise scales. To verify this, we conduct ablation of the timestep range on motion denoising. As evidenced in Table~\ref{tab:interval_ablation}, to achieve the best performance, larger \(t\) values are required for noisier inputs. Based on the above analyses, we select task-specific timestep intervals \([t_{max}, t_{min}]\) as follows: \([0.2, 0.05]\) for motion denoising (40 mm noise), \([0.15, 0.05]\) for pose completion and inverse kinematics, and \([0.12, 0.08]\) for mesh recovery. All the experiments, including body-only, hand-only, face-only, and whole-body, share the same timestep hyperparameters without more tuning.

\begin{figure*}[t]
    \centering
    \includegraphics[width=\linewidth]{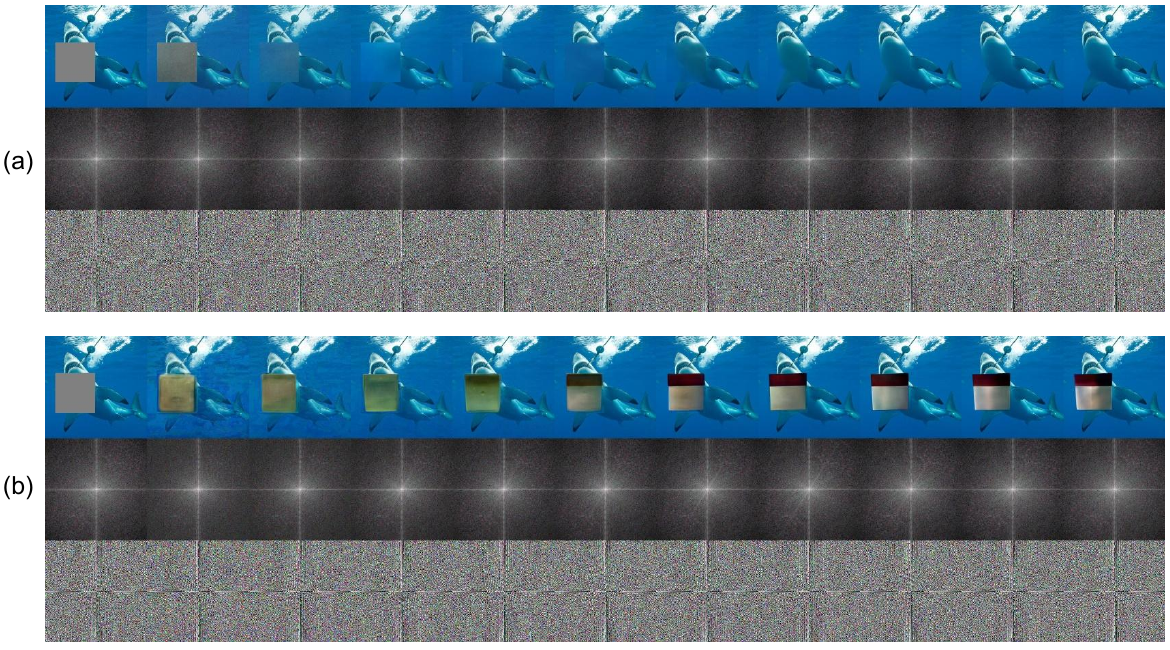}
    \caption{Image inpainting using standard and truncated timestep scheduling. The process evolution is shown over iterations with the middle row depicting the log-magnitude spectrum and the bottom row the phase spectrum. (a) The standard scheduling exhibits cohesive restoration with detail fidelity. (b) The truncated scheduling results in detail-rich patches that are perceptually incongruent with the original image context.}
    \label{fig:truncated_images}
\end{figure*}

It is also noteworthy that our truncated timestep scheduling is designed for human poses and does not work well on images. In image domains, the initial timesteps play a crucial role in generating foundational perceptual content.
In our study, we employed a 256x256 unconditional diffusion model~\cite{dhariwal2021diffusion} trained on ImageNet~\cite{deng2009imagenet} with variational diffusion sampling~\cite{mardani2023variational} for image inpainting. This model employs 1000 discrete timesteps during training.
We compared standard scheduling (timesteps 990 to 0) with truncated scheduling (timesteps 495 to 0), both using 100 steps. The results, shown in Fig.~\ref{fig:truncated_images}, indicate that truncation negatively affects image quality. While the standard approach preserved perceptual content, the truncated method produced disjointed patches that were misaligned with the original context. 
These results affirm that truncated timestep scheduling excels in pose data where key information emerges in later stages but falls short in image tasks where early timesteps are essential. This scheduling is thus bespoke to the characteristics of human poses and is unsuitable for image processes that rely on the full diffusion timeline for content fidelity.

\section{Dataset Description}
\label{sec:datasets}
This section provides a detailed overview of the datasets used in our experiments, categorized based on the body part they focus on. We describe each dataset's specific use case along with the number of samples available for each dataset.

\subsection{Body-only Dataset}
\noindent \textbf{AMASS}\
The AMASS dataset~\cite{mahmood2019amass} is a large-scale collection of high-quality 3D human body meshes derived from multiple motion capture sources. It provides motion sequences and human poses in a SMPL-based format, covering a broad range of activities such as walking, sitting, dancing, and running. Following the same splits as VPoser~\cite{pavlakos2019expressive}, and after sampling to de-duplicate the data, we use approximately 55 million body poses in the SMPL-X~\cite{pavlakos2019expressive} format to train our DPoser-body model. The test split consists of 54,000 body poses, which are used to evaluate model performance on tasks including body pose completion and motion denoising.

\noindent \textbf{HPS}\
The HPS dataset~\cite{guzov2021human} contains over 300K synchronized RGB images, paired with reference 3D poses and locations, captured from seven people interacting with large-scale 3D scenes. The dataset includes motion sequences of various activities such as exercising, reading, eating, lecturing, using a computer, making coffee, and dancing.
Following Pose-NDF~\cite{tiwari2022pose}, we use the HPS dataset to evaluate the motion denoising task without training on it. After sampling, we got 350 sequences, each consisting of 60 frames for testing.

\subsection{Hand-only Dataset}
\noindent \textbf{FreiHAND}\
FreiHAND~\cite{zimmermann2019freihand} is a large-scale dataset for 3D hand pose estimation, focusing on single-hand poses. It includes 130,240 training samples (4\(\times\)32560) and 3,960 evaluation samples. Each training hand pose is accompanied by 4 RGB images, providing diverse data for training robust models. We use 32,560 hand poses, represented in the MANO format~\cite{romero2022embodied}, for training the DPoser-hand model. The remaining 3,960 evaluation samples are used to assess hand mesh recovery performance during testing.

\noindent \textbf{DexYCB}\
DexYCB~\cite{chao2021dexycb} is a dataset for capturing 3D hand poses during hand-object interactions, focusing on single-hand poses. We use only the hand poses for training the DPoser-hand model. The training set includes 407,000 single hand poses.

\noindent \textbf{HO3D}\
HO3D~\cite{hampali2020honnotate} is a dataset that provides 3D annotations for both hand poses and object interactions. Similar to DexYCB, we utilize the hand poses for training the DPoser-hand model. The training set contains 83,000 hand poses.

\noindent \textbf{H2O}\
The H2O dataset~\cite{kwon2021h2o} provides 3D pose annotations for two-hand and object interactions. For the purpose of training DPoser-hand, we use only the right-hand poses. The training set contains 58,000 hand poses.

\noindent \textbf{ReInterHand}\
ReInterHand~\cite{moon2024dataset} is a high-quality synthetic dataset designed for 3D hand pose estimation, specifically focusing on interacting hands. It includes annotations for both hands. For training DPoser-hand, we flip the left-hand pose as right-hand to unify the format. The dataset is split into training, validation, and test sets with an 8:1:1 ratio. We use approximately 186,000 hand poses for training, and 23,000 poses for testing. The test set is used to evaluate hand inverse kinematics tasks.

\subsection{Face-only Dataset}
\noindent \textbf{MICA}\
The MICA dataset~\cite{zielonka2022towards} consists of eight smaller datasets that were unified to represent about 2315 subjects using the FLAME~\cite{li2017learning} model. It contains only shape geometry. We use the MICA dataset to train the shape component of DPoser-face, focusing on high-quality 3D face shapes.

\noindent \textbf{WCPA}\
WCPA~\cite{kao2022single} is a large-scale dataset focusing on 3D face reconstruction under perspective projection. It contains 200 subjects and 356,640 training instances, with detailed annotations for facial expressions. We use WCPA to train the expression component of DPoser-face, with 1/10 of the dataset reserved for testing. The test set is used to evaluate face reconstruction, considering both shape and expressions.

\noindent \textbf{NOW}\
NOW~\cite{sanyal2019learning} is a widely-used benchmark for face reconstruction. It introduces standard evaluation metrics for assessing the accuracy and robustness of 3D face reconstruction methods, especially under variations in viewing angle, lighting, and occlusions. 
The validation set containing 352 images is employed for our face reconstruction task. 
We focus on the non-metrical evaluation of face shape, as the ground truth (GT) only includes shape. As in previous works such as DECA~\cite{feng2021learning}, expressions are set to zero in the FLAME~\cite{li2017learning} model to obtain a neutral face mesh for final evaluation.

\subsection{Whole-body Dataset}
\noindent \textbf{BEAT2}\
BEAT2~\cite{liu2024emage} is a holistic co-speech dataset that combines the MoShed SMPL-X~\cite{pavlakos2019expressive} body with FLAME head parameters. It refines the modeling of head, neck, and finger movements, providing high-quality 3D motion-captured data. After de-duplication, we have 1.48 million whole-body poses for training DPoser-X, and the test set contains 172,000 whole-body poses used for whole-body pose completion.

\noindent \textbf{GRAB}\
GRAB~\cite{taheri2020grab} is a dataset containing full 3D shape and pose sequences of 10 subjects interacting with 51 everyday objects of varying shapes and sizes. We use this dataset for training DPoser-X, with 391,000 whole-body poses. The data helps train the model for whole-body mesh recovery during grasping actions.

\noindent \textbf{ARCTIC}\
ARCTIC~\cite{fan2023arctic} is a dataset focused on two-hand object manipulation, with 2.1 million video frames paired with 3D hand and object meshes. After de-duplication, we have 77,000 whole-body poses for training DPoser-X. The validation set, which contains 10,000 whole-body poses, is used for testing whole-body mesh recovery and pose completion. Note that the face expressions are not annotated, so we set models' output expressions as zeros for evaluation. Face-related metrics in whole-body mesh recovery are only influenced by the human shape.

\noindent \textbf{EgoBody}\
EgoBody~\cite{zhang2022egobody} is a large-scale dataset that captures 3D human motions during social interactions in 3D scenes. It provides SMPL-X~\cite{pavlakos2019expressive} annotations for 3D whole-body pose, shape, and motion for both the interactee and the camera wearer. The training set contains 38,896 instances of whole-body poses (x2 for each subject), and the test set contains 24,665 instances. The test set is used for the whole-body pose completion task.

\noindent \textbf{Fit3D}\
Fit3D~\cite{fieraru2021aifit} is a dataset with over 3 million images and corresponding 3D human shape and motion capture ground truth data, covering 37 exercises performed by instructors and trainees. We take the subject \textit{s04} which consists of 612 images after sampling, for whole-body mesh recovery tasks to test DPoser-X’s generalization, without training the model on this dataset.

\noindent \textbf{EHF}\
EHF~\cite{pavlakos2019expressive} is a curated dataset comprising 100 images with pseudo whole-body poses. Following Pose-NDF~\cite{tiwari2022pose}, we use this dataset to evaluate body mesh recovery performance, specifically calculating PA-MPJPE for body joints.

\section{Experimental Details}
\label{sec:exp}
In this section, we provide detailed descriptions of the experimental setups and specific loss functions for various tasks. These tasks include pose completion, motion denoising, inverse kinematics, face reconstruction, hand mesh recovery, and whole-body mesh recovery. In addition, we explain the evaluation metrics used in each task and implementation details of comparative methods.

\subsection{Pose Completion}
For partial observations \(\mathbf{y}\), the measurement operator \(\mathcal{A}\) is modeled as a known mask matrix \(M \in \mathbb{R}^{d\times n}\). Based on our optimization framework denoted in Alg.~1, we define the task-specific loss, \(L_\text{comp}\), as follows:
\begin{equation}
    L_\text{comp} = ||M \mathbf{x}_0 - \mathbf{y}||_2^2. 
\end{equation}
Here, \(\mathbf{x}_0\) denotes the complete body pose \(\theta\) we try to recover, where the unseen parts are initialized as random noise. In the following ablated studies, if not specified, the evaluation of the body pose completion is performed using 10 hypotheses on the AMASS dataset~\cite{mahmood2019amass} with left leg occlusion.

\subsection{Motion Denoising (Noisy Input)}
\label{sec:motion_denoising}
Adhering to Pose-NDF settings~\cite{tiwari2022pose}, we aim to refine noisy joint positions \(J_{\text{obs}}^t\) over \(N\) frames to obtain clean poses \(\theta^t\),  initialized from mean poses in SMPL with small noise. We formulate the task-specific loss combining an observation fidelity term \(L_\text{obs}\) and a temporal consistency term \(L_\text{temp}\):
\begin{equation}
    L_\text{obs} = \sum_{t=0}^{N-1}  ||M_J(\theta^t, \beta_0) - J_{\text{obs}}^t||_2^2,
    \label{eq:motion}
\end{equation}
\begin{equation}
    L_\text{temp} = \sum_{t=1}^{N-1}  ||M_J(\theta^{t-1}, \beta_0) - M_J(\theta^t, \beta_0)||_2^2,
\end{equation}
where \(M_J\) denotes the 3D joint positions regressed from SMPL~\cite{loper2015smpl} and \(\beta_0\) is the constant mean shape parameters.

\subsection{Motion Denoising (Partial Input)}
This task focuses on reconstructing clean poses, \(\theta^t\), from partially observed joint positions, \(J_{\text{obs}}^t\), across \(N\) frames, employing a known mask matrix to identify visible joints. The optimization objective mirrors that of motion denoising (Section~\ref{sec:motion_denoising}), but incorporates a mask in Eq.~\eqref{eq:motion} to specifically target visible parts, ensuring that only these segments guide the recovery process.

\subsection{Inverse Kinematics}
Inverse kinematics (IK) aims to estimate clean poses from noisy or partially observed 3D joint positions, similar to the motion denoising task. The key difference in the implementation is that the inputs are single-frame data, meaning the temporal consistency term \(L_\text{temp}\) is not required. 

For inverse kinematics applied to hand poses, we optimize only the hand poses while keeping the hand shape parameters fixed. This simplifies the optimization, focusing solely on pose adjustments. For the face, we optimize both the face expression and shape parameters, as face-related tasks require accurate modeling of both shape and dynamic expressions. In all cases, we employ a similar optimization framework as in the motion denoising task, using only the fidelity loss for observed 3D joints.

\subsection{Face Reconstruction}
Reconstructing human faces using only 2D keypoints is challenging and typically insufficient for high-quality reconstructions. To address this, we utilize the photometric optimization approach described in \cite{photometric_optimization} to fit a textured FLAME model~\cite{li2017learning}. The optimization aims to refine the face shape and expression parameters, as well as adjust the appearance and lighting parameters for the face rendering. 
We use a combination of two key loss functions: the photometric loss (L1-loss between rendered and target images) and the reprojection loss (for 2D face keypoints).

We observe that face shape plays a crucial role in tasks like face reconstruction. To this end, DPoser-face is designed to separately model face shape and expression. Given that these two components (shape and expression) are largely independent, we train the face shape and expression models separately using the WCPA~\cite{kao2022single} and MICA~\cite{zielonka2022towards} datasets, respectively. For face-only tasks such as face reconstruction, DPoser regularization is applied to both the face shape and expression models. It is important to note that only the expression component of DPoser-face contributes to the broader DPoser-X framework, with the shape component being reserved for face-specific tasks. For a fair comparison, we implement the same strategy for training the VPoser-face model.

\subsection{Hand Mesh Recovery}
For hand mesh recovery, we optimize the hand poses using the MANO model~\cite{romero2022embodied} instead of the SMPL model. Similar to the body mesh recovery task, we employ a reprojection loss based on 2D hand keypoints. 
In addition to using our DPoser loss for plausible hand poses, we also employ the L2 prior for hand shape, similar to Eq.~(11) in the main text, to maintain natural hand geometry.

\subsection{Whole-body Mesh Recovery}
Whole-body mesh recovery shares similarities with body mesh recovery (as discussed in Section~2.5) but additionally incorporates the face and hands into the optimization. The goal is to recover the whole-body poses \(\theta\) (including body, hands, and face) and shape parameters \(\beta\) by optimizing a reprojection loss based on whole-body 2D keypoints. A distinguishing feature is the inclusion of two root-relative reprojection losses, one for the hands and another for the face, to refine local poses. Specifically, the wrists for hands and the mouth for the face are chosen as the root, and the root coordinates are subtracted before calculating the reprojection losses. This ensures that the hand and face poses are localized relative to the body, improving the accuracy of hand and facial mesh recovery.

\subsection{Evaluation Metrics}
For comprehensive assessment across various tasks, following recent works like NRDF~\cite{he2024nrdf} and SMPLer-X~\cite{cai2024smpler}, we adopt task-specific metrics:

\begin{itemize}
\item \textit{Pose Generation}: Diversity and fidelity are evaluated using Average Pairwise Distance (APD) and \(d_{NN}\)~\cite{he2024nrdf}, respectively. \(d_{NN}\) measures the distance between the generated pose and its nearest neighbor from the training data. We also report the common metrics for generative models, including FID~\cite{heusel2017gans} (distribution similarity), Precision~\cite{kynkaanniemi2019improved} (fidelity), and Recall~\cite{kynkaanniemi2019improved} (diversity).

\item \textit{Human Mesh Recovery}: Procrustes-aligned Mean Per-Vertex Position Error (PA-MPJPE) and Procrustes-aligned Mean Per-Joint Position Error (PA-MPVPE) measures the accuracy of recovered human meshes.

\item \textit{Multi-hypothesis Pose Completion}: MPVPE and APD on masked parts across multiple hypotheses measure solution accuracy and diversity, respectively.

\item \textit{Motion Denoising \& Inverse Kinematics}: Both MPJPE and MPVPE are calculated to assess the performance.
\end{itemize}
All errors are reported in \textit{millimeter} units.

\subsection{Implementation of Comparative Methods}
\label{sec:comparative_implementation}
In pose generation experiments, we employ standard sampling techniques for generative models, including GMM~\cite{bogo2016keep}, VPoser (VAE)~\cite{pavlakos2019expressive}, and GAN-S (GAN)~\cite{davydov2022adversarial}. For Pose-NDF~\cite{tiwari2022pose} and NRDF~\cite{he2024nrdf}, we reproduce their projection algorithms using their official repositories. For other tasks during testing, to ensure a fair comparison, we implement all pose priors within the same optimization framework—using identical task-specific loss functions and optimization iterations—while tuning hyperparameters such as loss weights for each method.

VPoser and GAN-S function as pose priors due to their learned meaningful latent representations. We optimize the pose latents for both methods. Given VPoser's Gaussian assumption, it naturally incorporates L2 regularization on the latent pose~\cite{pavlakos2019expressive}. However, we observe that applying spherical loss to the latents of GAN-S~\cite{davydov2022adversarial} degrades human mesh recovery performance. Therefore, we use only GAN-S’s generator for decoding without imposing additional constraints on the pose latents. 
Both NDF~\cite{tiwari2022pose} and NRDF~\cite{he2024nrdf} directly optimize pose rotation representations by minimizing the predicted distance between the current pose and their learned plausible pose fields. We implement these methods using their official code and model weights. 
Since GAN-S does not provide pre-trained models, we train it from scratch on the same datasets as our DPoser. Additionally, for hand, face, and whole-body models, we train the comparative methods ourselves.

\section{Additional Experiments}
\label{sec:addtional_experiments}
In this section, we present a series of additional experiments that further demonstrate the efficacy of DPoser-X across various tasks. These experiments cover body mesh recovery, body pose completion, motion denoising, hand/face generation, hand/face inverse kinematics, face reconstruction, and whole-body mesh recovery, with a focus on different input types and datasets.

\subsection{Body Mesh recovery}
\label{sec:addtional_hmr}
In addition to the priors compared in the main text, we evaluate DPoser against two recent state-of-the-art, generation-based methods: GFPose~\cite{ci2023gfpose} and HuProSO3~\cite{dunkel2024normalizing}. Unlike optimization-based priors, these methods are designed to produce multiple, diverse hypotheses for a given input.

We report the results for the Human Mesh Recovery (HMR) task on the EHF dataset~\cite{pavlakos2019expressive} in Table~\ref{tab:comparison_gen_methods}. The comparison is conducted with both a single hypothesis (hypotheses\_num=1) and multiple hypotheses (hypotheses\_num=10), reporting the minimum PA-MPJPE and MPJPE. The results clearly show that while GFPose and HuProSO3 can generate diverse potential poses, our DPoser achieves significantly higher accuracy (i.e., lower error) in both evaluation settings. This suggests that DPoser provides a more precise and reliable pose prior for this task.

\begin{table}[t]
    \centering
    \resizebox{0.48\textwidth}{!}{
    \begin{tabular}{lcc}
        \toprule[\heavyrulewidth]
            Methods & ~~~hypotheses\_num=1~~~ & ~~~hypotheses\_num=10~~~ \\
            \midrule[\lightrulewidth]
            GFPose~\cite{ci2023gfpose} & 68.64/89.88 & 62.80/83.39 \\
            HuProSO3~\cite{dunkel2024normalizing} & 72.00/104.52 & 57.42/84.21 \\
            DPoser (ours) &  \textbf{56.05}/\textbf{79.82} & \textbf{53.28}/\textbf{76.53} \\
          \bottomrule[\heavyrulewidth]
          \end{tabular}
        }
    \caption{Comparison with generation-based methods on the HMR task using the EHF dataset~\cite{pavlakos2019expressive}. We report the minimum PA-MPJPE/MPJPE across multiple hypotheses.}
    \label{tab:comparison_gen_methods}
\end{table}

\subsection{Body Pose Completion}

\begin{table*}[t]
\centering
\small
\vspace{-1mm}
\resizebox{\linewidth}{!}{
    \begin{tabular}{lcccccccc}
        \toprule[\heavyrulewidth]
        \makecell[c]{\multirow{2}{*}[-1.0ex]{Methods}} & \multicolumn{2}{c}{Occ. left leg} & \multicolumn{2}{c}{Occ. legs} & \multicolumn{2}{c}{Occ. arms} & \multicolumn{2}{c}{Occ. torso} \\
        \cmidrule(r){2-3} \cmidrule(lr){4-5} \cmidrule(lr){6-7} \cmidrule(l){8-9}
        & \makecell[c]{MPVPE $\downarrow$} & \makecell[c]{APD $\uparrow$} & \makecell[c]{MPVPE $\downarrow$} & \makecell[c]{APD $\uparrow$} & \makecell[c]{MPVPE $\downarrow$} & \makecell[c]{APD $\uparrow$} & \makecell[c]{MPVPE $\downarrow$} & \makecell[c]{APD $\uparrow$} \\
        \midrule[\lightrulewidth]
        Pose-NDF~\cite{tiwari2022pose} (\(S=1\)) & 168.61 & NAN & 169.92 & NAN & 261.11 & NAN & 115.03 & NAN \\
        Pose-NDF (\(S=5\)) & 157.62/168.49/7.94 & 1.95 & 162.30/169.94/5.54 & 1.96 & 254.97/261.01/4.38 & 1.22 & 108.07/114.98/4.98 & 0.93 \\
        Pose-NDF (\(S=10\)) & 154.21/168.45/8.66 & 1.95 & 159.75/169.86/6.12 & 1.97 & 252.90/260.94/4.81 & 1.20 & 105.87/114.97/5.43 & 0.93 \\
        VPoser~\cite{pavlakos2019expressive} (\(S=1\)) & 200.23 & NAN & 221.21 & NAN & 206.83 & NAN & 58.66 & NAN \\
        VPoser (\(S=5\)) & 187.38/200.73/10.52 & 2.38 & 201.70/221.16/14.57 & 5.49 & 191.27/206.55/11.54 & 4.06 & 49.88/58.67/6.71 & 1.59 \\
        VPoser (\(S=10\)) & 182.31/200.51/12.20 & 2.41 & 195.76/221.34/16.40 & 5.44 & 186.55/206.72/12.91 & 4.08 & 47.31/58.71/7.38 & 1.56 \\
        CVPoser\(^\dagger\) (\(S=10\)) & 113.48/128.04/10.36 & 1.91 & 121.00/134.35/10.17 & 2.43 & 153.12/162.82/5.58 & 1.08 & 45.16/51.23/4.32 & 0.57 \\
        \midrule[\lightrulewidth]
        \rowcolor{colorTab}
        DPoser (\(S=1\)) & 78.78 & NAN & 103.12 & NAN & 104.59 & NAN & 44.60 & NAN \\
        \rowcolor{colorTab}
        DPoser (\(S=5\)) & 46.23/78.13/24.96 & \textbf{6.58} & 72.37/102.73/23.05 & 7.72 & 74.32/105.70/24.15 & 5.67 & 27.47/44.63/13.26 & 2.19 \\
        \rowcolor{colorTab}
        DPoser (\(S=10\)) & \textbf{38.79}/78.31/27.13 & 6.53 & \textbf{63.65}/102.46/25.39 & \textbf{7.75} & \textbf{64.72}/104.94/26.44 & \textbf{5.69} & \textbf{22.63}/44.60/14.65 & \textbf{2.21} \\
        \bottomrule[\heavyrulewidth]
    \end{tabular}
    }
\caption{Performance metrics (min/mean/std of MPVPE and APD) for body pose completion on the AMASS dataset~\cite{mahmood2019amass} under varying occlusion scenarios. \(S\) denotes the number of hypotheses. \(^\dagger\) Task-specific baseline trained with partial poses as conditional input.}
\label{tab:completion}
\end{table*}

In practical scenarios, HMR algorithms often grapple with occlusions leading to incomplete 3D pose estimates.
In this context, the task is to recover full 3D poses from partially observed data, initializing the occluded parts with noise. Our DPoser model is employed to refine these initially implausible poses into feasible ones, utilizing an L2 loss on the visible parts to ensure data consistency.
In parallel, we employ a comparable optimization strategy for both Pose-NDF~\cite{tiwari2022pose} and VPoser~\cite{pavlakos2019expressive}. As a task-specific baseline, we adapt the original VPoser model into CVPoser by incorporating conditional inputs within its VAE framework~\cite{kingma2013auto} for end-to-end training and conditional sampling. The completion experiment is conducted on the AMASS dataset~\cite{mahmood2019amass} with occlusion of various body parts.

\begin{figure*}[t]
\centering
\includegraphics[width=0.98\linewidth]{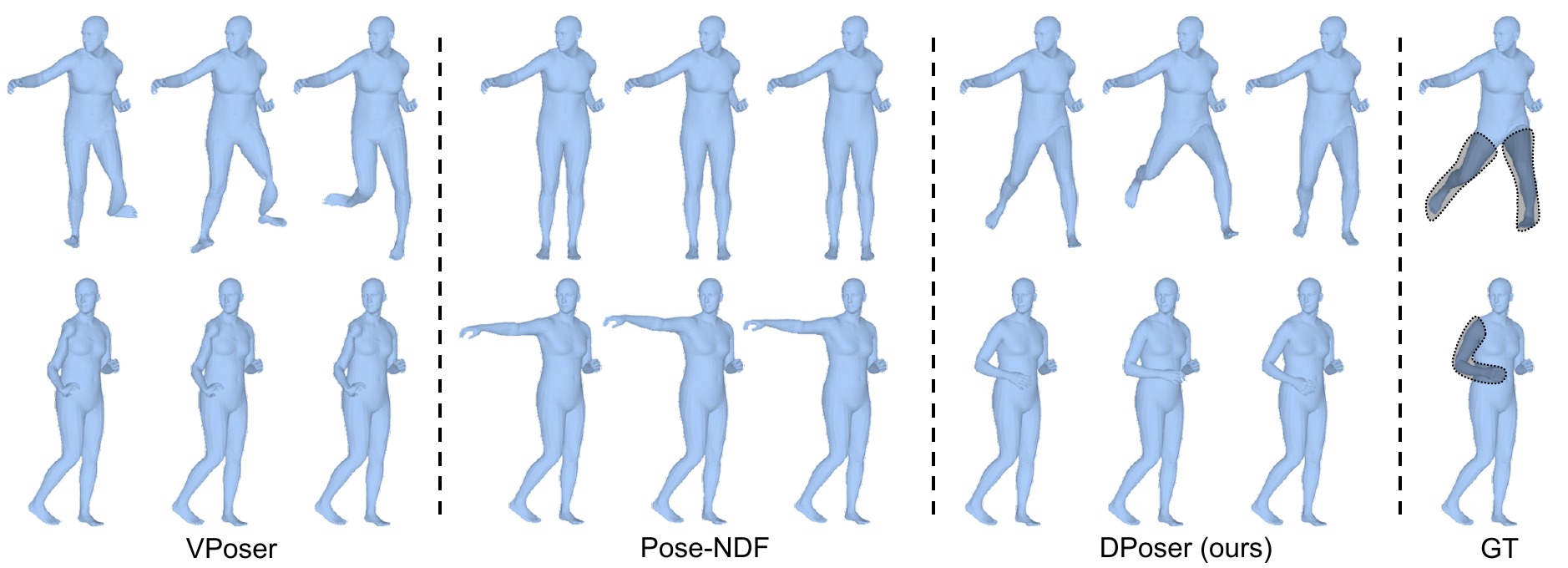}
\vspace{-1mm}
\caption{Visual comparisons of body pose completion. Three hypotheses are drawn for each method. DPoser uniquely offers multiple plausible solutions for partial poses, a scenario where competitors often struggle due to limited generalization. 
}
\label{fig:completion}
\vspace{-3mm}
\end{figure*}

Given the uncertainties in this task, we generate multiple hypotheses and evaluate them using minimum, mean, and standard deviation errors against the ground truth. We calculate APD across solutions to assess diversity.
As illustrated in Table \ref{tab:completion}, DPoser exhibits superior performance across different occlusion scenarios compared to existing pose priors and even the task-specific CVPoser, highlighting its effectiveness in pose completion. 
The qualitative evaluations are presented in Fig. \ref{fig:completion}. Here, we observe that DPoser can generate a multitude of plausible poses, a capability lacking in VPoser~\cite{pavlakos2019expressive}. Pose-NDF~\cite{tiwari2022pose}, meanwhile, struggles with generalizing to unseen noisy poses and making plausible adjustments from the mean pose initialization.

\subsection{Motion Denoising (Noisy Input)}
\begin{table}[t]
\centering
\setlength\tabcolsep{6pt}
\begin{tabular}{lcccc}
    \toprule[\heavyrulewidth]
    \makecell[c]{\multirow{2}{*}[-1.0ex]{Methods}} & \multicolumn{2}{c}{AMASS~\cite{mahmood2019amass}} & \multicolumn{2}{c}{HPS~\cite{guzov2021human}}\\
    \cmidrule(r){2-3} \cmidrule(l){4-5}
    & \makecell[c]{20mm} & \makecell[c]{100mm} & \makecell[c]{20mm} & \makecell[c]{100mm} \\
    \midrule[\lightrulewidth]
    w/o prior & 15.33 & 51.48 & 16.26 & 50.87 \\
    VPoser~\cite{pavlakos2019expressive} & 15.20 & 49.10 & 17.24 & 46.69 \\
    Pose-NDF~\cite{tiwari2022pose} & 13.84 & 46.10 & 15.62 & 47.50  \\
    \rowcolor{colorTab}
    DPoser & \textbf{13.64} & \textbf{33.18} & \textbf{13.45} & \textbf{35.32} \\
    \bottomrule[\heavyrulewidth]
  \end{tabular}
  \caption{Performance comparison of motion denoising under varying noise scales. MPJPE is reported afters denoising.}
  \label{tab:more motion}
\end{table}
To further evaluate DPoser's performance in motion denoising, we extend our analysis to scenarios with varying noise levels. In complement to the results presented in Table~3 of our main text, we conduct an in-depth examination that spans a broader range of noise conditions. The extended results, detailed in Table~\ref{tab:more motion}, showcase DPoser's exceptional performance against state-of-the-art (SOTA) pose priors, especially under high noise conditions, manifesting DPoser's resilience to noise.

\subsection{Motion Denoising (Partial Input)}
\begin{table}[t]
  \centering
  \setlength\tabcolsep{6pt}
  \resizebox{\linewidth}{!}{
  \begin{tabular}{llcccc}
    \toprule[\heavyrulewidth]
    \makecell[c]{\multirow{2}{*}[-1.0ex]{Methods}} & \makecell[c]{\multirow{2}{*}[-1.0ex]{Occlusion}} & \multicolumn{3}{c}{MPJPE} & MPVPE \\
    \cmidrule(lr){3-5} \cmidrule(l){6-6}
    & & Vis. & Occ. & All & All \\
    \midrule[\lightrulewidth]
    w/o prior & legs & 0.26 & 14.72 & 5.52 & 5.45 \\
    VPoser~\cite{pavlakos2019expressive} & legs & 1.75 & 14.29 & 6.31 & 7.38 \\
    Pose-NDF~\cite{tiwari2022pose} & legs & \textbf{0.25} & 15.71 & 5.87 & 5.64 \\
    \rowcolor{colorTab}
    DPoser & legs & 0.28 & \textbf{12.24} & \textbf{4.63} & \textbf{3.65} \\
    \midrule[\lightrulewidth]
    w/o prior & left arm & 0.26 & 24.87 & 4.74 & 9.91 \\
    VPoser~\cite{pavlakos2019expressive} & left arm & 1.21 & 13.23 & 3.40 & 7.68 \\
    Pose-NDF~\cite{tiwari2022pose} & left arm & \textbf{0.25} & 17.70 & 3.42 & 7.86 \\
    \rowcolor{colorTab}
    DPoser & left arm & 0.27 & \textbf{7.80} & \textbf{1.64} & \textbf{3.81} \\
    \bottomrule[\heavyrulewidth]
  \end{tabular}
  }
  \caption{Comparative analysis of methods for motion denoising with different occlusions (legs or left arm) on the AMASS dataset~\cite{mahmood2019amass}.}
  \label{tab:motion_completion}
\end{table}
We next assess the performance of our model in scenarios involving partial input using the AMASS dataset~\cite{mahmood2019amass}. Two types of occlusions were considered: legs and left arm. The quantitative results of these experiments are presented in Table~\ref{tab:motion_completion}, while visual examples can be found in Section~\ref{sec:visual}. Errors (in \emph{cm}) are evaluated in terms of MPJPE across visible (Vis.), occluded (Occ.), and all joints, along with MPVPE for all vertices.

In the leg occlusion scenario, where the AMASS dataset primarily consists of straight poses, the lack of diversity allows for reasonable results even without incorporating a pose prior. In this case, the optimization starts from an initial point that closely matches these common poses. However, while VPoser's mean-centered approach struggles to faithfully replicate visible areas, DPoser accurately handles the visible portions and guides the reconstruction of occluded parts, yielding more realistic results. In contrast, Pose-NDF does not effectively enhance the occluded regions. For left arm occlusions, which involve more varied movements, DPoser markedly surpasses other methods, underlining its adaptability and precision in handling diverse motion patterns.

\begin{table}[t]
  \centering
  \vspace{-1mm}
  \small
  \setlength{\tabcolsep}{3pt}
  \begin{tabular}{lcccccc}
    \toprule[\heavyrulewidth]
    Methods & APD $\uparrow$ & FID $\downarrow$ & Prec. $\uparrow$ & Rec. $\uparrow$ & \(d_{NN}\) $\downarrow$ \\
    \midrule[\lightrulewidth]
    VPoser~\cite{pavlakos2019expressive} & 1.99 & 0.21 & 0.68 & 0.65 & 1.85 \\
    NRDF~\cite{he2024nrdf} & 1.76 & 5.20 & 0.17 & 0.65 & 5.37 \\
    \rowcolor{colorTab}
    DPoser-hand & \textbf{2.36} & \textbf{0.01} & \textbf{0.82} & \textbf{0.87} & \textbf{1.45} \\
    \bottomrule[\heavyrulewidth]
  \end{tabular}
  \caption{Quantitative evaluation of hand pose generation.}
  \label{tab:hand_generation}
  \vspace{-2mm}
\end{table}

\subsection{Hand Pose Generation}
We evaluate the generated hand poses based on their diversity and realism. As shown in Table~\ref{tab:hand_generation}, DPoser produces a strong combination of both, outperforming methods like VPoser~\cite{pavlakos2019expressive} and NRDF~\cite{he2024nrdf}. Specifically, NRDF shows poor realism, reflected in high FID and \(d_{NN}\) scores. VPoser, while achieving moderate precision, suffers from limited diversity, as indicated by its low APD. See Fig.~\ref{fig:hand_generation} visualization comparison.

\begin{table*}[t]
  \centering
  \begin{tabular}{lcccc}
    \toprule[\heavyrulewidth]
    Methods & PA-MPJPE $\downarrow$ & PA-MPVPE $\downarrow$ & F@5 $\uparrow$ & F@15 $\uparrow$ \\
    \midrule[\lightrulewidth]
    w/o prior & 17.71/16.12 & 18.40/17.04 & 0.396/0.446 & 0.875/0.895 \\
    L2 prior & 12.87/11.49 & 12.71/11.59 & 0.512/0.533 & 0.924/0.927 \\
    VPoser~\cite{pavlakos2019expressive} & 12.31/10.62 & 12.23/10.91 & 0.524/0.609 & 0.931/0.943 \\
    NRDF~\cite{he2024nrdf} & 13.19/11.04 & 13.39/11.59 & 0.469/0.554 & 0.914/0.937 \\
    \rowcolor{colorTab}
    DPoser-hand & \textbf{10.71}/\textbf{8.68} & \textbf{10.48}/\textbf{8.70} & \textbf{0.574}/\textbf{0.679} & \textbf{0.947}/\textbf{0.963} \\
    \midrule[\lightrulewidth]
    hand4whole & 8.50 & 7.81 & 0.651 & 0.97 \\
    + w/o prior & 9.13/6.04 & 8.97/6.06 & 0.609/0.749 & 0.965/0.985 \\
    + L2 prior & 9.91/7.16 & 9.69/7.11 & 0.568/0.686 & 0.953/0.974 \\
    + VPoser~\cite{pavlakos2019expressive} & 9.13/6.42 & 9.04/6.55 & 0.605/0.717 & 0.964/0.981 \\
    + NRDF~\cite{he2024nrdf} & 9.00/6.15 & 8.99/6.29 & 0.595/0.726 & 0.964/0.983 \\
    \rowcolor{colorTab}
    + DPoser-hand & \textbf{7.96}/\textbf{5.36} & \textbf{7.69}/\textbf{5.20} & \textbf{0.663}/\textbf{0.793} & \textbf{0.973}/\textbf{0.990} \\
    \bottomrule[\heavyrulewidth]
  \end{tabular}
  \caption{Performance evaluation of hand mesh recovery on the FreiHAND dataset~\cite{zimmermann2019freihand}. Results are reported using 2D keypoints detected by RTMPose~\cite{jiang2023rtmpose} / ground truth.}
  \label{tab:HMR_hand}
\end{table*}

\subsection{Hand Mesh Recovery}

To evaluate DPoser's ability to recover hand meshes, we test its performance on the FreiHAND dataset~\cite{zimmermann2019freihand} under two initialization strategies: mean poses and the Hand4Whole~\cite{moon2022accurate} prediction poses. The results, detailed in Table~\ref{tab:HMR_hand} and visually represented in Section~\ref{sec:visual} (Fig.~\ref{fig:freihand_gt_appendix} and Fig.~\ref{fig:freihand_mmpose_appendix}), show DPoser's superior performance across various metrics and initialization settings. 

DPoser consistently outperforms competing methods, such as VPoser~\cite{pavlakos2019expressive} and NRDF~\cite{he2024nrdf}, achieving the lowest PA-MPJPE and PA-MPVPE values. For example, when using keypoints detected by RTMPose~\cite{jiang2023rtmpose}, DPoser reduces PA-MPJPE by 20\% compared to VPoser. Moreover, the performance is further enhanced when using Hand4Whole~\cite{moon2022accurate} initialization, highlighting DPoser's ability to refine results from existing SOTA mesh recovery models. In contrast, methods like the L2 prior and VPoser, which rely on mean-centered priors, fail to match the quality of the initializations, producing poorer results. Additionally, DPoser demonstrates significant advantages over NRDF in modeling hand pose distributions, offering more reliable guidance in mesh recovery. By leveraging ground truth (GT) keypoints, DPoser consistently recovers natural hand meshes that align well with observed 2D keypoints.

\begin{table*}[t]
\centering
\begin{tabular}{lcccccc}
    \toprule[\heavyrulewidth]
    \makecell[c]{\multirow{2}{*}[-1.0ex]{Methods}} & \multicolumn{2}{c}{2mm} & \multicolumn{2}{c}{5mm} & \multicolumn{2}{c}{10mm} \\
    \cmidrule(r){2-3} \cmidrule(lr){4-5} \cmidrule(l){6-7}
    & \makecell[c]{MPVPE $\downarrow$} & \makecell[c]{MPJPE $\downarrow$} & \makecell[c]{MPVPE $\downarrow$} & \makecell[c]{MPJPE $\downarrow$} & \makecell[c]{MPVPE $\downarrow$} & \makecell[c]{MPJPE $\downarrow$} \\
    \midrule[\lightrulewidth]
    No prior & 3.95 & 1.50 & 5.82 & 3.46 & 8.62 & 5.97 \\
    L2 prior & 2.10 & 1.43 & 4.06 & 2.92 & 6.06 & 4.27 \\
    VPoser~\cite{pavlakos2019expressive} & 2.47 & 1.36 & 4.15 & 2.85 & 6.32 & 4.40 \\
    NRDF~\cite{he2024nrdf} & 2.67 & 1.40 & 4.57 & 3.11 & 7.18 & 5.06 \\
    \rowcolor{colorTab}
    DPoser-hand & \textbf{1.71} & \textbf{1.17} & \textbf{3.30} & \textbf{2.39} & \textbf{5.39} & \textbf{3.87} \\
    \bottomrule[\heavyrulewidth]
\end{tabular}
\caption{Performance of hand inverse kinematics on the ReInterHand dataset~\cite{moon2024dataset} under noisy settings.}
\label{tab:noisy_ik_hand}
\end{table*}

\subsection{Hand Inverse Kinematics (Noisy Input)}

For hand inverse kinematics, we extend our experiments to noisy settings using the ReInterHand dataset~\cite{moon2024dataset}. Table~\ref{tab:noisy_ik_hand} shows that DPoser consistently outperforms alternative methods across different noise levels (2mm, 5mm, 10mm), achieving the lowest MPVPE and MPJPE. While methods like the L2 prior and VPoser~\cite{pavlakos2019expressive} perform competitively at lower noise levels, their accuracy deteriorates significantly as noise increases. In contrast, DPoser maintains both stability and precision, showcasing its superior ability to handle noisy input and recover plausible hand poses even under challenging conditions.

\subsection{Face Generation}
\begin{table}[t]
  \centering
  \resizebox{\linewidth}{!}{
  \begin{tabular}{lcccc}
    \toprule[\heavyrulewidth]
    Methods & FID $\downarrow$ & Prec. $\uparrow$ & Rec. $\uparrow$ & \(d_{NN}\) $\downarrow$ \\
    \midrule[\lightrulewidth]
    VPoser~\cite{pavlakos2019expressive} (shape) & 31.91 & \textbf{0.984} & 0.105 & \textbf{6.52} \\
    \rowcolor{colorTab}
    DPoser-face (shape) & \textbf{5.331} & 0.689 & \textbf{0.396} & 8.29 \\
    \midrule[\lightrulewidth]
    VPoser~\cite{pavlakos2019expressive} (expression) & 0.888 & \textbf{0.993} & 0.019 & \textbf{0.79} \\
    \rowcolor{colorTab}
    DPoser-face (expression) & \textbf{0.156} & 0.818 & \textbf{0.697} & 1.01 \\
    \bottomrule[\heavyrulewidth]
  \end{tabular}
  }
  \caption{Quantitative evaluation for face generation.}
  \label{tab:face_generation}
\end{table}

\begin{figure*}[t]
    \centering
    \subfloat[VPoser]{
        \includegraphics[width=0.98\linewidth]{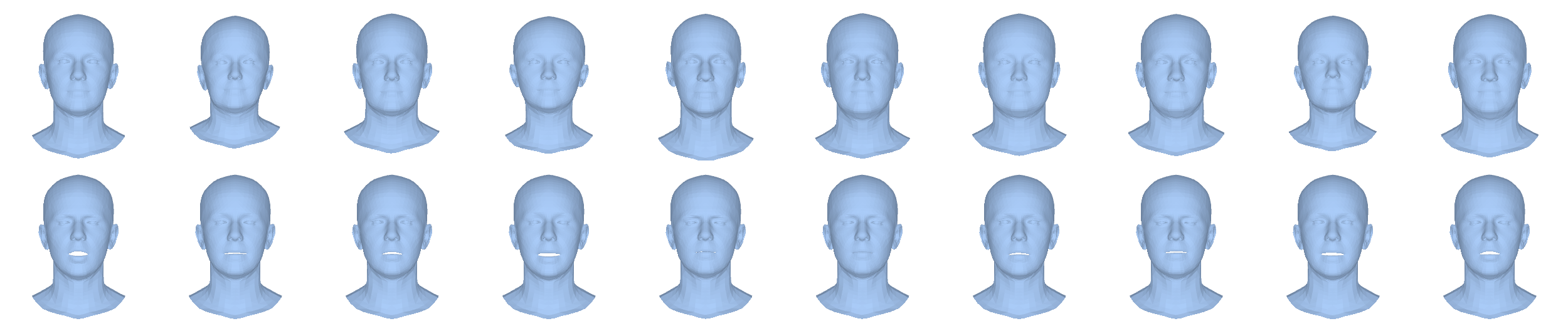}
    }
    \hfill
    \subfloat[DPoser (ours)]{
        \includegraphics[width=0.98\linewidth]{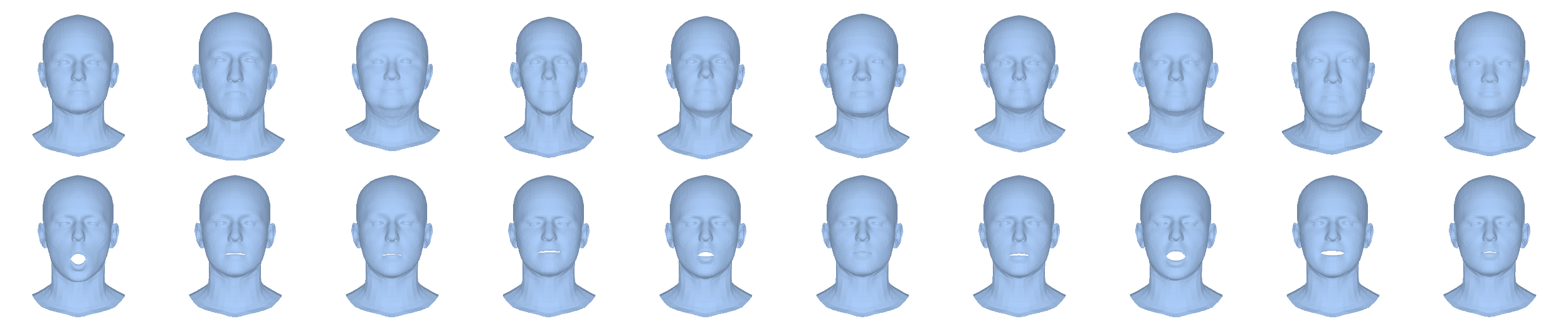}
    }
    \caption{Visualization of face generation results. Top row shows varying face shapes; bottom row shows varying expressions.}
    \label{fig:face_generation}
\end{figure*}
We conduct the face generation experiments for the shape and expression separately since they are uncorrelated attrbutes.
As detailed in Table~\ref{tab:face_generation}, DPoser outperforms VPoser~\cite{pavlakos2019expressive} in terms of FID, achieving values of 5.331 for shape and 0.156 for expression, which highlights DPoser's superior ability to model the distribution of face shapes and expressions. While VPoser achieves higher precision scores, its recall values are considerably lower, indicating a lack of variability in the generated samples. This observation is further corroborated by qualitative results shown in Fig.~\ref{fig:face_generation}, which demonstrate DPoser’s ability to generate a wide variety of realistic face shapes and expressions. Compared to VPoser, DPoser captures a broader range of subtle variations, especially in expressions, while maintaining fidelity.

\subsection{Face Reconstruction}
\begin{table}[t]
  \centering
  \begin{tabular}{lcc}
    \toprule[\heavyrulewidth]
    Methods & all & side-view \\
    \midrule[\lightrulewidth]
    w/o prior & 3.67/4.19 & 3.77/4.46 \\
    L2 prior & 3.56/3.90 & 3.58/4.01 \\
    VPoser~\cite{pavlakos2019expressive} & 3.59/4.01 & 3.62/4.13 \\
    \rowcolor{colorTab}
    DPoser-face & \textbf{3.34}/\textbf{3.65} & \textbf{3.32}/\textbf{3.61} \\
    \midrule[\lightrulewidth]
    EMOCA~\cite{danvevcek2022emoca} & 3.58/4.07 & 3.78/4.43 \\
    + w/o prior & 3.49/3.88 & 3.92/4.56 \\
    + L2 prior & 3.49/3.82 & 3.68/4.28 \\
    + VPoser~\cite{pavlakos2019expressive} & 3.39/3.65 & 3.56/4.05 \\
    \rowcolor{colorTab}
    + DPoser-face & \textbf{3.10}/\textbf{3.54} & \textbf{3.16}/\textbf{3.72} \\
    \bottomrule[\heavyrulewidth]
  \end{tabular}
  \caption{Face reconstruction performance (PA-MPVPE/PA-MPJPE) on the WCPA dataset~\cite{kao2022single}.}
  \label{tab:face_recon_wcpa}
\end{table}

For face reconstruction, along with the NOW beachmark~\cite{sanyal2019learning}, we test on the WCPA~\cite{kao2022single} dataset, which evaluates both face shape and expression. 
As shown in Table~\ref{tab:face_recon_wcpa}, DPoser consistently outperforms other methods. It achieves the lowest PA-MPVPE and PA-MPJPE errors across all configurations, with a notable reduction in errors for both overall and side-view cases. When combined with EMOCA~\cite{danvevcek2022emoca} initialization, DPoser further refines the reconstruction quality, reducing the mean PA-MPVPE error to 3.10 mmm compared to 3.58 mm for EMOCA alone.

Qualitative visualizations in Fig.~\ref{fig:wcpa_comparison} illustrate DPoser’s ability to reconstruct detailed and realistic face meshes, even in challenging scenarios involving variations in side-view poses and complex expressions. While other methods often struggle to generalize across such cases, DPoser remains robust and highly accurate, demonstrating its capability to handle the full diversity of facial shapes and expressions in real-world conditions.

\subsection{Face Inverse Kinematics}
\begin{table*}[t]
  \centering
  \begin{tabular}{lcccccccc}
    \toprule[\heavyrulewidth]
    \makecell[c]{\multirow{2}{*}{Methods}} & \multicolumn{2}{c}{1mm Noise} & \multicolumn{2}{c}{2mm Noise} & \multicolumn{2}{c}{5mm Noise} & \multicolumn{2}{c}{Half Face Occ.} \\
    \cmidrule(lr){2-3} \cmidrule(lr){4-5} \cmidrule(lr){6-7} \cmidrule(l){8-9}
    & MPVPE $\downarrow$ & MPJPE $\downarrow$ & MPVPE $\downarrow$ & MPJPE $\downarrow$ & MPVPE $\downarrow$ & MPJPE $\downarrow$ & MPVPE $\downarrow$ & MPJPE $\downarrow$ \\
    \midrule[\lightrulewidth]
    w/o prior & 1.460 & 0.878 & 2.230 & 1.702 & 4.701 & 4.028 & 0.752 & 0.632 \\
    L2 prior & 1.121 & 0.865 & 1.626 & 1.288 & 2.570 & 2.344 & 0.698 & 0.512 \\
    VPoser~\cite{pavlakos2019expressive} & 1.153 & 0.803 & 1.688 & 1.480 & 2.716 & 2.688 & 0.671 & 0.361 \\
    \rowcolor{colorTab}
    DPoser-face & \textbf{0.784} & \textbf{0.584} & \textbf{1.098} & \textbf{0.963} & \textbf{1.902} & \textbf{1.936} & \textbf{0.427} & \textbf{0.228} \\
    \bottomrule[\heavyrulewidth]
  \end{tabular}
  \caption{Performance of face inverse kinematics on the WCPA dataset~\cite{kao2022single} under noisy and occlusion settings.}
  \label{tab:face_IK_wcpa}
\end{table*}
To evaluate DPoser’s robustness in face inverse kinematics, we conduct experiments under various noise levels and occlusion scenarios using the WCPA dataset~\cite{kao2022single}. The results in Table~\ref{tab:face_IK_wcpa} demonstrate that DPoser consistently achieves the lowest MPVPE and MPJPE errors across all tested conditions. Notably, DPoser retains its strong performance even under extreme noise conditions, whereas VPoser~\cite{pavlakos2019expressive} experiences significant degradation as noise levels increase. Qualitative results, visualized in Fig.~\ref{fig:ik_face_comparison}, further confirm DPoser’s ability to reconstruct realistic and aligned facial details under noisy and occluded conditions.

\subsection{Whole-body Mesh Recovery}
\begin{table}[t]
  \centering
  \small
  \vspace{-1mm}
  \resizebox{\linewidth}{!}{
  \begin{tabular}{lcccc}
    \toprule[\heavyrulewidth]
    \makecell[c]{\multirow{2}{*}[-1.0ex]{Methods}} & \multicolumn{3}{c}{PA-MPVPE$\downarrow$} & {PA-MPJPE$\downarrow$} \\
    \cmidrule(r){2-4} \cmidrule(l){5-5}
    & \makecell[c]{All} & \makecell[c]{Hands} & \makecell[c]{Face} & \makecell[c]{Body} \\
    \midrule[\lightrulewidth]
    w/o prior & 89.72 & 23.51 & 7.26 & 91.18 \\
    GMM~\cite{bogo2016keep} \& L2 prior & 86.95 & 18.22 & 5.38 & 83.58 \\
    VPoser-X~\cite{pavlakos2019expressive} & 81.96 & 17.59 & 6.37 & 86.50 \\
    \rowcolor{colorTab}
    DPoser-X & \textbf{70.91} & \textbf{15.83} & \textbf{5.27} & \textbf{74.33} \\
    \midrule[\lightrulewidth]
    SMPLerX & 25.49 & 18.89 & 2.85 & 28.30 \\
    + w/o prior & 24.72 & 11.92 & 2.78 & 22.98 \\
    + GMM~\cite{bogo2016keep} \& L2 prior & 24.28 & 11.09 & \textbf{2.58} & 22.95 \\
    + VPoser-X~\cite{pavlakos2019expressive} & 24.41 & 10.21 & 2.65 & 23.03 \\
    \rowcolor{colorTab}
    + DPoser-X & \textbf{23.20} & \textbf{8.91} & 2.62 & \textbf{21.22} \\
    \bottomrule[\heavyrulewidth]
  \end{tabular}
  }
  \caption{Whole-body mesh recovery results on the Fit3d dataset~\cite{fieraru2021aifit}.}
  \label{tab:HMR_wholebody_fit3d}
\end{table}
We extend our evaluation of whole-body mesh recovery to include the Fit3D dataset~\cite{fieraru2021aifit}, in addition to the comparative results on ARCTIC~\cite{fan2023arctic}. For this evaluation, we compare DPoser-X with VPoser-X~\cite{pavlakos2019expressive} and the GMM baseline, which utilizes a Gaussian Mixture Model (GMM)~\cite{bogo2016keep} for body poses and an L2 prior for hands and face. 

As shown in Table~\ref{tab:HMR_wholebody_fit3d}, DPoser-X outperforms both VPoser-X~\cite{pavlakos2019expressive} and GMM~\cite{bogo2016keep} across most metrics, for both hands and the entire body. However, we observe that the L2-prior baseline performs better than DPoser-X in terms of PA-MPVPE on the face. We attribute this result to the low-resolution images in the Fit3D dataset, where the face is depicted with limited pixel density and the 2D keypoints are less expressive. In this case, the neutral face produced by the L2 prior is more likely to yield better results due to the lack of detailed facial features in the input. Nonetheless, DPoser-X still outperforms other methods in handling the full-body mesh recovery, showing its robustness in both body and hand mesh reconstruction.

\section{Ablated DPoser's Training}
\label{sec:training}
\begin{table*}[t]
  \centering
  \begin{tabular}{lcccccc}
    \toprule[\heavyrulewidth]
    \makecell[c]{\multirow{2}{*}[-1.5ex]{Normalization}} & \multicolumn{1}{c}{Body Mesh Recovery} & \multicolumn{1}{c}{Body Pose Completion} & \multicolumn{2}{c}{Motion Denoising} \\
    \cmidrule(r){2-2} \cmidrule(lr){3-3} \cmidrule(l){4-5}
    & \makecell[c]{PA-MPJPE $\downarrow$} & \makecell[c]{MPJPE (\(S=10\)) $\downarrow$} & \makecell[c]{MPVPE $\downarrow$} & \makecell[c]{MPJPE $\downarrow$} \\
    \midrule[\lightrulewidth]
    w/o norm & 57.88 & 45.37/102.28/41.08 & 44.82 & 24.04 \\
    min-max & 59.17 & 47.41/107.00/43.42 & 42.70 & 21.29 \\
    \rowcolor{colorTab}
    z-score & \textbf{56.49} & \textbf{34.37}/72.47/26.32 & \textbf{38.57} & \textbf{20.24} \\ 
    \bottomrule[\heavyrulewidth]
  \end{tabular}
  \caption{Comparative performance of normalization methods using axis-angle rotation representation across multiple tasks.}
  \label{tab:normalize}
\end{table*}

\begin{table*}[t]
  \centering
  \begin{tabular}{lcccccc}
    \toprule[\heavyrulewidth]
    \makecell[c]{\multirow{2}{*}[-1.5ex]{Representation}} & \multicolumn{1}{c}{Body Mesh Recovery} & \multicolumn{1}{c}{Body Pose Completion} & \multicolumn{2}{c}{Motion Denoising} \\
    \cmidrule(r){2-2} \cmidrule(lr){3-3} \cmidrule(l){4-5}
    & \makecell[c]{PA-MPJPE $\downarrow$} & \makecell[c]{MPJPE (\(S=10\)) $\downarrow$} & \makecell[c]{MPVPE $\downarrow$} & \makecell[c]{MPJPE $\downarrow$} \\
    \midrule[\lightrulewidth]
    \rowcolor{colorTab}
    axis-angle & \textbf{56.05} & \textbf{34.76}/72.41/26.09 & \textbf{38.21} & \textbf{19.87} \\
    6D rotations & 57.54 & 40.89/81.43/27.31 & 38.44 & 20.12 \\
    \bottomrule[\heavyrulewidth]
  \end{tabular}
  \caption{Comparative performance of rotation representations using z-score normalization across multiple tasks.}
  \label{tab:rotation}
\end{table*}

This section dissects the impact of different rotation representations and normalization techniques on DPoser's performance. The ablation of training experiments is conducted for the DPoser-body model trained on AMASS~\cite{mahmood2019amass}. Initially, we examine axis-angle representation, comparing various normalization strategies: min-max scaling, z-score normalization, and no normalization. Our findings, summarized in Table~\ref{tab:normalize}, indicate that z-score normalization is generally the most effective. Subsequently, using this optimal normalization, we explore 6D rotations~\cite{zhou2019continuity} as an alternative. As evidenced by Table~\ref{tab:rotation}, axis-angle representation offers superior performance. This preference can be attributed to the effective modeling capabilities of diffusion models, which do not benefit much from a more continuous data representation.

Inspired by HuMoR~\cite{rempe2021humor}, we experiment with integrating the SMPL body model~\cite{loper2015smpl} as a regularization term during training. Alongside the prediction of additive noise, as outlined in Eq.~(4) in the main text, we employ a 10-step DDIM sampler~\cite{song2020denoising} to recover a ``clean'' version of the pose, denoted as \(\tilde{\mathbf{x}}_0\), from the diffused \(\mathbf{x}_t\). The regularization loss aims to minimize the discrepancy between the original and recovered poses under the SMPL body model \(M\):
\begin{align}
L_\mathrm{reg} &= ||M_J(\tilde{\mathbf{x}}_0, \beta_0) - M_J(\mathbf{x}_0, \beta_0)||_2^2 \nonumber \\
& \quad + ||M_V(\tilde{\mathbf{x}}_0, \beta_0) - M_V(\mathbf{x}_0, \beta_0)||_2^2.
\end{align}
Here, \(\beta_0\) represents the mean shape parameters in SMPL. To account for denoising errors, we scale the regularization loss by \(\mathrm{log} (1+\frac{\alpha_t}{\sigma_t})\), thereby increasing the weight for samples with smaller \(t\) values (less noise). 

\begin{figure}
    \centering
    \includegraphics[width=\linewidth]{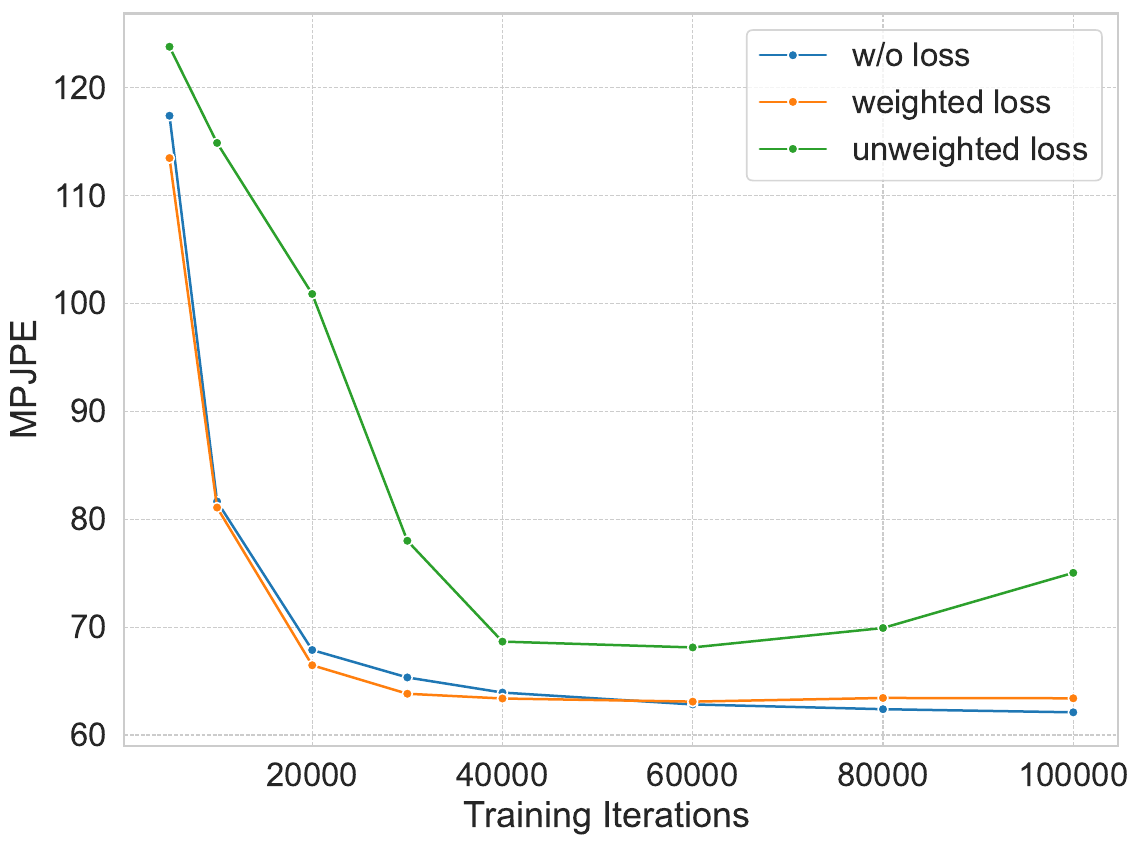}
    \caption{MPJPE evolution in DPoser training with different regularization loss settings for body pose completion, assessed on AMASS~\cite{mahmood2019amass} with 10 hypotheses under legs occlusion scenarios.}
    \label{fig:regloss}
    \vspace{-3mm}
\end{figure}

Fig.~\ref{fig:regloss} visualizes the impact of this regularization on MPJPE during the training, specifically for pose completion tasks with occlusion of both legs. 
We observe that weighted regularization offers slight performance gains in the early training process, while the absence of weighting introduces instability and deterioration in results. Despite these insights, the computational cost of incorporating the SMPL model—especially for our large batch size of 1280—makes the training approximately 8 times slower. Therefore, we opted not to include this regularization in our main experiments.

\begin{table}[t]
    \centering
    \resizebox{0.47\textwidth}{!}{\noindent
        \begin{tabular}{ccccc}
            \toprule[\heavyrulewidth]
             Steps & Blocks & Hidden Dim & HMR (PA-MPJPE) & Runtime (s) \\
            \midrule[\lightrulewidth]
            500 & 2 & 1024 & 56.05 & 17.34 \\
            \midrule[\lightrulewidth]
            250 & 2 & 1024 & 56.53 & \textbf{8.44} \\
            1000 & 2 & 1024 & \textbf{55.74} & 34.11 \\
            500 & 4 & 1024 & 56.47 & 18.72 \\
            500 & 2 & 2048 & 56.67 & 19.12 \\
          \bottomrule[\heavyrulewidth]
          \end{tabular}
      }
    \caption{Ablation of DPoser's architecture and optimization steps for the HMR task.}
    \label{tab:network_ablation}
    \vspace{-3.5mm}
\end{table}
    
We ablate the architectural hyperparameters of DPoser-body and the number of optimization steps on the HMR task, with results shown in Table~\ref{tab:network_ablation}. Our findings indicate that a more complex architecture (i.e., 4 blocks or a 2048 hidden dimension) does not improve accuracy. Regarding the optimization, increasing the steps to 1000 offers the best accuracy (55.74 PA-MPJPE) but at a high computational cost (34.11s), while 250 steps are fastest but less accurate. Based on this analysis, we adopt the configuration of \emph{500 steps, 2 blocks, and a 1024 hidden dimension} for our experiments, as it provides a solid trade-off between accuracy and runtime efficiency.

\begin{table*}[t]
  \centering
    \begin{tabular}{lcccc}
    \toprule[\heavyrulewidth]
    Methods & Occ. left leg & Occ. legs & Occ. arms & Occ. torso\\
    \midrule[\lightrulewidth]
    ScoreSDE~\cite{song2020score} & 48.73/106.32/41.30 & 74.68/128.32/37.27 & 66.89/127.86/48.15 & 16.69/34.54/12.21 \\
    DPS~\cite{chung2022diffusion} & 40.51/104.32/54.57 & 64.26/113.46/33.71 & 60.63/119.85/42.78 & 15.10/33.90/13.27 \\
    MCG~\cite{chung2022improving} & 49.04/106.37/41.07 & 74.90/128.53/37.40 & 66.17/127.72/48.15 & 16.69/34.66/12.23 \\
    \rowcolor{colorTab}
    DPoser & \textbf{35.37}/74.01/26.47 & \textbf{59.25}/96.77/24.55 & \textbf{51.27}/81.76/20.04 & \textbf{13.95}/28.57/9.85 \\
    \bottomrule[\heavyrulewidth]
    \end{tabular}
    \caption{Comparative evaluation of diffusion-based solvers for body pose completion on the AMASS dataset~\cite{mahmood2019amass}. The min/mean/std of MPJPE are reported (hypotheses number \(S=10\)).}
  \label{tab:solvers}
\end{table*}

\section{Extended DPoser's Optimization}
\label{sec:testing}

In addressing pose-centric tasks as inverse problems, we propose a versatile optimization framework, which employs variational diffusion sampling as its foundational approach~\cite{mardani2023variational}. Our exploration extends to an array of diffusion-based methodologies for solving these complex inverse problems. Among the techniques considered are ScoreSDE~\cite{song2020score}, MCG~\cite{chung2022improving}, and DPS~\cite{chung2022diffusion}.
These methods augment standard generative processes with observational data, either by employing gradient-based guidance or back-projection techniques.
We compare these methods with our DPoser for body pose completion tasks. Our findings, captured in Table~\ref{tab:solvers}, reveal that DPoser outperforms the competitors under most occlusion conditions. Consequently, DPoser emerges not merely as a universally applicable solution to pose-related tasks, but also as an exceptionally efficient one.

It is worth mentioning that methods rooted in generative frameworks~\cite{song2020score, chung2022improving, chung2022diffusion, kawar2022denoising} can pose challenges for broader applicability in pose-centric tasks. For instance, in blind inverse problems—certain parameters in \(\mathcal{A}\) (e.g., camera models in HMR) are unknown—generative methods are less straightforward to implement. 
ZeDO~\cite{jiang2023back}, a recent study focusing on the 2D-3D lifting task, adopts the ScoreSDE~\cite{song2020score} framework and refines camera translations by solving an optimization sub-problem after each generative step.
However, directly porting this strategy to HMR is non-trivial, owing to the added complexity of body shape parameter optimization—a feature currently absent in our DPoser model. 
Although some state-of-the-art techniques~\cite{chung2023parallel, murata2023gibbsddrm} offer solutions by jointly modeling operator \(\mathcal{A}\) and data distributions, a full-fledged discussion on this subject is beyond this paper's purview and remains an open question for future work.

\begin{figure}[t]
    \centering
    \includegraphics[width=\linewidth]{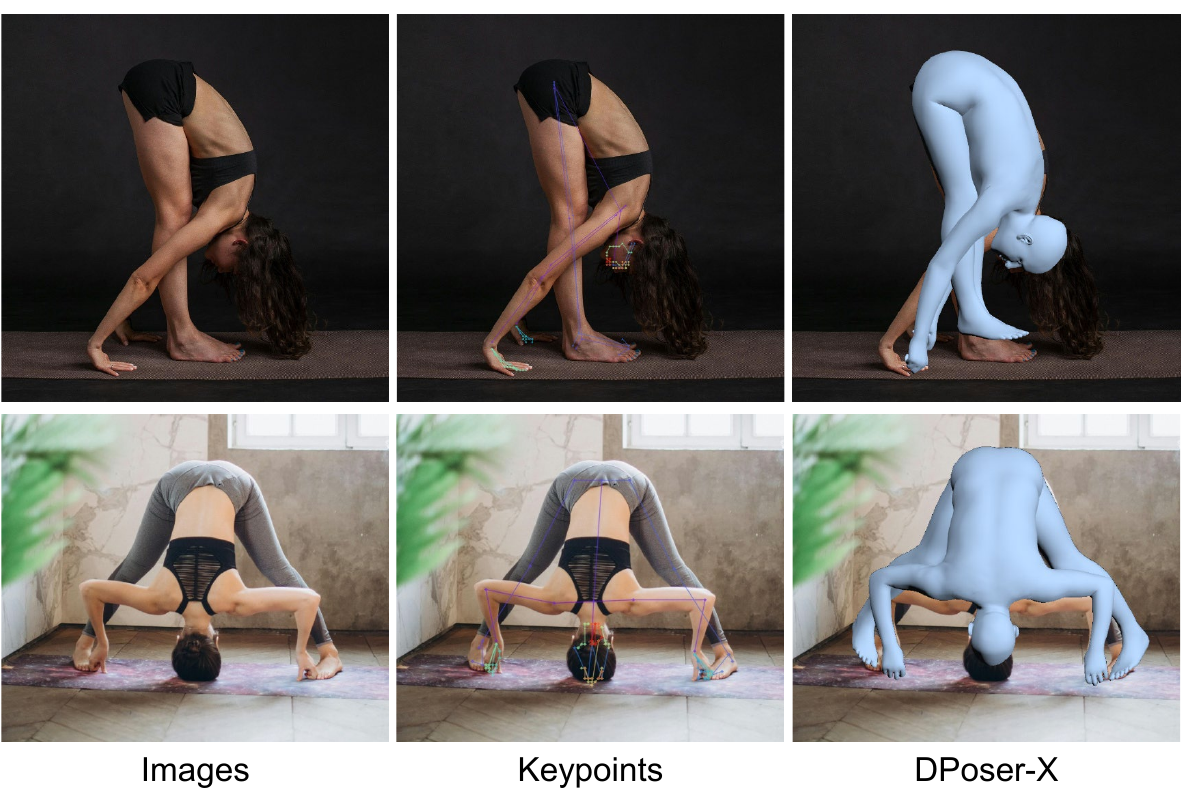}
    \vspace{-5mm}
    \caption{Failure cases of our method on challenging yoga poses. Inaccuracies in the estimated 2D keypoints (middle column), combined with our model's limited exposure to such out-of-distribution poses during training, lead to flawed 3D mesh reconstructions (right column).}
    \vspace{-2mm}
\label{fig:failure_cases}
\end{figure}

\section{Limitation and future work}
\label{sec:limitation}
A primary limitation of our work is the dependency on the training data's distribution. Our body pose prior is trained on the AMASS dataset~\cite{mahmood2019amass}, which, while diverse in common daily actions, contains limited examples of challenging or extreme poses like those found in yoga. This data imbalance leads to two main issues. First, the learned prior is inherently biased towards common standing poses. Second, when confronted with out-of-distribution inputs, as illustrated in Fig.~\ref{fig:failure_cases}, the prior may offer limited or even incorrect guidance. This problem is often exacerbated by the failure of off-the-shelf 2D keypoint detectors like ViTPose~\cite{xu2022vitpose} to produce accurate keypoints for such complex images, which in turn misguides the optimization.

Future work could address these data-driven limitations in several ways. To mitigate the action imbalance, techniques like clustering motions with action labels~\cite{punnakkal2021babel} and performing importance sampling during training could be effective. To improve robustness on challenging poses, incorporating more diverse training data and exploring more robust fitting strategies, such as using predicted dense depth maps for supervision, are promising directions.

Our framework also inherits certain limitations from the variational diffusion sampling~\cite{mardani2023variational} process it employs, most notably a tendency towards mode-seeking. For example, minimizing the DPoser regularization loss alone for ``generation'' results in a high Precision of 0.995 but a low Recall of 0.163. The low recall, compared to standard generative diffusion samplers (see Table~1 in the main text), indicates that the optimization framework captures the primary modes of the data distribution accurately but lacks diversity. To address this, future research could explore techniques like particle-based variational inference~\cite{liu2016stein, wang2023prolificdreamer} to enhance solution diversity. Finally, within the broader context of inverse problems we have framed, a plethora of existing methods~\cite{daras2024survey} could be adapted to leverage our diffusion-based pose prior. Exploring these methods holds great potential for future progress.

\section{More Qualitative Results}
\label{sec:visual}
\begin{figure*}[t]
    \centering
    \includegraphics[width=0.9\linewidth]{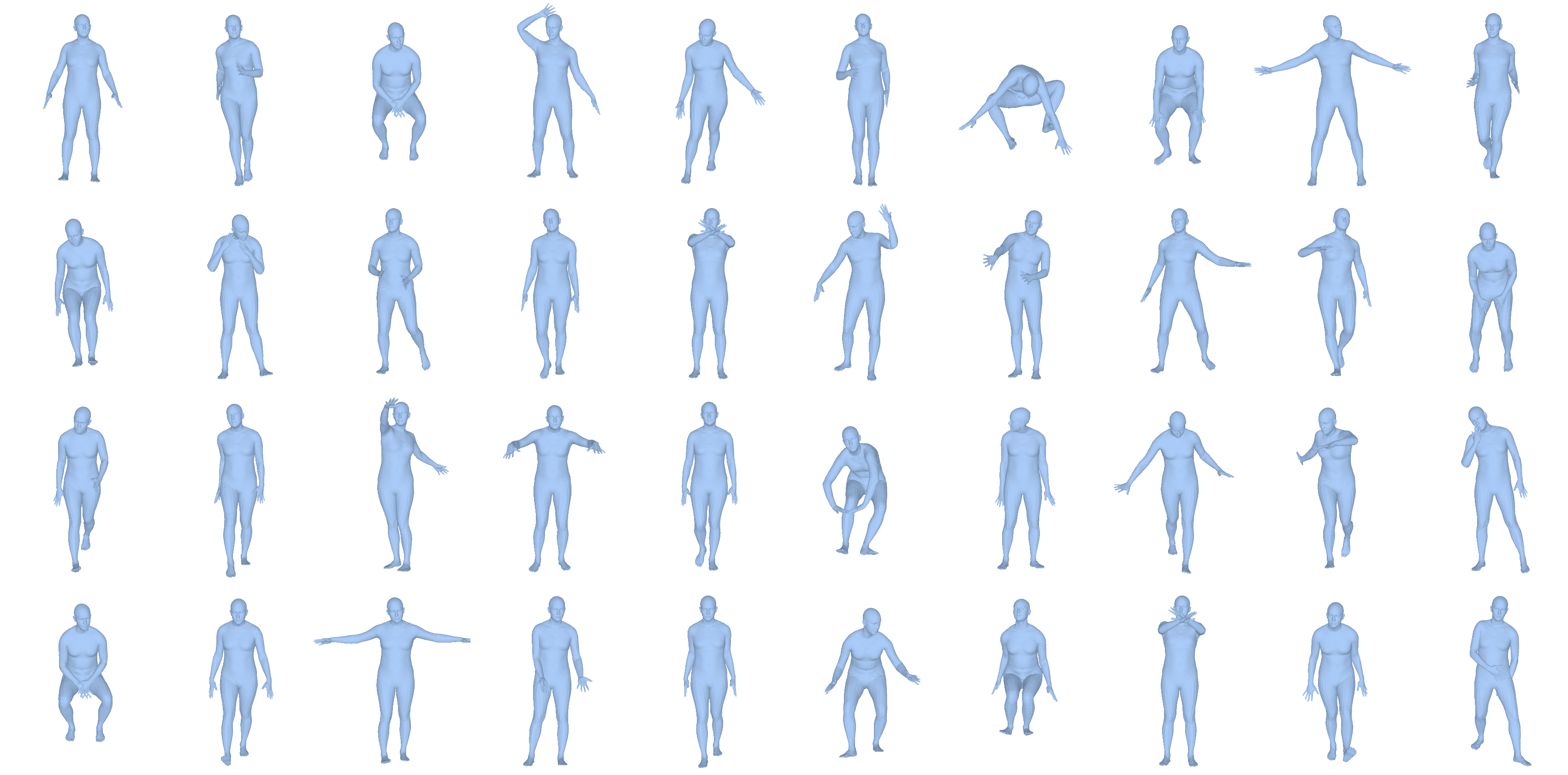}
    \vspace{-3mm}
    \caption{Visualization of body pose generation. DPoser can generate diverse and realistic body poses.}
    \label{fig:more generation}
\end{figure*}

\begin{figure*}[t]
    \centering
    \subfloat{\includegraphics[width=0.9\linewidth]{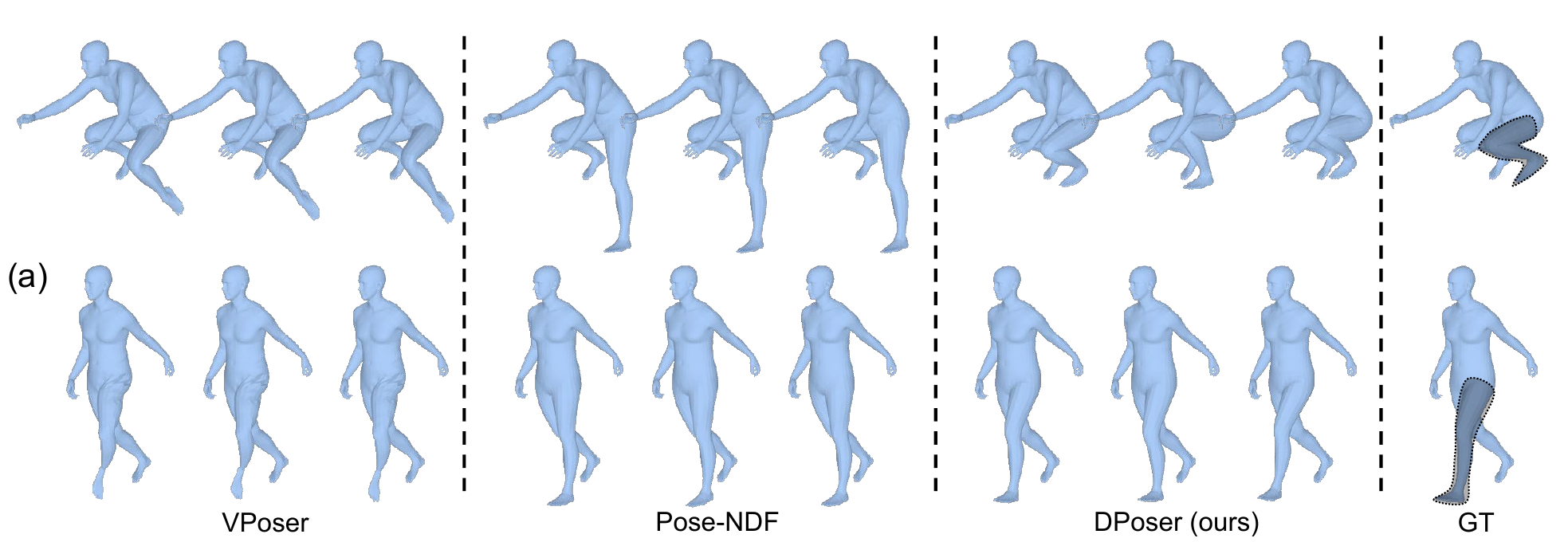}} \hfill
    \subfloat{\includegraphics[width=0.9\linewidth]{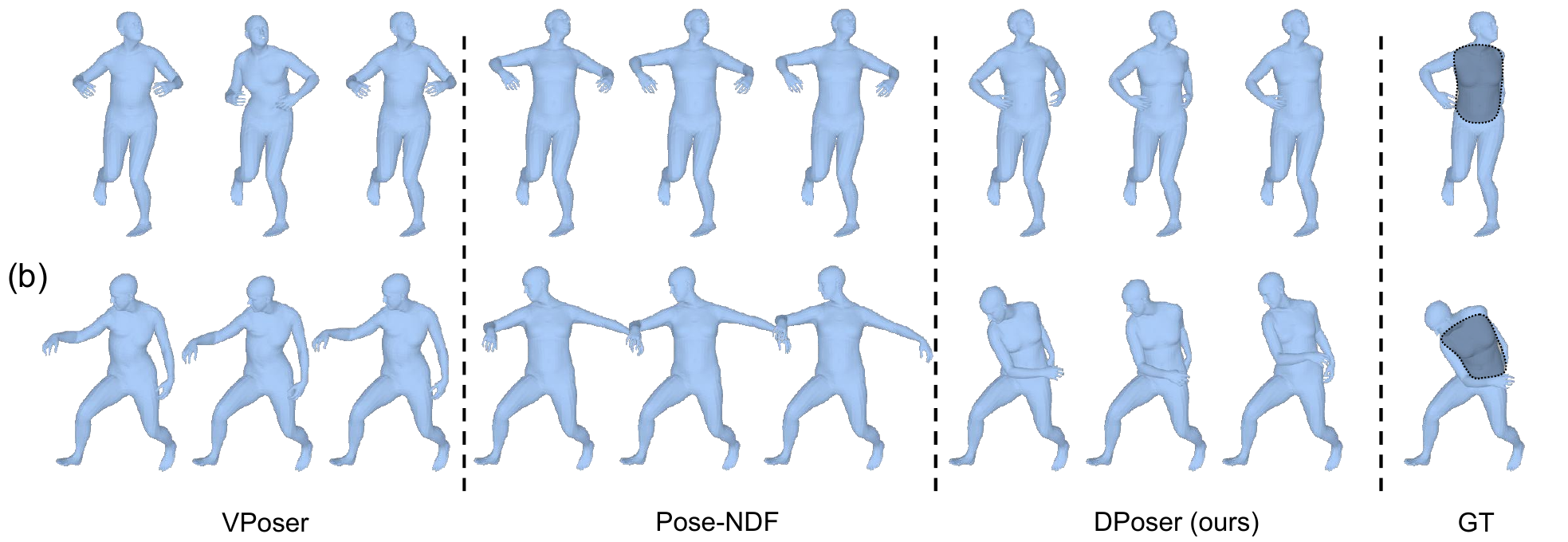}}
    \vspace{-2mm}
    \caption{Visualization of body pose completion. (a) Left leg under occlusion. (b) Torso under occlusion.}
    \label{fig:more completion}
\end{figure*}

\begin{figure*}[t]
    \centering
    \subfloat{\includegraphics[width=\linewidth]{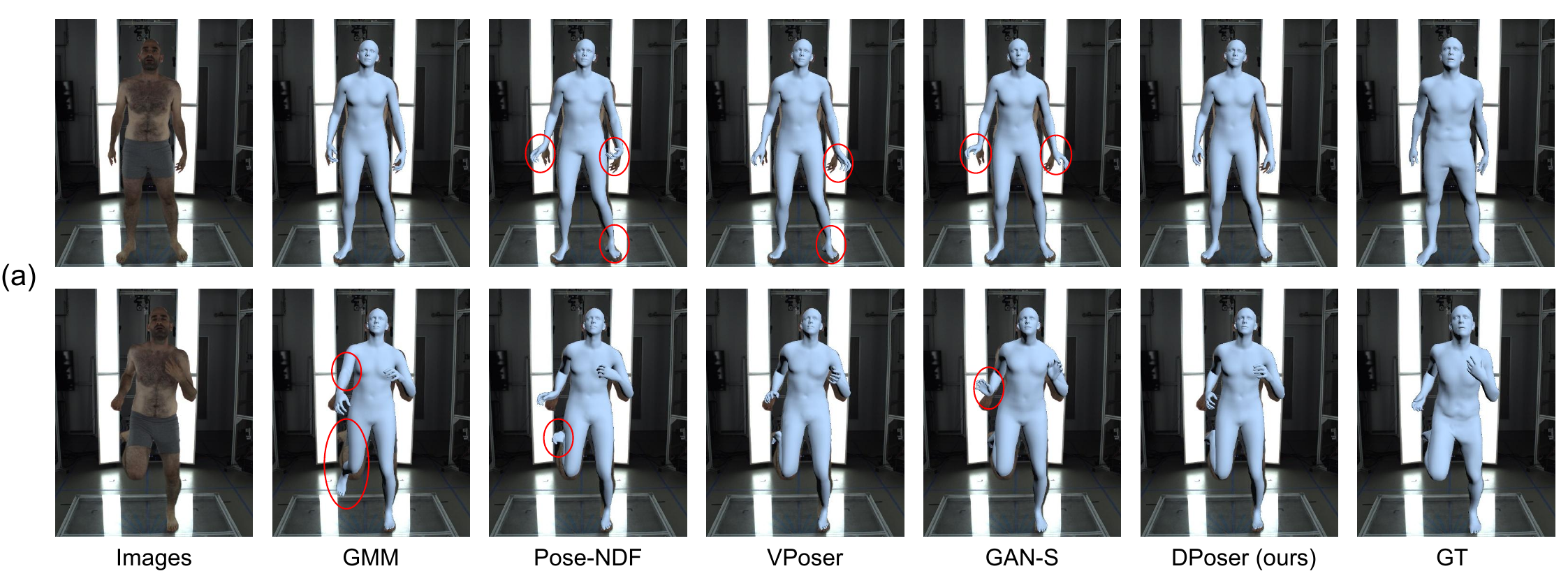}} \hfill
    \subfloat{\includegraphics[width=\linewidth]{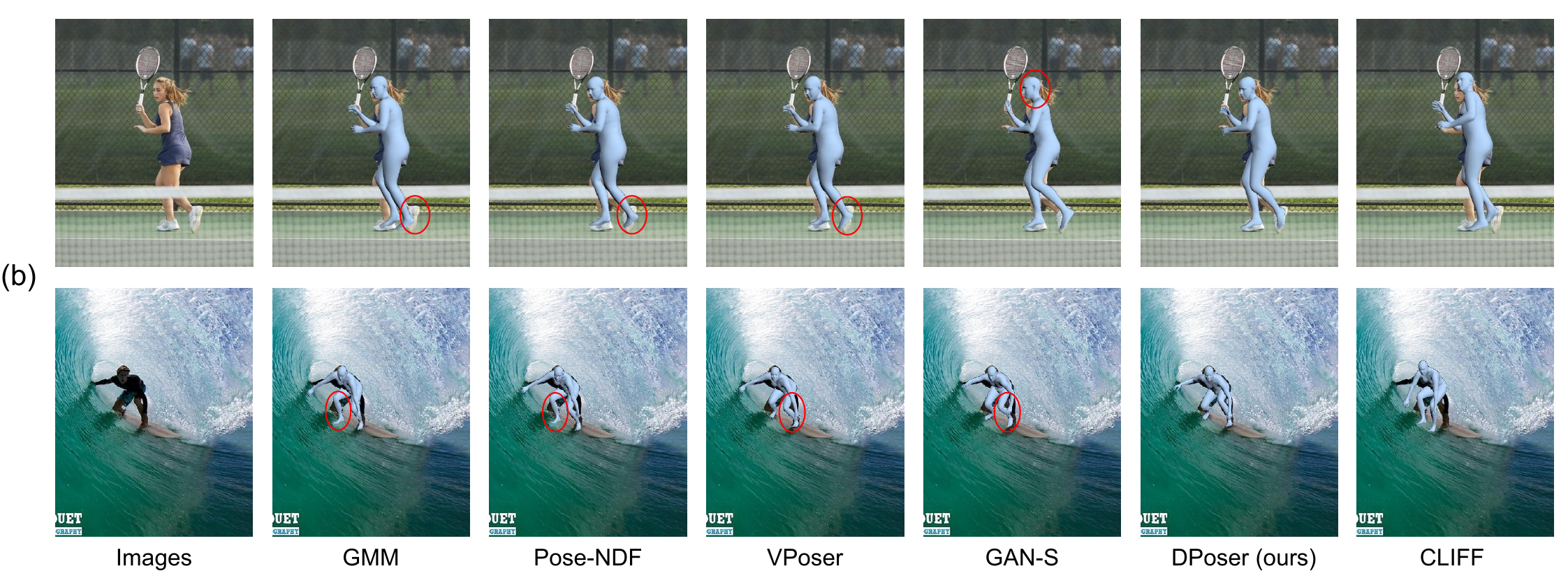}} \hfill
    \subfloat{\includegraphics[width=\linewidth]{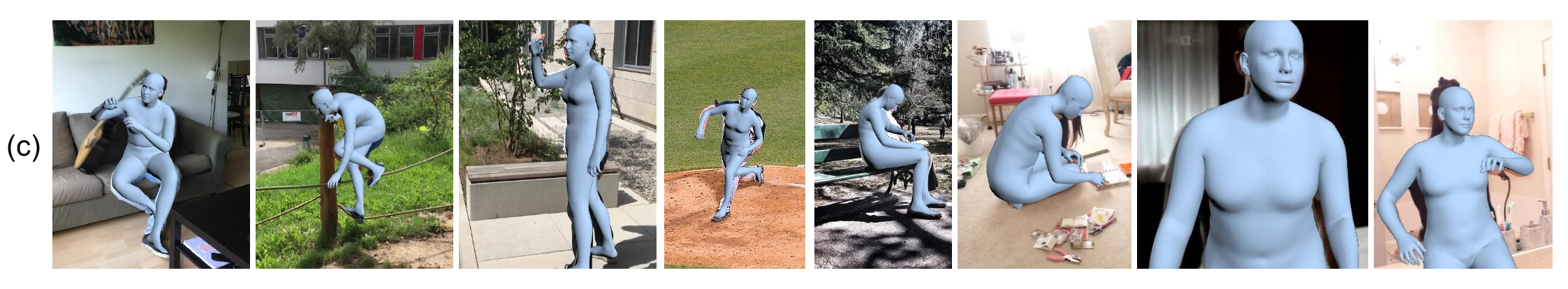}}
    \caption{Visualization of body mesh recovery. (a) Fitting from scratch. (b) Initialization using the CLIFF~\cite{li2022cliff} prediction results. (c) More results of DPoser optimization with CLIFF initialization on in-the-wild images.}
    \label{fig:more hmr}
\end{figure*}

\begin{figure*}[t]
    \centering
    \subfloat[Gaussian noise with 40 mm standard deviation.]{\includegraphics[width=\linewidth]{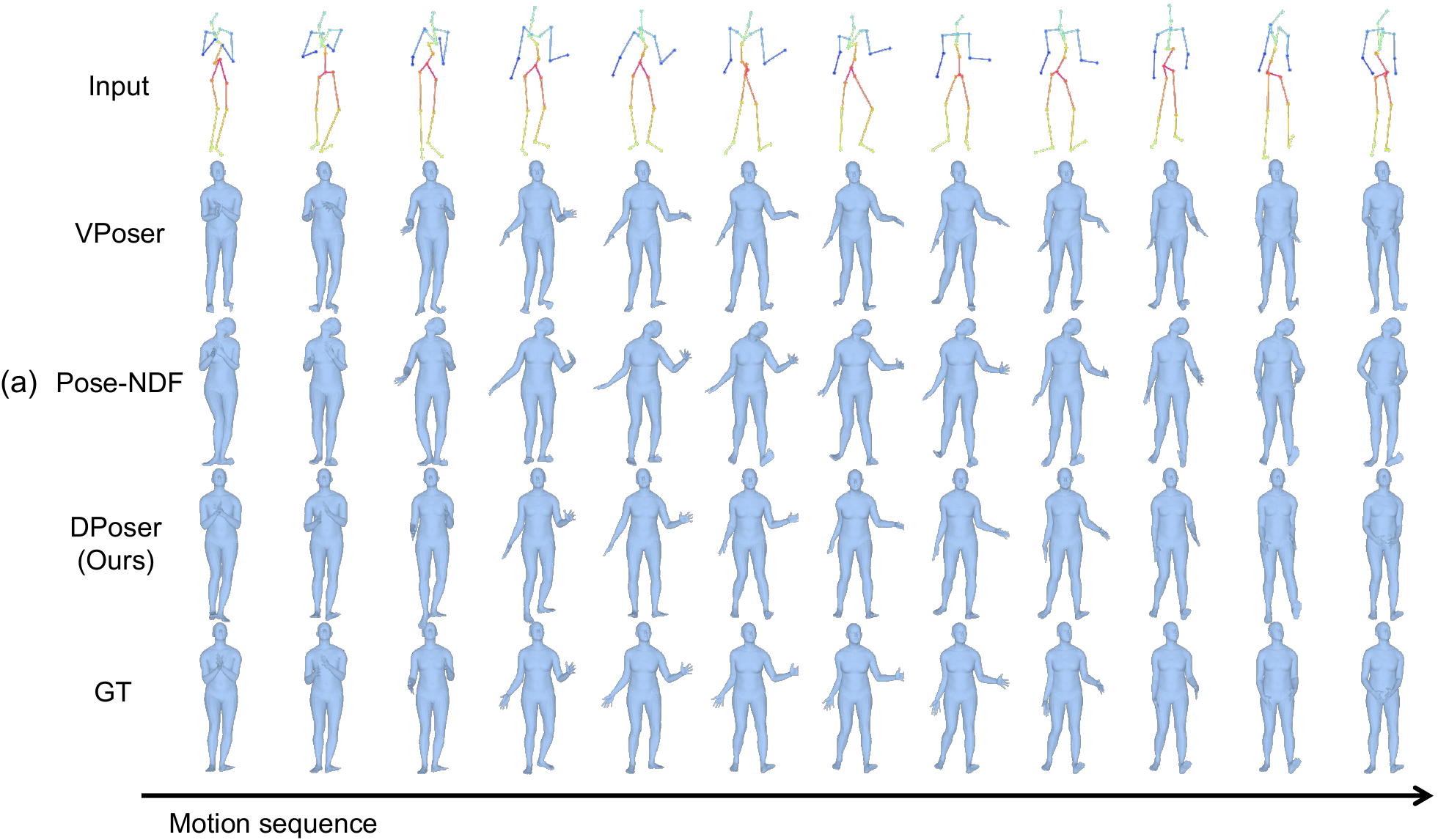}} \hfill
    \subfloat[Gaussian noise with 100 mm standard deviation. ]{\includegraphics[width=\linewidth]{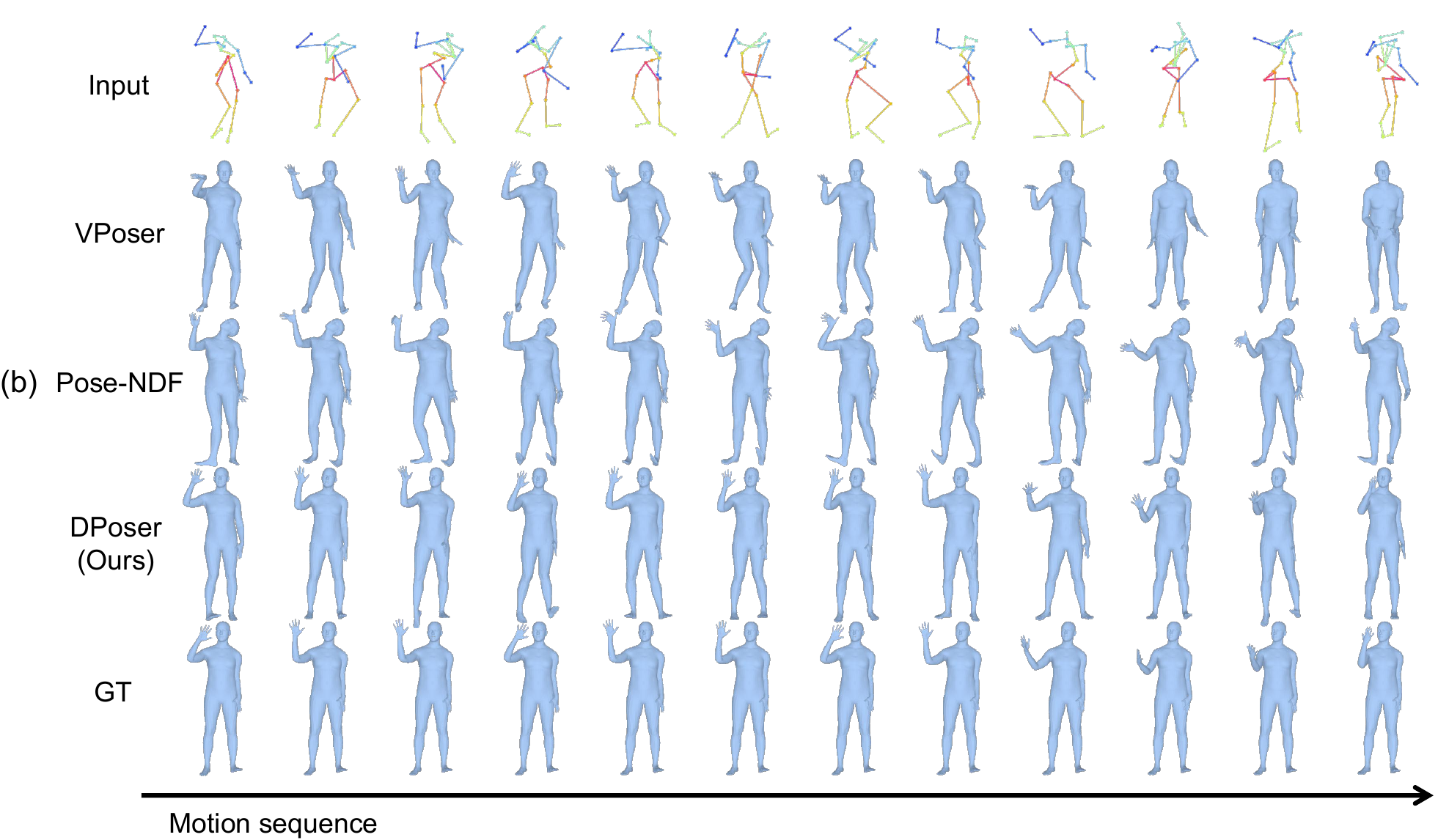}}
    \caption{Visualization of motion denoising with noisy observations. We visualize every 20\(^{th}\) of the sequence.}
    \label{fig:more motion}
\end{figure*}

\begin{figure*}[t]
    \centering
    \subfloat[Legs under occlusion.]{\includegraphics[width=\linewidth]{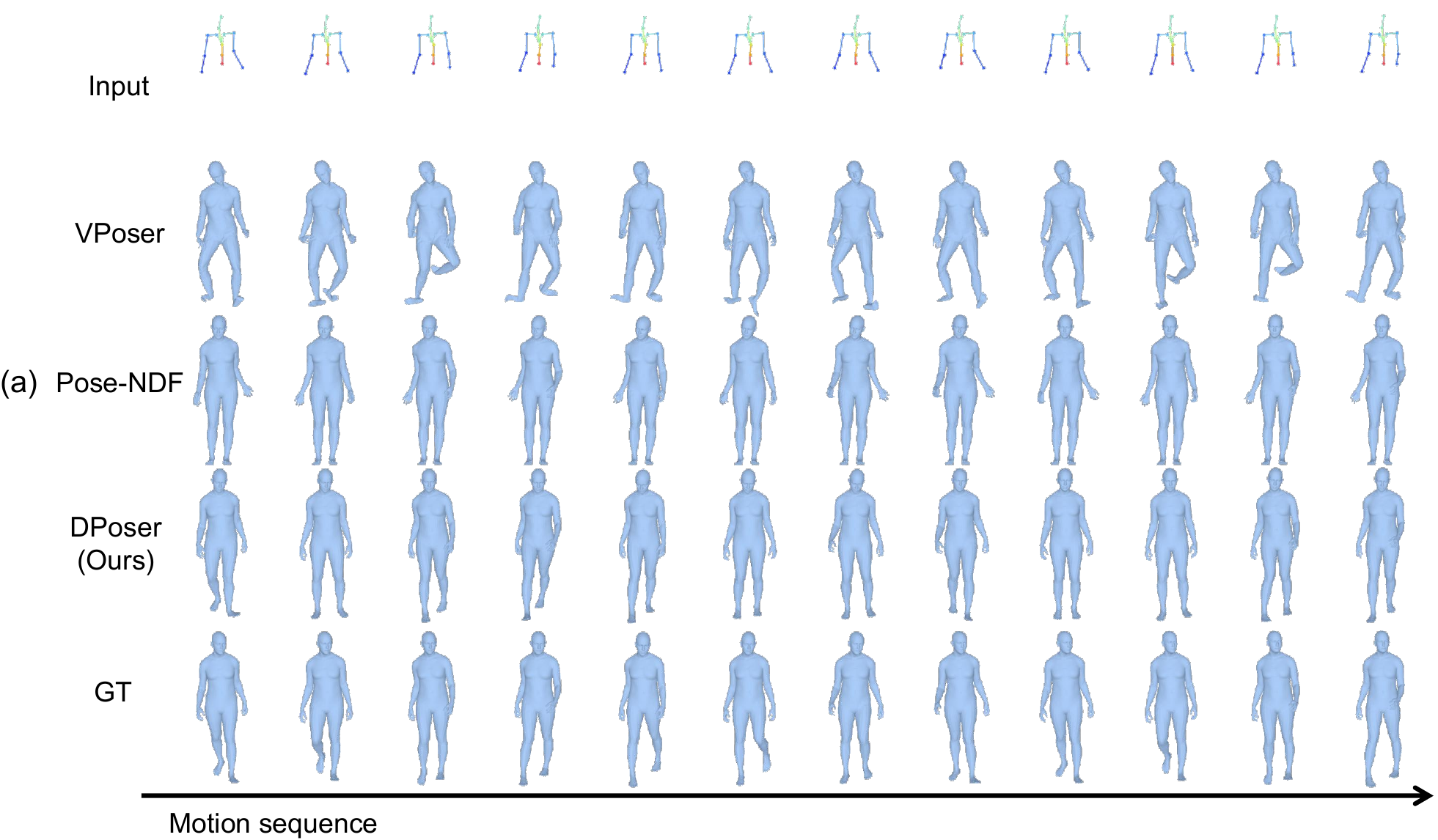}} \hfill
    \subfloat[Left arm under occlusion.]{\includegraphics[width=\linewidth]{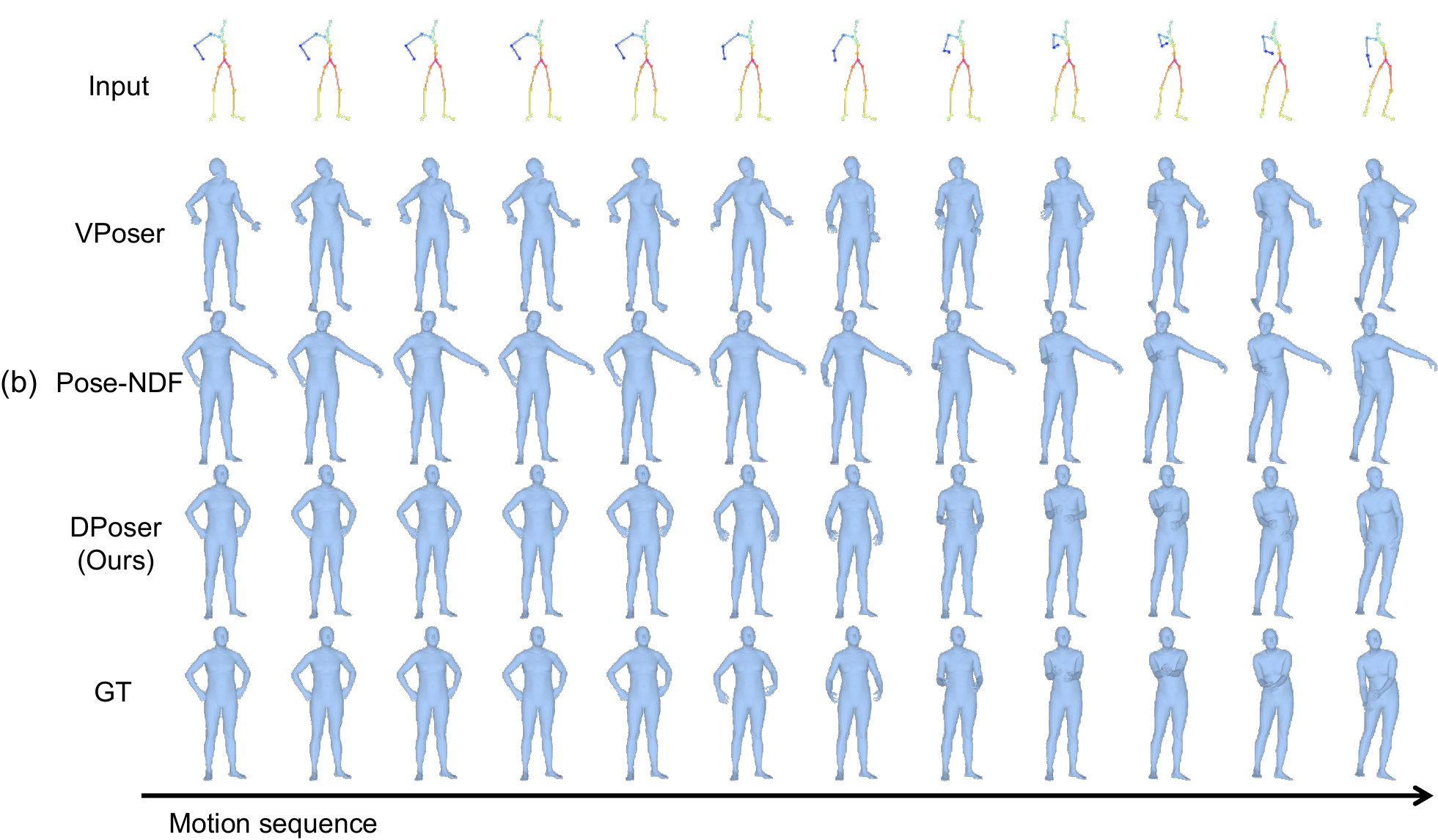}}
    \caption{Visualization of motion denoising with partial observations. We visualize every 20\(^{th}\) of the sequence.}
    \label{fig:motion completion}
\end{figure*}

\begin{figure*}[t]
    \centering
    \subfloat[DPoser (ours)]{
        \includegraphics[width=0.31\linewidth]{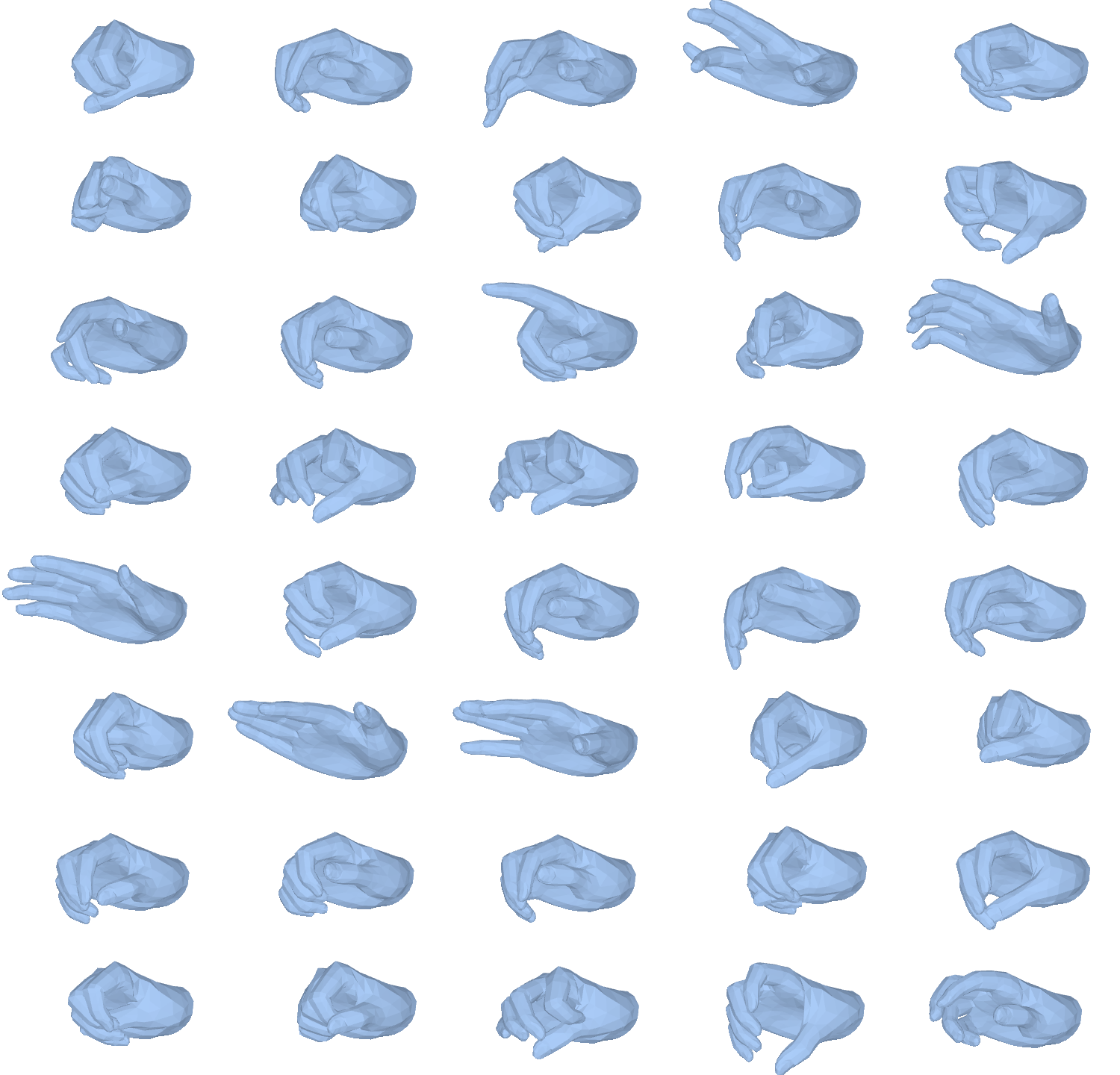}
    }
    \hfill
    \subfloat[VPoser]{
        \includegraphics[width=0.31\linewidth]{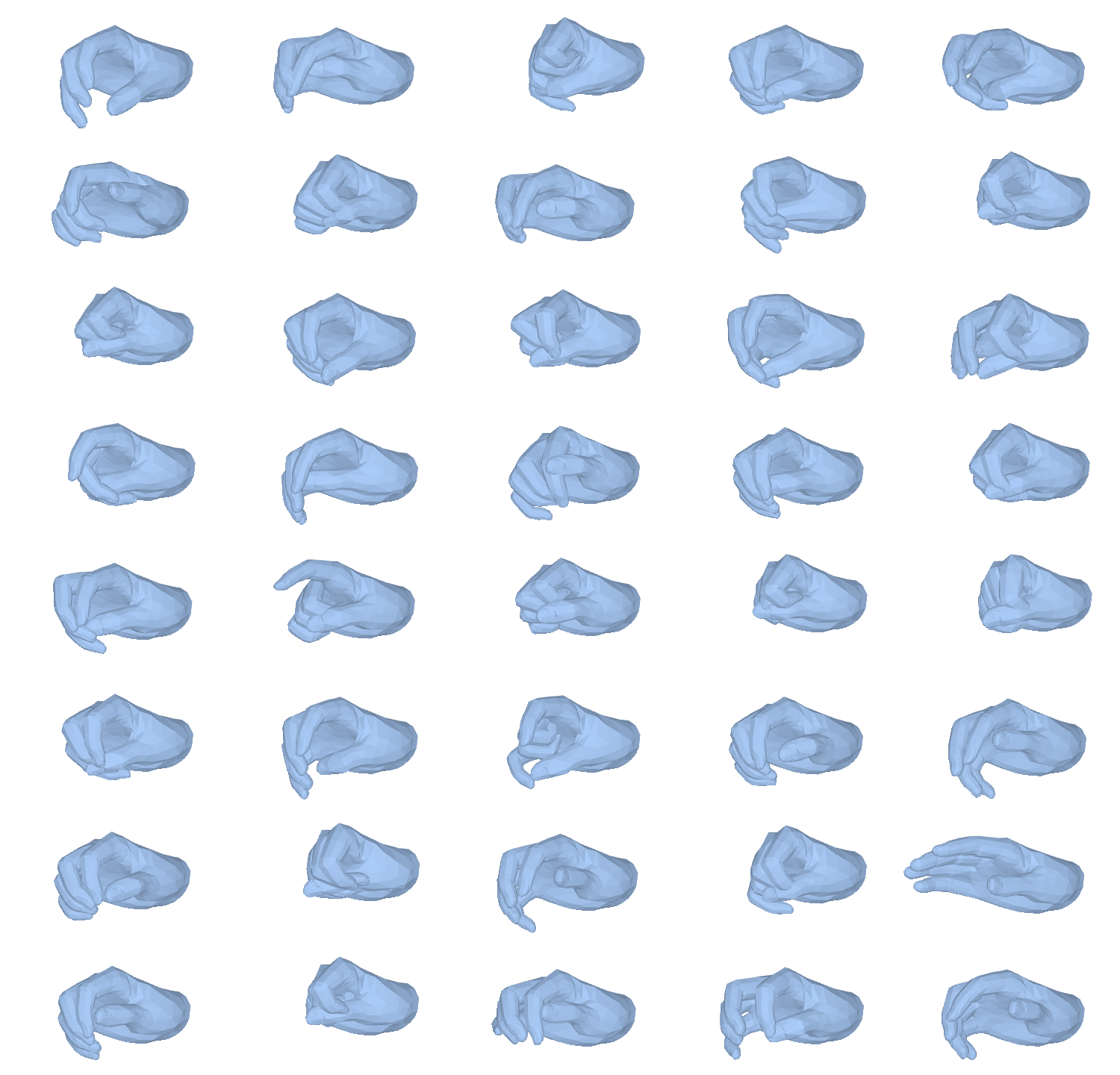}
    }
    \hfill
    \subfloat[NRDF]{
        \includegraphics[width=0.31\linewidth]{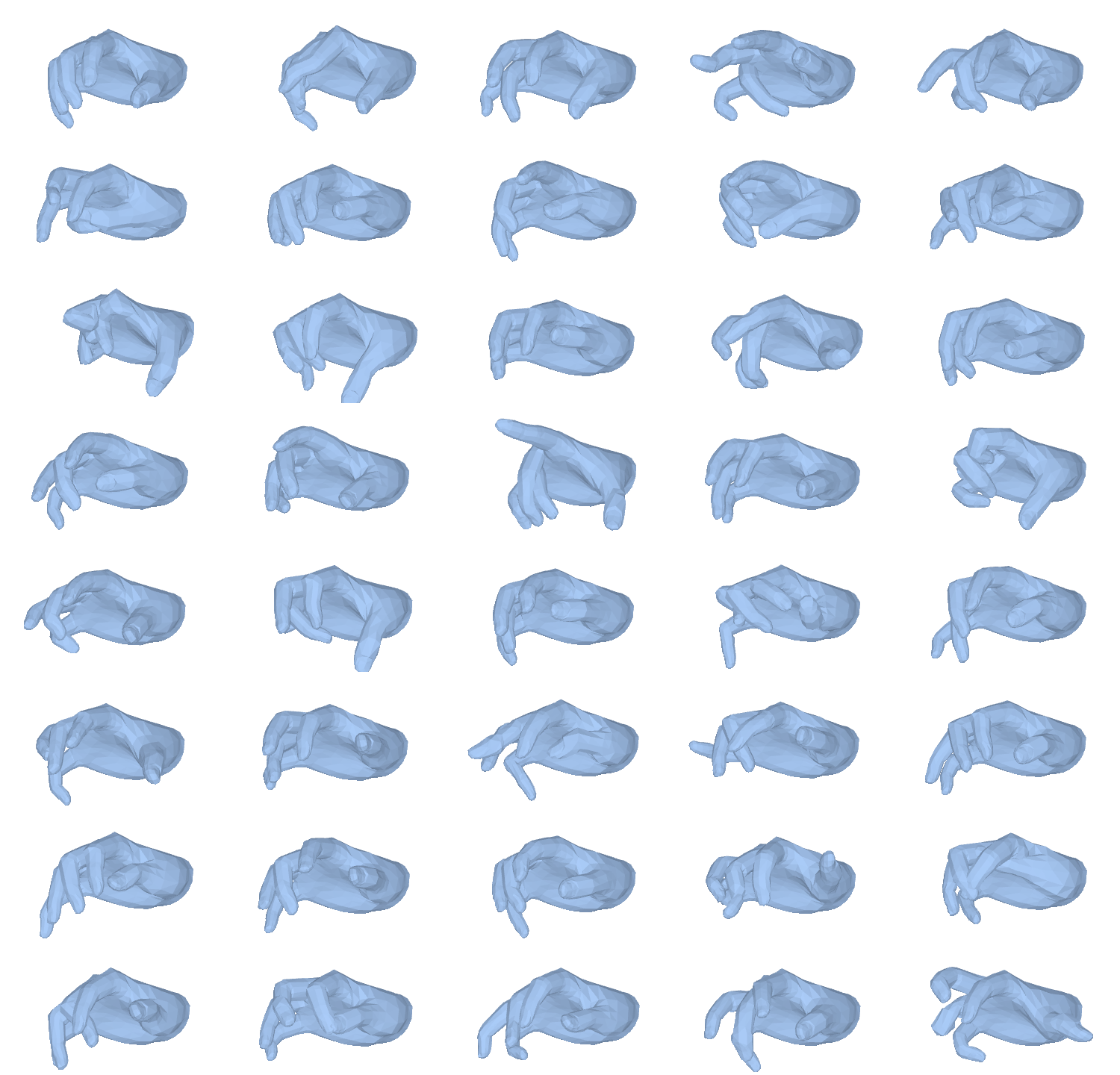}
    }
    \caption{Visualization of hand pose generation. DPoser produces more diverse and realistic hand poses compared to VPoser~\cite{pavlakos2019expressive} and NRDF~\cite{he2024nrdf}.}
    \label{fig:hand_generation}
\end{figure*}

\begin{figure*}[t]
    \centering
    \includegraphics[width=0.98\linewidth]{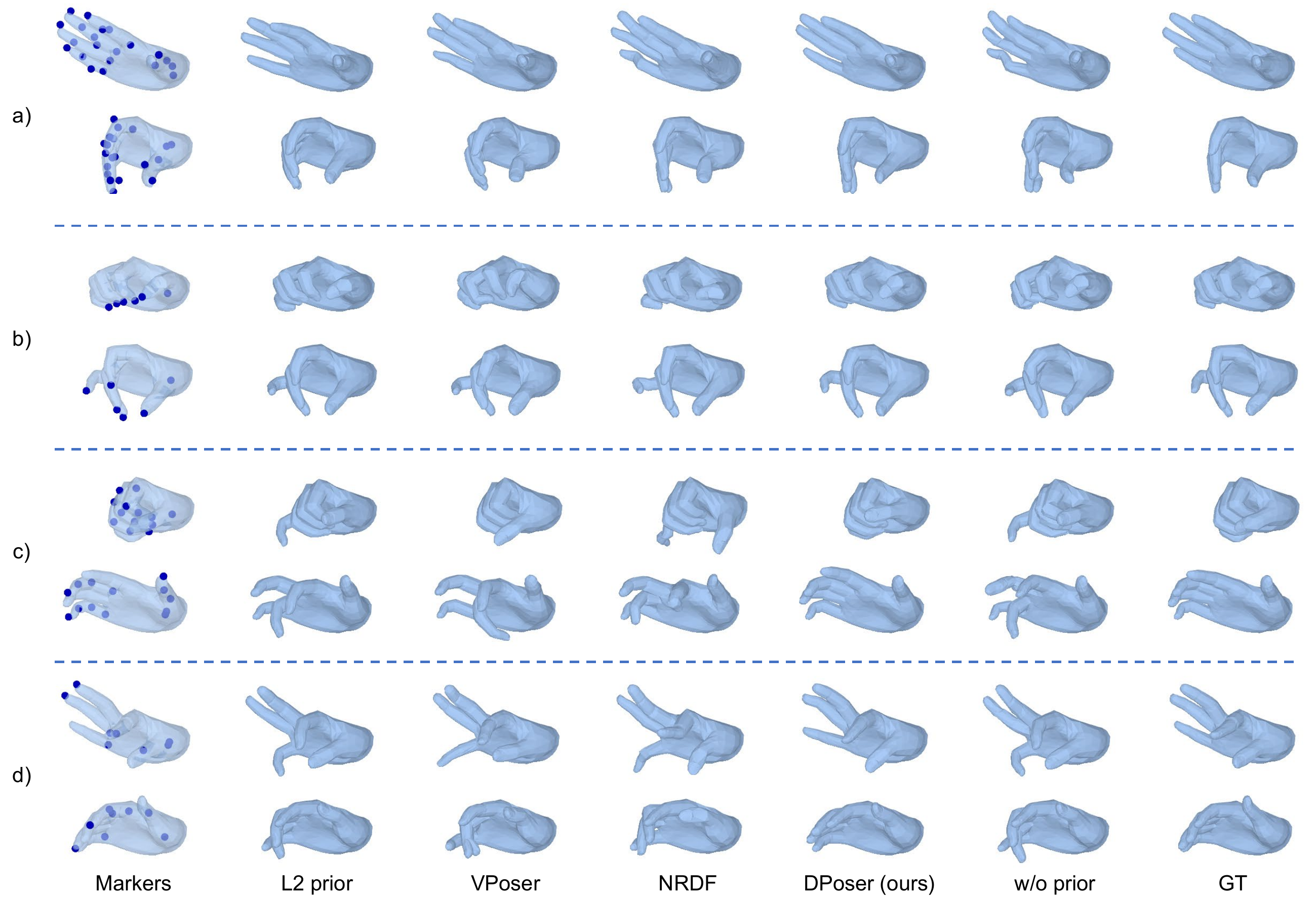}
    \caption{Visualization of hand inverse kinematics under multiple challenging settings. Comparison across (a) noisy keypoints, (b) fingertip keypoints, (c) partial finger keypoints, and (d) sparse keypoints settings.}
    \label{fig:ik_hand_comparison}
\end{figure*}

\begin{figure*}[t]
    \centering
    \includegraphics[width=0.85\linewidth]{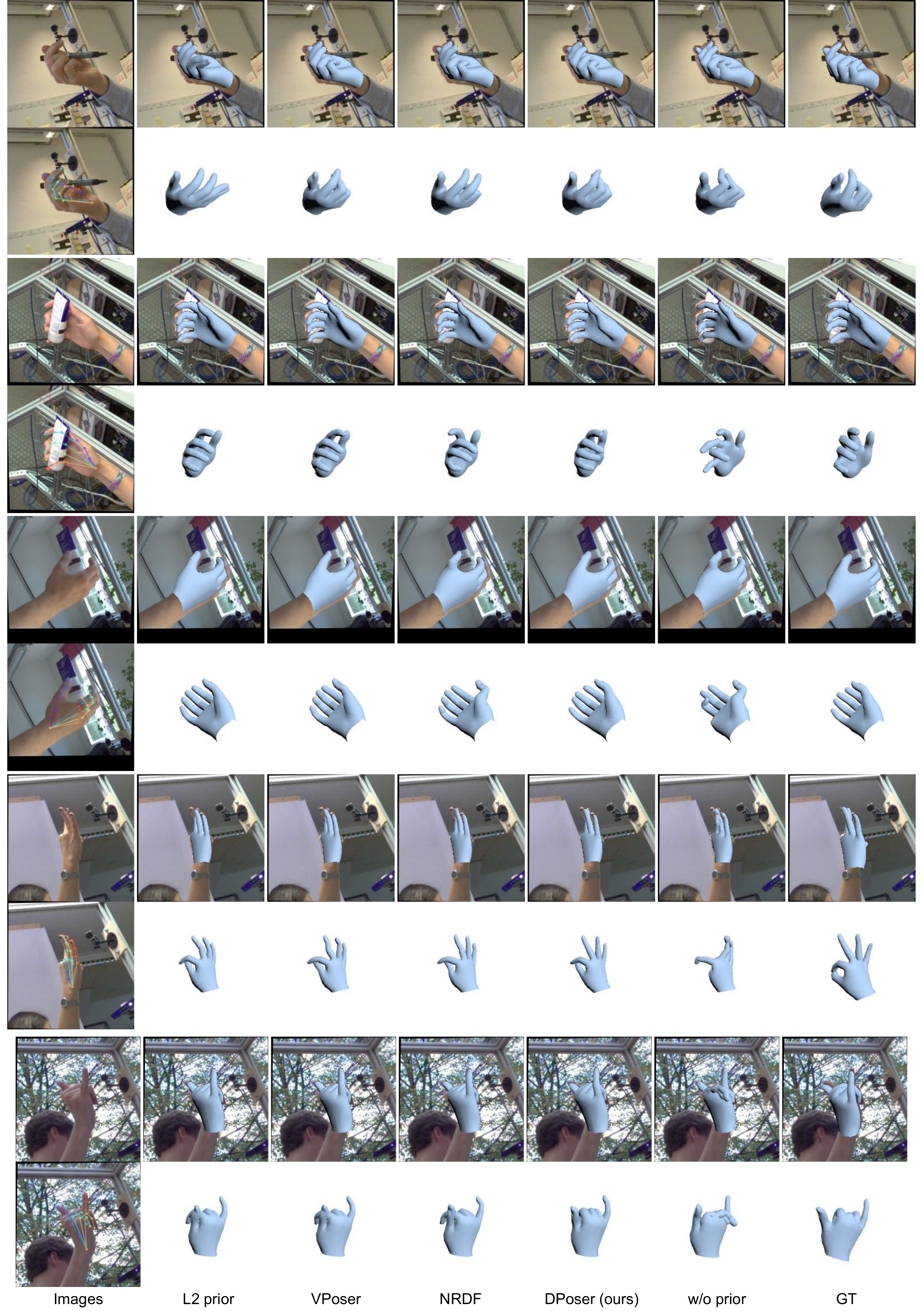}
    \caption{Visualization of hand mesh recovery with mean pose initialization.}
    \label{fig:freihand_mmpose_appendix}
\end{figure*}

\begin{figure*}[t]
    \centering
    \includegraphics[width=0.85\linewidth]{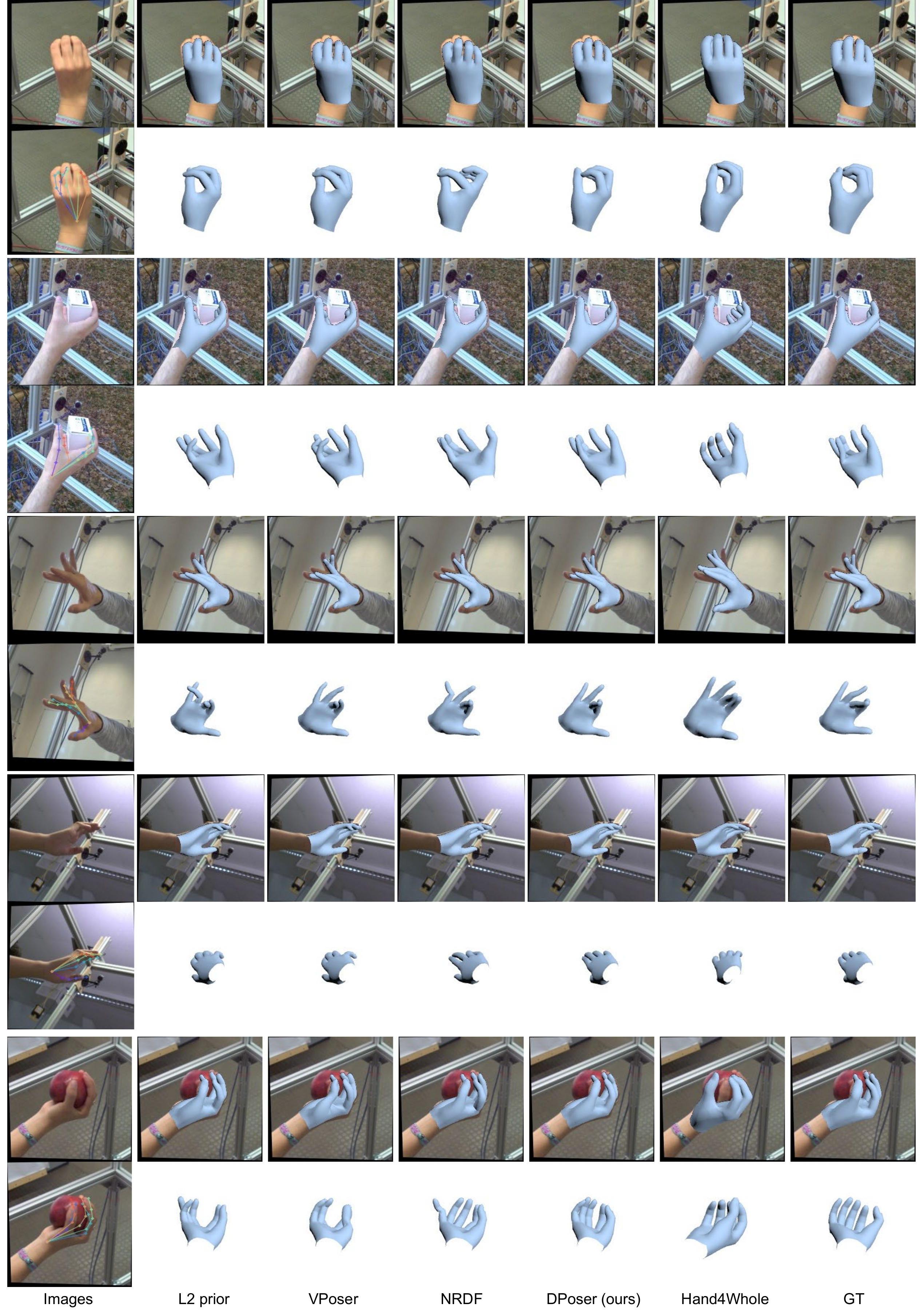}
    \caption{Visualization of hand mesh recovery with Hand4Whole~\cite{moon2022accurate} initialization.}
    \label{fig:freihand_gt_appendix}
\end{figure*}

\begin{figure*}[t]
    \centering
    \includegraphics[width=0.98\linewidth]{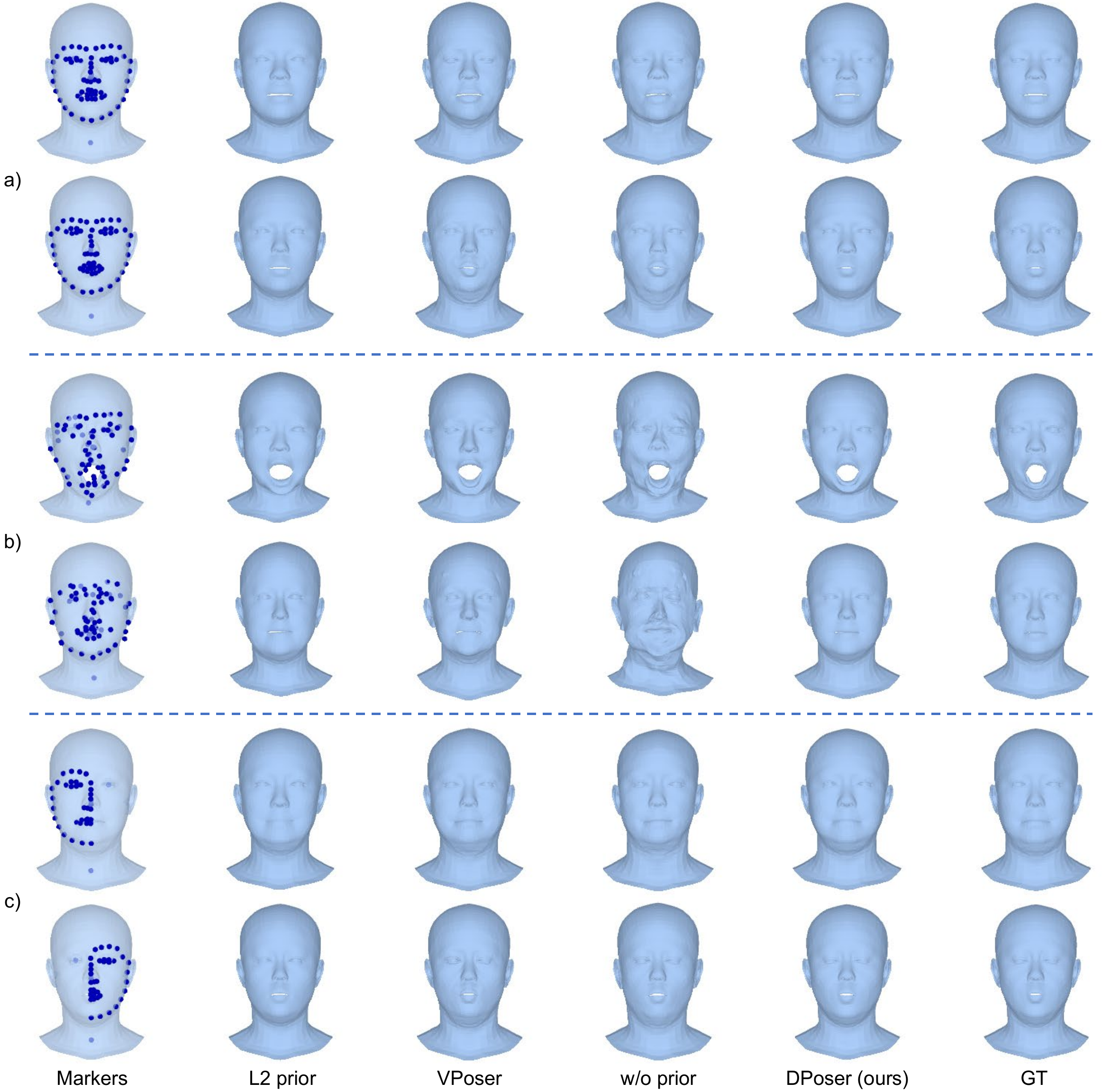}
    \caption{Qualitative results of face inverse kinematics on the WCPA dataset~\cite{kao2022single}. Comparison across (a) 1 mm noise, (b) 5 mm noise, and (c) half-face occlusion. Better zoom in and compare the human eyes and chin.}
    \label{fig:ik_face_comparison}
\end{figure*}

\begin{figure*}[t]
    \centering
    \includegraphics[width=0.98\linewidth]{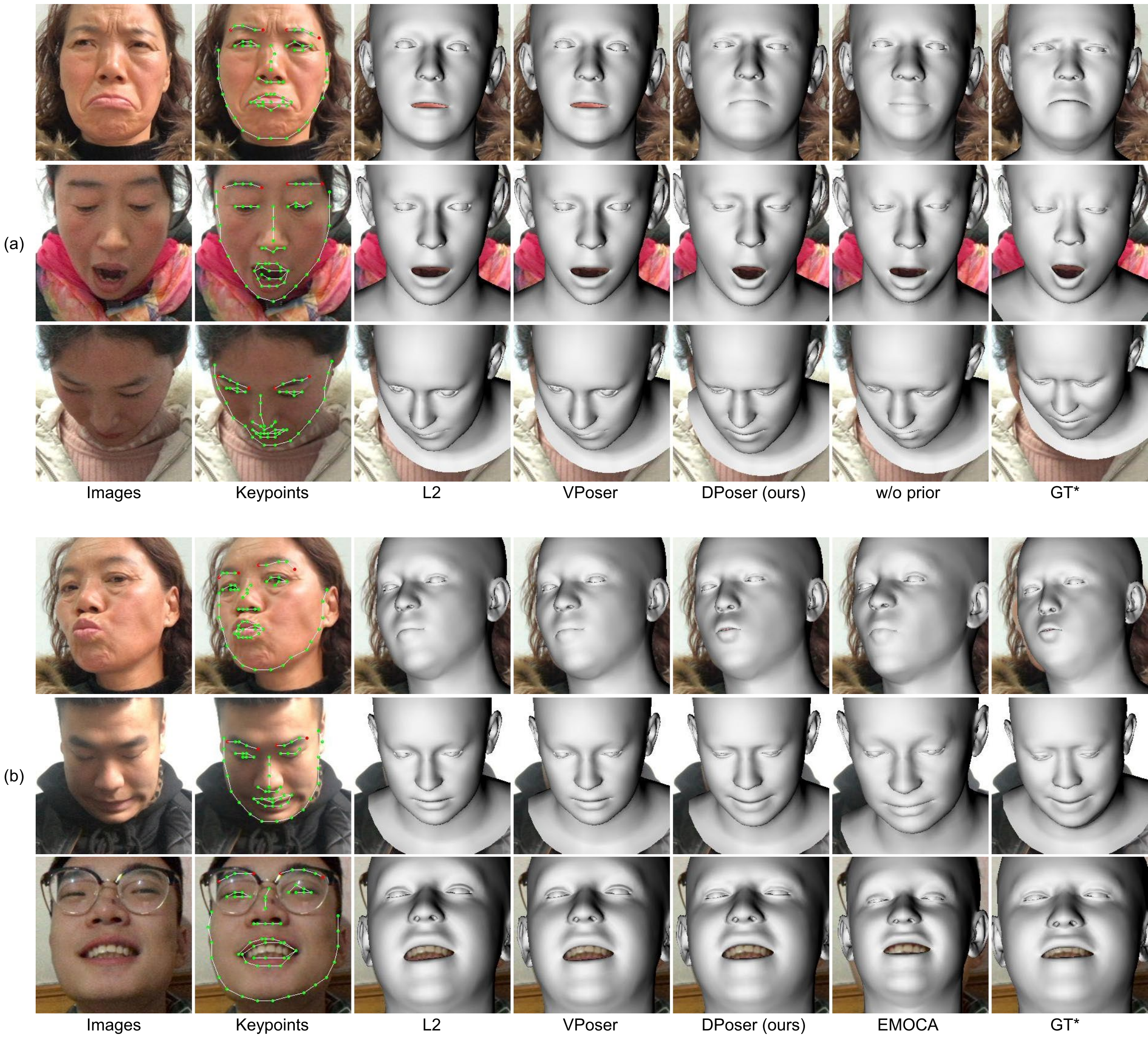}
    \caption{
    Visualization of face reconstruction results on the WCPA dataset~\cite{kao2022single}. Comparisons include (a) fitting from scratch and (b) initialization using EMOCA~\cite{danvevcek2022emoca} results. *Ground truth lacks global orientation and translational data; these are fitted for visualization.}
    \label{fig:wcpa_comparison}
\end{figure*}

\begin{figure*}[t]
    \centering
    \subfloat[VPoser-X]{
        \includegraphics[width=0.48\linewidth]{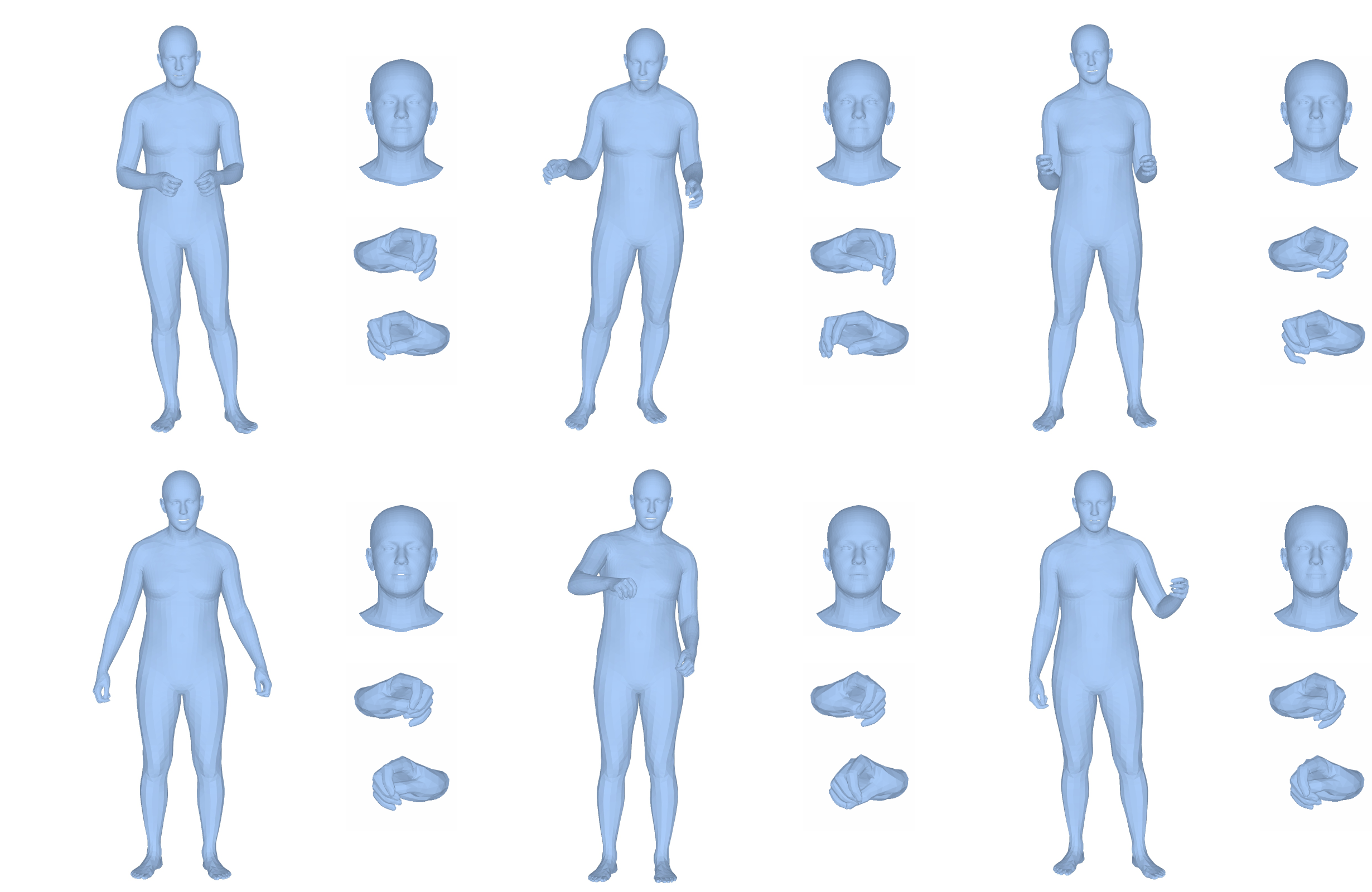}
    }
    \hfill
    \subfloat[DPoser-X-base]{
        \includegraphics[width=0.48\linewidth]{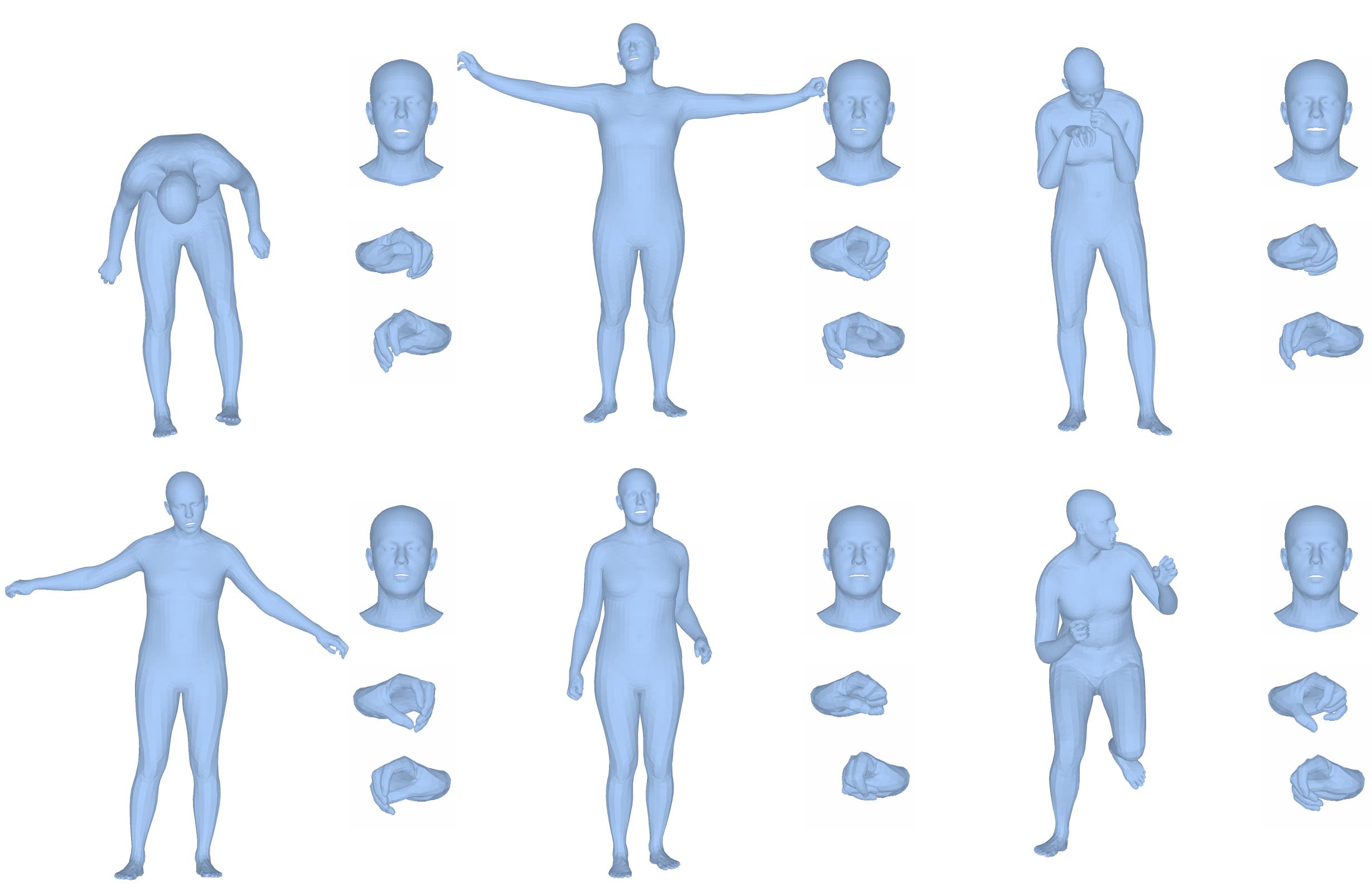}
    }
    \hfill
    \subfloat[DPoser-X-fused]{
        \includegraphics[width=0.48\linewidth]{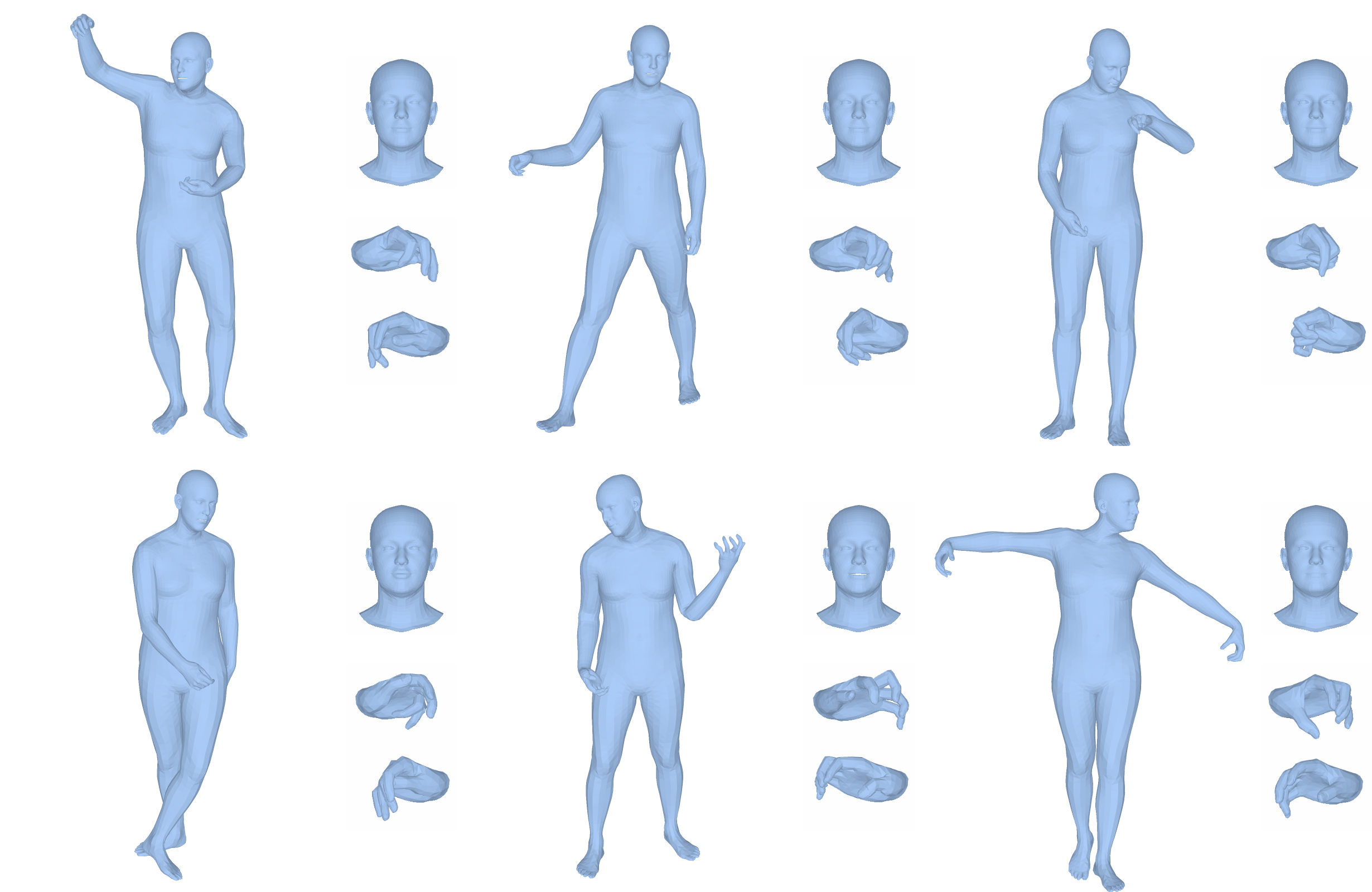}
    }
    \hfill
    \subfloat[DPoser-X-mixed]{
        \includegraphics[width=0.48\linewidth]{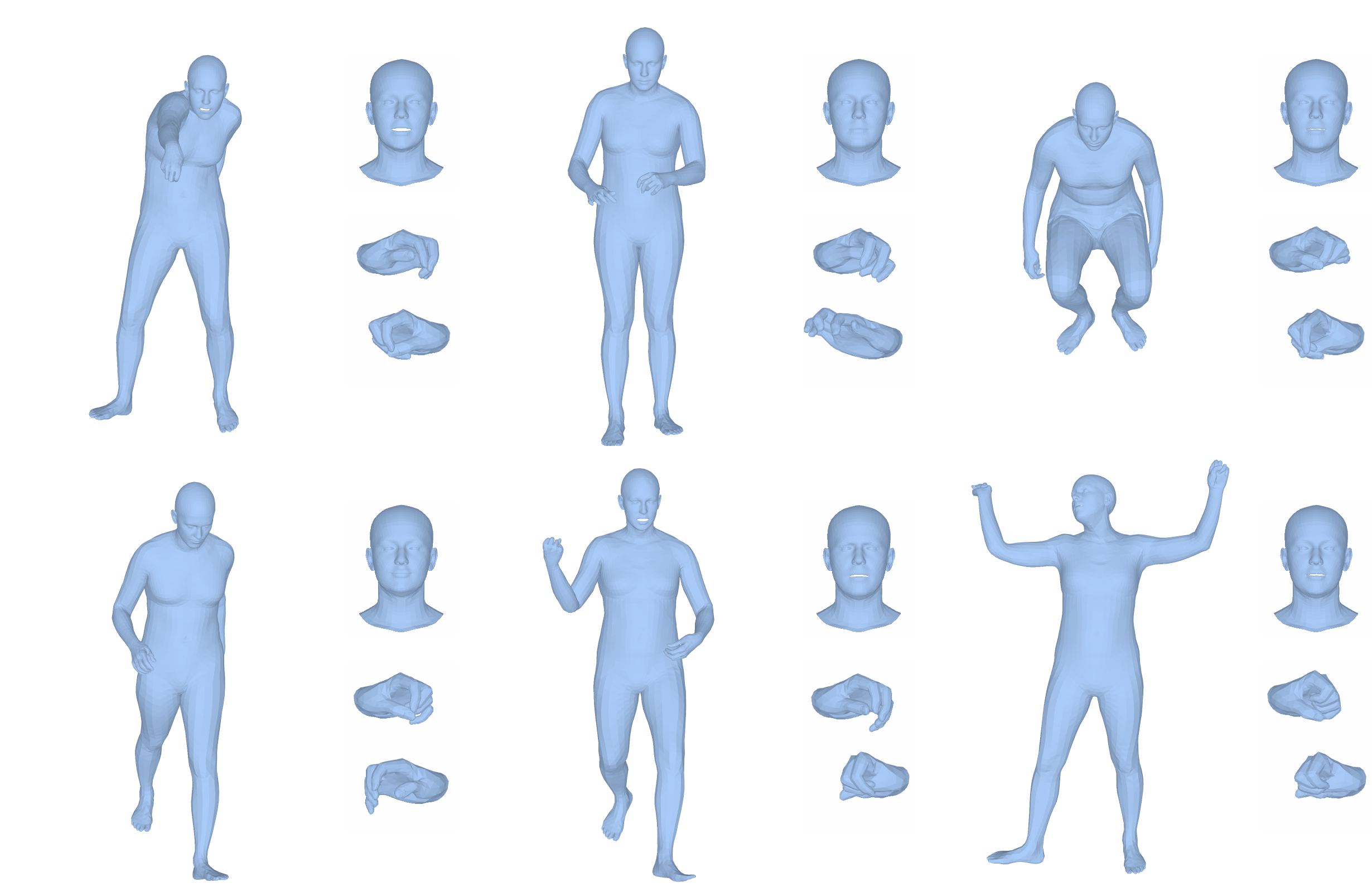}
    }
    
    \caption{Visualization of whole-body pose generation. (a) VPoser-X primarily generates standing poses with limited diversity. (b) DPoser-X-base generates diverse samples but lacks realism in hand interactions and facial expressions. (c) DPoser-X-fused produces less diverse samples while maintaining plausible whole-body poses. (d) DPoser-X-mixed achieves a well-balanced trade-off between diversity and realism.}
    \label{fig:wholebody_generation_paper}
\end{figure*}

\begin{figure*}[t]
    \centering
    \subfloat[DPoser-X-mixed]{
        \includegraphics[width=0.98\linewidth]{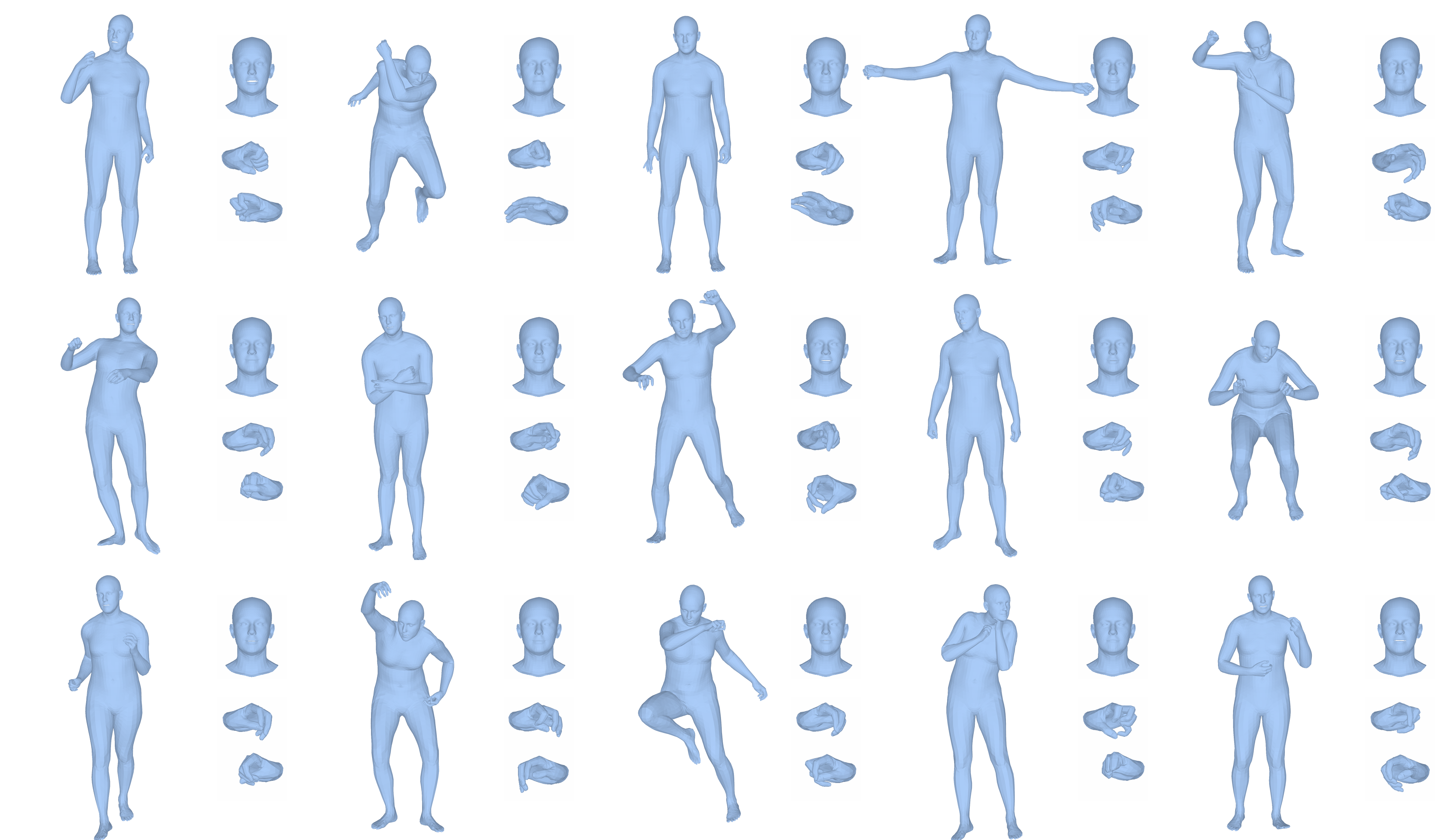}
    }
    \hfill
    \subfloat[DPoser-X-fused]{
        \includegraphics[width=0.98\linewidth]{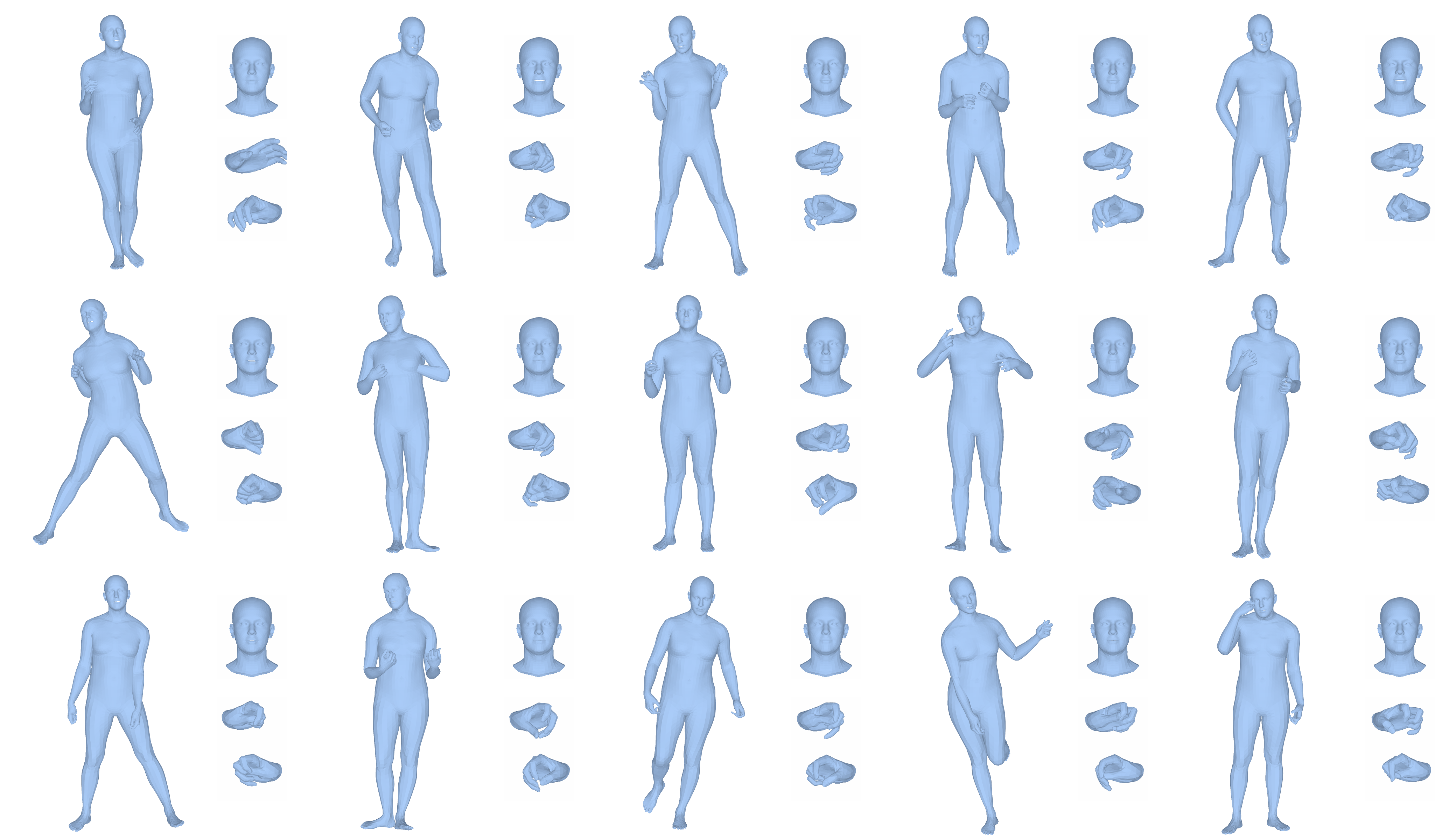}
    }
    \caption{Extended visualization of whole-body pose generation. DPoser-X-mixed generates a diverse range of whole-body poses while maintaining realistic hand interactions and facial expressions. In contrast, DPoser-X-fused retains high realism but produces less diverse results.}
    \label{fig:wholebody_generation_appendix}
\end{figure*}

\begin{figure*}[t]
    \centering
    \includegraphics[width=0.98\linewidth]{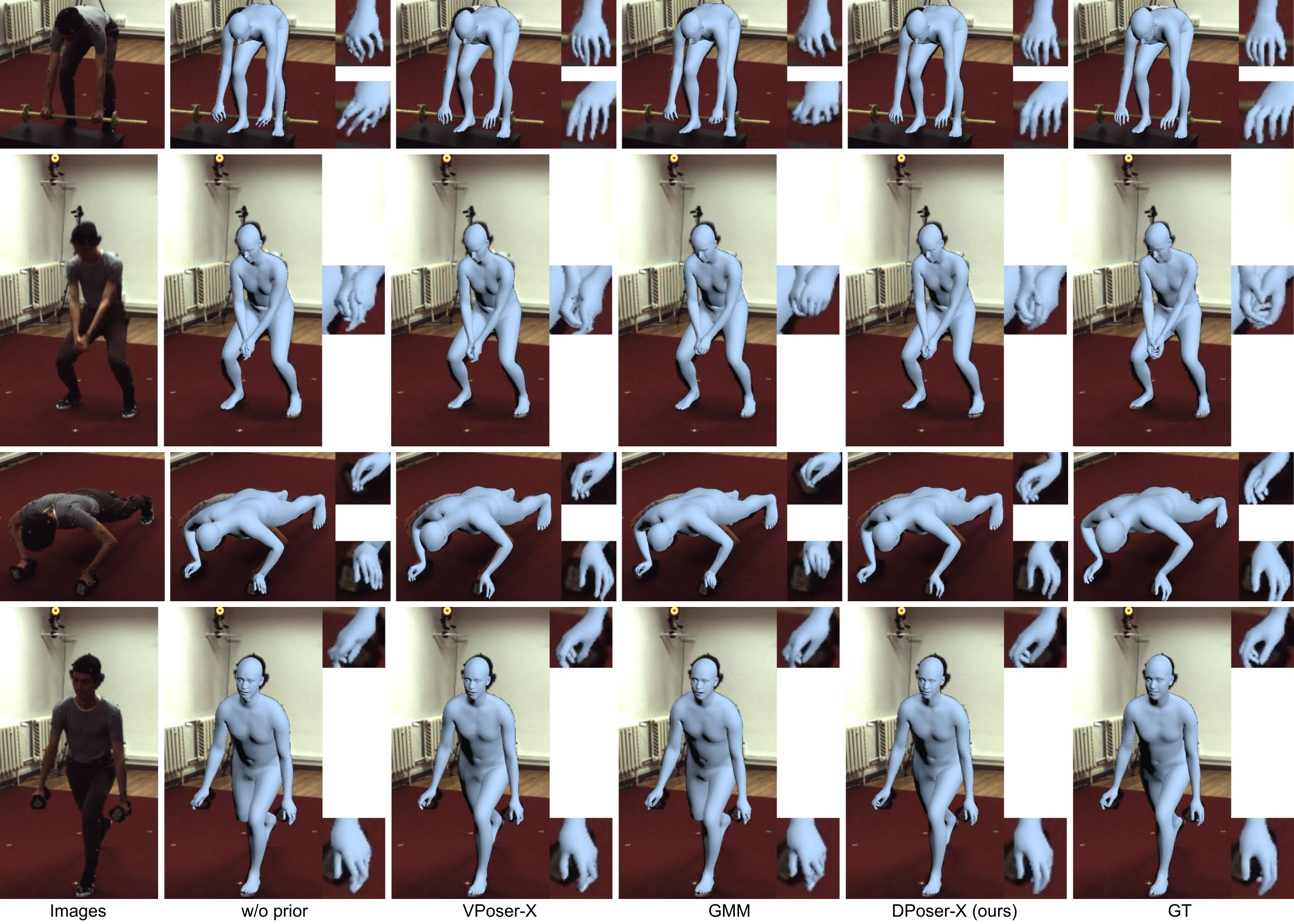}
    \caption{Visualization of whole-body mesh recovery on the Fit3d dataset~\cite{fieraru2021aifit}.}
    \label{fig:fit3d_comparison}
\end{figure*}

\begin{figure*}[t]
    \centering
    \includegraphics[width=0.98\linewidth]{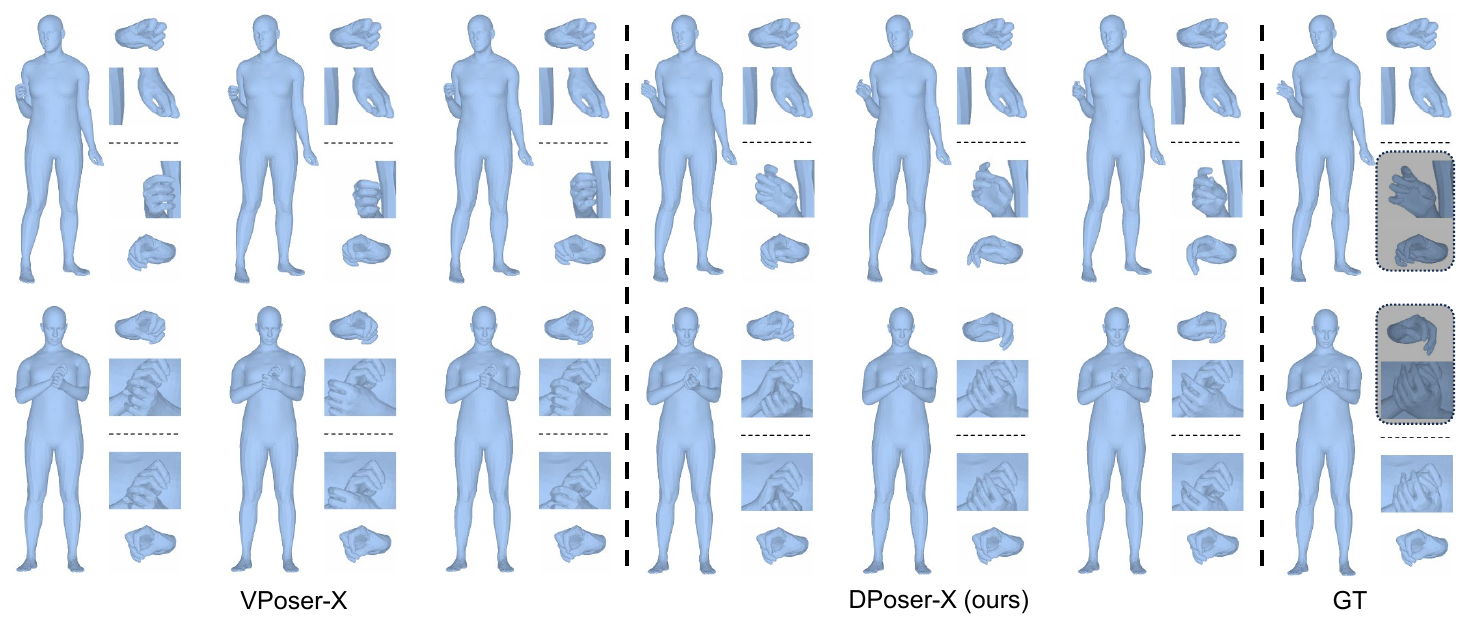}
    \caption{Qualitative comparison of whole-body pose completion. One hand is masked randomly.}
    \label{fig:wholebody_completion_comparison}
\end{figure*}

\begin{figure*}[t]
    \centering
    \includegraphics[width=0.98\linewidth]{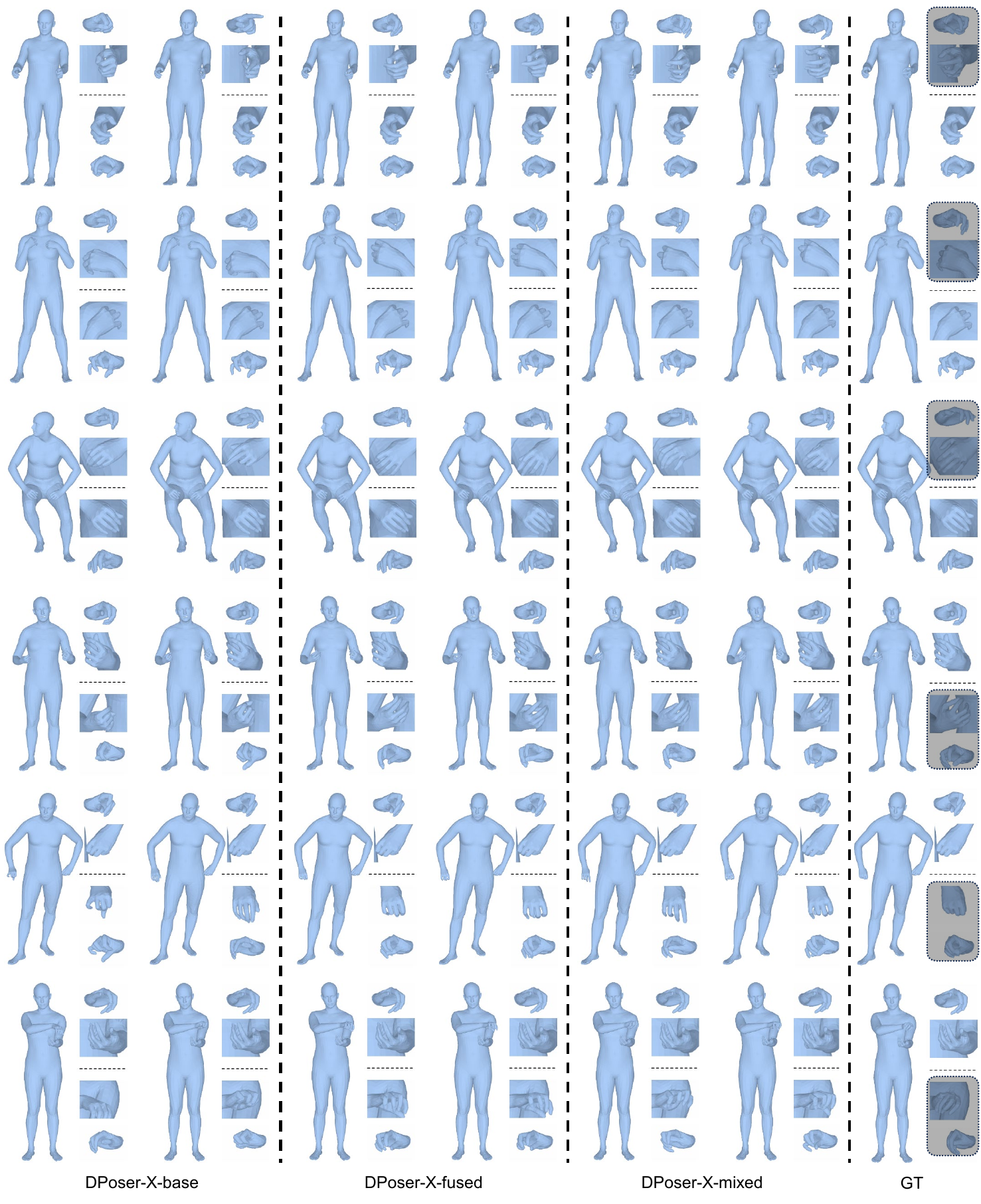}
    \caption{Visualization of whole-body pose completion for three DPoser-X variants. One hand is masked randomly.}
    \label{fig:wholebody_completion_ablation}
\end{figure*}

We show more qualitative results for body pose generation (Fig.~\ref{fig:more generation}), body pose completion (Fig.~\ref{fig:more completion}), body mesh recovery (Fig.~\ref{fig:more hmr}), motion denoising (Fig.~\ref{fig:more motion} and Fig.~\ref{fig:motion completion}), hand generation (Fig~\ref{fig:hand_generation}), hand inverse kinematics (Fig~\ref{fig:ik_hand_comparison}), hand mesh recovery (Fig~\ref{fig:freihand_mmpose_appendix} and Fig~\ref{fig:freihand_gt_appendix}), face inverse kinematics (Fig~\ref{fig:ik_face_comparison}), face reconstruction (Fig~\ref{fig:wcpa_comparison}), whole-body pose generation (Fig~\ref{fig:wholebody_generation_paper} and Fig.~\ref{fig:wholebody_generation_appendix}), whole-body mesh recovery (Fig~\ref{fig:fit3d_comparison}), whole-body pose completion (Fig~\ref{fig:wholebody_completion_comparison} and Fig.~\ref{fig:wholebody_completion_ablation}).

\newpage

\end{document}